\documentclass{article}
\usepackage{iclr2027_conference,times}
\iclrfinalcopy

\usepackage{amsmath,amsfonts,bm}

\def\eqref#1{equation~\ref{#1}}

\def\1{\bm{1}}

\DeclareMathAlphabet{\mathsfit}{\encodingdefault}{\sfdefault}{m}{sl}
\SetMathAlphabet{\mathsfit}{bold}{\encodingdefault}{\sfdefault}{bx}{n}

\usepackage{hyperref}
\usepackage{url}
\usepackage{booktabs}
\usepackage{array}
\usepackage{graphicx}
\usepackage{flafter}
\usepackage{amsmath,amssymb}
\usepackage{microtype}
\usepackage{placeins}

\title{RoutePrism: Tracing Construction Order\\Effects in Agent Memory}

\author{
\textbf{Dong Xu$^{1,2}$, Zhangfan Yang$^{3}$, Jiantao Wu$^{1}$, Shipeng Zhang$^{1}$, Zexuan Zhu$^{1}$,}\\
\textbf{Jiangqiang Li$^{1}$, Jun Zhang$^{1}$, Junkai Ji$^{1,2,\dagger}$}\\
\normalfont$^{1}$School of Artificial Intelligence, Shenzhen University\\
$^{2}$EasternDawn\\
$^{3}$School of Computer Science, University of Nottingham Ningbo\\
$^\dagger$Corresponding author.}

\begin{document}

\maketitle
\fancyhead[L]{}

\begin{abstract}
Processing the same records in a different order can discard different evidence, yet endpoint accuracy alone cannot reveal what changed or whether it mattered.
We introduce RoutePrism, a diagnostic protocol that builds memory twice from the same source pool in two processing orders, then traces which sources, compiled contexts, and answers differ. Because record content, timestamps, policy, and the answer model all stay fixed, any observed difference is localized to the memory construction step. A matched four-condition intervention tests whether a record displaced by reordering actually carried task-relevant evidence: restoring that single record recovers over 60 percentage points of lost accuracy, while substituting a non-supporting record of equal length does not.
We evaluate the protocol on PersonaMem-32K (63 primary queries, 29 users) and 470 LongMemEval-S questions with histories spanning 38 to 62 sessions, replicating the core intervention across five answer models. Survivor selection, defined as the choice of which record a cluster retains, drives most source-level changes, while different memory policies (compaction, bounded recency, MemoChat-style summarization, A-MEM) produce distinct failure signatures at the source, context, and metadata layers.
\end{abstract}

\begin{figure*}[t]
\centering
\includegraphics[width=0.90\textwidth]{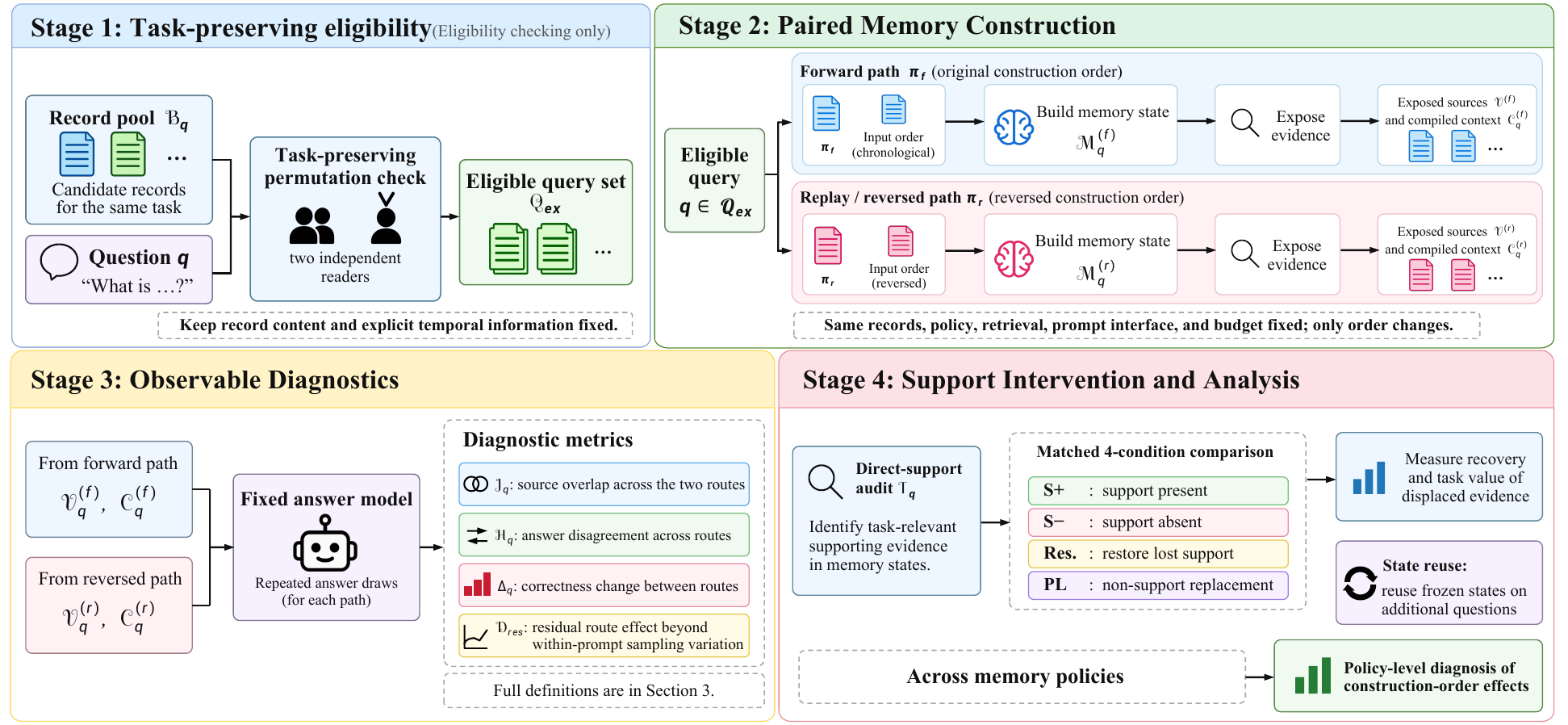}
\vspace{-2mm}
\caption{\textbf{RoutePrism overview.} Given a question and its source pool, we build memory in two processing orders, record which sources and contexts differ, and test whether the displaced evidence matters via a four-condition support-restoration intervention. The protocol isolates construction-order effects from answer-model noise.}
\label{fig:overview}
\vspace{-4mm}
\end{figure*}

\section{Introduction}

A user tells an assistant about the joy of creating a budget plan with friends. Days later, the same user discusses unrelated financial activities. When a new question asks the assistant to recall that budgeting experience, the answer depends on which record survived memory construction. This selection, in turn, depends on the order in which the records were processed. Reverse the processing order, and a different record survives; the assistant loses its only evidence for the correct answer. Accuracy on a test set will not flag this: the score conflates source selection, context compilation, and answer sampling into a single number.

Memory systems such as MemGPT~\citep{packer2023memgpt} and MemoryBank~\citep{zhong2023memorybank} must select and organize past interactions under bounded budgets. As records accumulate, a policy may cluster related observations and keep one representative, or summarize several records into a single entry. Each of these decisions can depend on arrival order. Under a fixed budget, reordering the same records can therefore change which evidence remains available---even when the input history is otherwise identical~\citep{liu2026segtreemem}.
Whether a replaced record matters depends on its contents. If two records state the same fact, swapping one for the other changes nothing. But if only the discarded record supports the correct answer, the agent loses its evidence and may answer incorrectly. Standard accuracy metrics hide this: a model can produce the right answer from wrong evidence, and gains on some questions can mask losses on others. Benchmarks such as LongMemEval~\citep{wu2025longmemeval} and PersonaMem~\citep{personamem2025} measure memory-dependent performance, but to interpret construction-order effects we also need to know \emph{what evidence changed} and \emph{whether the change mattered}.

Comparing two memory configurations on a test set reveals \emph{that} accuracy changed, but not \emph{why}: did the source set change, did the compiled context change with fixed sources, or did the answer model simply sample differently? Disentangling these stages requires an evaluation protocol that controls everything except the variable under study and records intermediate states alongside the final score.

Prior work on order sensitivity focuses on prompts and retrieval-augmented generation~\citep{lu2022ordered,liu2024lostmiddle}. Stable-RAG~\citep{zhang-etal-2026-stable} permutes an already-retrieved document set. SegTreeMem~\citep{liu2026segtreemem} studies temporally ordered memory construction. RoutePrism traces changes in the sources and contexts produced by construction, then restores a displaced source to measure its task consequence. This requires separating processing order from task chronology---dates and within-record event relations stay fixed across reorderings, so the correct answer is preserved even as the processing order changes.

We introduce RoutePrism (Figure~\ref{fig:overview}), a diagnostic protocol that pairs two construction paths over a fixed source pool and traces their effects at three layers: exposed sources, compiled contexts, and generated answers. RoutePrism asks a specific question that accuracy alone cannot answer: \emph{did this memory policy lose evidence that matters, and if so, where in the pipeline did the loss occur?} We evaluate the protocol on PersonaMem-32K (63 primary queries from 29 users) and 470 LongMemEval-S questions with longer native histories (38--62 sessions), replicating the core intervention across five answer models (Astra, Sol, Terra, Luna, Qwen3-8B). Unless stated otherwise, extended results use Luna. Our contributions:
\vspace{-2mm}
\begingroup
\setlength{\leftmargini}{0.5\leftmargini}
\begin{itemize}
\item \textbf{A diagnostic protocol for construction-order effects.} RoutePrism screens for reorderings that preserve the correct answer, then records which sources and contexts change between the two construction paths. Repeated answers at each fixed prompt separate route-level variation from answer-model sampling noise.
\item \textbf{Evidence that displaced records carry task value.} In matched support-loss pairs, restoring a single displaced record recovers over 60 percentage points of lost accuracy under both compaction and bounded recency. Substituting a non-supporting record of equal length does not. This pattern replicates across all five answer models, with model-dependent magnitude.
\item \textbf{Distinct failure signatures across memory policies.} Compaction changes which source records survive. MemoChat-style summarization changes the compiled text while largely preserving source identities. A-MEM changes internal metadata without affecting the answer prompt. These cases show that no single observation layer is sufficient.
\end{itemize}
\endgroup

\section{Related Work}

\textbf{Long-horizon memory.} Research on persistent memory for LLM assistants has explored explicit storage mechanisms~\citep{packer2023memgpt,zhong2023memorybank}. Evaluations cover long-term conversational memory~\citep{maharana-etal-2024-evaluating}, multi-session temporal reasoning~\citep{wu2025longmemeval}, user profiling~\citep{personamem2025}, and stateful agent workflows~\citep{hu2026memoryagentbench,ama-bench2026}. We build on PersonaMem's released evaluation split and LongMemEval's evidence labels. Appendix~\ref{app:related} reviews a broader set of benchmarks.

\textbf{Order and construction.} Order sensitivity in prompts and retrieval is well documented~\citep{lu2022ordered,liu2024lostmiddle}. Stable-RAG~\citep{zhang-etal-2026-stable} extends this analysis to permutations of retrieved document sets. Recent work examines how order affects memory construction itself: SegTreeMem~\citep{liu2026segtreemem} studies temporally ordered construction, while \citet{zhang2026faultymemory} trace consolidation failures. RoutePrism pairs two construction paths that preserve the correct answer, then traces which sources survive and whether displaced records carry task-relevant evidence.

\textbf{Structure and provenance.} A growing line of work organizes persistent state through hierarchical schemas~\citep{rezazadeh2024memtree}, structured retention and belief separation~\citep{latimer2026hindsight}, compiled instructions~\citep{rhodes2026compiled}, or evolving notes~\citep{xu2025amem}. On the evaluation side, \citet{wang2026tracetrust} survey evidence tracing. AttriMem~\citep{li2026attrimem} uses attribution feedback to improve construction, and AcquaBench~\citep{luo2026acquabench} audits success provenance through gold-versus-sham substitutions. RoutePrism restores a record that was displaced during construction and compares the resulting recovery with a matched non-supporting replacement, directly measuring the task consequence of the observed loss.

\section{RoutePrism}
\label{sec:routeprism}

\subsection{A Paired Construction Intervention}

RoutePrism builds memory twice from the same records, changing only the processing order, and observes the resulting differences (Figure~\ref{fig:overview}). The records themselves, the policy, retrieval settings, prompt interface, and context budget stay fixed, so an observed difference is attributable to how the construction step handled the changed arrival sequence.

\noindent\textbf{Two routes and what we observe.} We feed the same source pool $B_q$ to the memory policy in two orders: the benchmark listing order (forward) and its exact reverse (replay). Record text, identifiers, dates, and within-record event relations are unchanged---only the arrival sequence differs. Construction receives the pool and the schedule; it never sees the question, answer options, or gold label (Appendix~\ref{app:setting-formal-interface}).
After construction, we record which source identifiers the policy exposed ($V_q^{(o)}$) and the compiled context ($C_q^{(o)}$) before generating an answer. The same answer model then produces a response for each route. By recording sources before answers, we can tell whether a changed answer traces back to a changed source set or to a changed compilation.

\noindent\textbf{Example.} A user has four conversation records; a question asks about the user's budgeting experience. The compactor clusters all four records and keeps only the last to arrive. Forward construction keeps record 3 (about later financial activities); reversal keeps record 0 (about the joy of creating the budget). Record 0 directly supports the correct answer; record 3 does not. A single change in arrival order replaces the only relevant evidence.

Eligibility requires that the complete source pool support the same correct answer regardless of record arrangement. Two independent annotators screen each question from the question text, answer options, and full source pool, without seeing the gold label or any route outcomes. Their agreed-admissible intersection defines the primary set of 63 queries from 29 users ($\kappa=0.634$). Appendix~\ref{app:semantic} reports the full readings; Appendix~\ref{app:checks} evaluates a stricter chronology-aware rubric.

\subsection{Observable Coordinates}

We summarize each paired comparison with three numbers. Let $c_q^{(o)}=\mathbf{1}[a_q^{(o)}=y_q]$ indicate whether route $o$ produces the correct answer:
\begin{equation}
J_q=\frac{|V_q^{(f)}\cap V_q^{(r)}|}{|V_q^{(f)}\cup V_q^{(r)}|},\qquad H_q=\mathbf{1}[a_q^{(f)}\ne a_q^{(r)}],\qquad \Delta_q=c_q^{(r)}-c_q^{(f)}.
\label{eq:profile}
\end{equation}
$J_q$ measures how much the exposed source sets overlap (0 = completely different sources, 1 = identical). $H_q$ flags whether the two routes produce different answers. $\Delta_q$ captures the direction of correctness change: $+1$ when replay corrects a forward error, $-1$ when it introduces one. Note that gains and losses can cancel across questions even when answers frequently differ ($H_q=1$ with $\Delta_q=0$ when both answers are wrong).

The three coordinates are supplemented by the compiled context $C_q^{(o)}$ and, where available, a policy-specific internal-state signature: a summary can change with fixed source identifiers ($J_q=1$), and different source sets can yield equivalent text. Throughout, we average within each user first, then give users equal weight; 95\% bootstrap intervals resample users (details in Appendix~\ref{app:setting-estimands}).

\subsection{Support Restoration and Answer-Stage Controls}
\label{sec:support-answer}

The next test asks whether a displaced record carried evidence that matters for the task. We use a four-condition intervention:
\begingroup
\setlength{\leftmargini}{0.5\leftmargini}
\begin{itemize}
\item Support present ($S^+$): the context includes the record that directly supports the correct answer.
\item Support absent ($S^-$): that record has been displaced by the alternative construction path.
\item Restored (Res.): we put the displaced record back into the context.
\item Replacement (Pl.): we substitute a record from the same pool that does \emph{not} directly support the answer.
\end{itemize}
\endgroup
All four prompts share the same question, answer options, source count, canonical ordering, and total token length (matched under \texttt{o200k\_base}); only the identity of one record differs. In the budget example, the four conditions expose record~0, record~3, restored record~0, and record~1 (a non-supporting replacement). If restoration recovers accuracy but replacement does not, the displaced record's task value comes from its specific content, not just its slot.

An offline audit identifies which records directly support the correct answer. The audit sees the question, options, correct answer, and complete source pool, but never sees route outcomes or generated answers. We retain only pairs where one supporting record is displaced and all four conditions pass the structural matching requirements (Appendix~\ref{app:additional-route-rescue}).

\noindent\textbf{Separating route effects from answer noise.} Even with identical prompts, a language model may produce different answers across calls. To distinguish genuine route effects from this sampling noise, we draw $m\geq2$ answers from each fixed-route prompt and compute a residual:
\begin{equation}
D_{q,\mathrm{res}}=H_{q,\times}-H_{q,\mathrm{within}},
\label{eq:residual}
\end{equation}
where $H_{q,\times}$ is the average disagreement across all forward--replay answer pairs and $H_{q,\mathrm{within}}$ is the average disagreement between draws from the \emph{same} route. A positive $D_{\mathrm{res}}$ means the two routes disagree more than same-prompt sampling alone would predict. Its population interpretation is given in Appendix~\ref{app:setting-estimands}.

\section{Experiments}
\label{sec:results}

\begin{table}[t]
\vspace{-4mm}
\centering
\small
\caption{Experiment overview. Queries and Users give each experiment's analysis units; NR means the benchmark supplies no user identifier, and NA denotes a source-only measurement without an answer model. Best Matching 25 (BM25) is the lexical retriever for LongMemEval-S. Extended answer-bearing rows report Luna estimates except for model comparisons; the complete Qwen3-8B collection is retained, and its LongMemEval answers receive the same masked Terra judge.}
\label{tab:experiment-overview-main}
\setlength{\tabcolsep}{3pt}
\begin{tabular*}{\linewidth}{@{\extracolsep{\fill}}llllrr@{}}
\toprule
\textbf{Experiment} & \textbf{Policy or interface} & \textbf{Benchmark} & \textbf{Answer model} & \textbf{Queries} & \textbf{Users}\\
\midrule
Route diagnostics & Focal, Recent-3 & PersonaMem & Luna & 63 & 29\\
Support restoration & Focal & PersonaMem & Luna & 33 & 21\\
Support restoration & Recent-3 & PersonaMem & Luna & 36 & 18\\
Mechanism localization & Focal schedules & PersonaMem & Luna & 63 & 29\\
Policy interfaces & Lexical, cross-fitted & PersonaMem & Luna & 589 & 37\\
Policy interfaces & MemoChat, A-MEM & PersonaMem & Luna & 63 & 29\\
Long-history coverage & Recent-8 + BM25 & LongMemEval-S & Luna/Qwen3 + Terra & 470 & NR\\
Long-history retention & Recent-8/16/32, Focal & LongMemEval-S & NA & 470 & NR\\
Core route replication & Focal exact & PersonaMem & Five models & 63 & 29\\
Core support restoration & Focal & PersonaMem & Five models & 33 & 21\\
Core support restoration & Recent-3 & PersonaMem & Five models & 36 & 18\\
\bottomrule
\end{tabular*}
\vspace{-3mm}
\end{table}

\noindent\textbf{Data.} We use the released PersonaMem-32K evaluation split: 589 queries from 37 users, each with 4--7 source records. Questions from the same user share one source pool. From the fact-recall subset (146 queries), two independent model readings screen for task-preserving reorderings ($\kappa=0.634$); their agreed-admissible intersection gives the primary set of 63 queries from 29 users. A separate audit labels which records directly support the correct answer.

\noindent\textbf{Construction policies.} We evaluate six memory interfaces, chosen to test different layers of the construction process (Table~\ref{tab:experiment-overview-main}). The \emph{focal compactor} clusters arriving records by token Jaccard overlap ($\geq 0.055$) and keeps the last arrival in each cluster---a controlled design that isolates survivor selection as a failure mode. \emph{Bounded recent} (Recent-3) keeps the three most recent arrivals. A \emph{lexical reference} preserves all records and isolates index order. \emph{Cross-fitted streaming} uses a learned retention score. \emph{MemoChat} summarizes and retrieves topic-organized chunks. \emph{A-MEM} maintains evolving notes with metadata. Each policy is paired with a source-order comparator that fixes all order-dependent decisions to benchmark order. Full construction details are in Appendices~\ref{app:construction}--\ref{app:policy}.

\noindent\textbf{Answer models.} Extended results use Luna (\texttt{gpt-5.6-luna}) unless stated otherwise. The core fixed-input matrices are answered by five models: Astra, Sol, Terra, Luna, and Qwen3-8B (greedy decoding). This separates answer-model variation from memory-state variation. All proprietary calls use \texttt{max} reasoning effort. Prompt controls (exact-token matching, offset alignment) are detailed in Appendix~\ref{app:answer}.

\subsection{Does Construction Order Change the Evidence?}

Reversing the processing order changes the exposed sources. For the focal compactor, nearly all source overlap is lost ($J=0.057$). The bounded recent policy retains more common evidence ($J=0.166$) and shows a smaller correctness shift (Table~\ref{tab:main-route}; full profiles in Appendix~\ref{app:answer-exact-token}).
Under both policies, route disagreement substantially exceeds variation within a fixed prompt ($D_{\mathrm{res}}>0.4$), indicating that the two construction paths produce different answer distributions. The residual is not concentrated in a few queries: all 63 focal pairs change their exposed source set, and the majority produce different answers across routes.

\begin{table}[htbp]
\vspace{-5mm}
\centering
\small
\caption{Primary exact-token route diagnostics. Each row uses 63 query pairs from 29 equally weighted users with \texttt{one} padding; the \texttt{x} control and full direct profiles are in Appendix~I.3. Each route combines one formal and two fresh answers. Lower and Upper are 95\% user-bootstrap limits.}
\label{tab:main-route}
\setlength{\tabcolsep}{2pt}
\begin{tabular*}{\linewidth}{@{\extracolsep{\fill}}lrrrrrrr@{}}
\toprule
 & & \multicolumn{3}{c}{$D_{\mathrm{res}}$} & \multicolumn{3}{c}{$\Delta_{\mathrm{rep}}$}\\
\cmidrule(lr){3-5}\cmidrule(lr){6-8}
\textbf{Policy} & \textbf{$J$} & \textbf{Estimate} & \textbf{Lower} & \textbf{Upper} & \textbf{Estimate} & \textbf{Lower} & \textbf{Upper}\\
\midrule
Focal & 0.057 & 0.529 & 0.361 & 0.690 & 0.444 & 0.261 & 0.612\\
Recent-3 & 0.166 & 0.493 & 0.376 & 0.604 & 0.028 & -0.183 & 0.240\\
\bottomrule
\end{tabular*}
\vspace{-2mm}
\end{table}

Source overlap alone does not establish task relevance: sources can change without affecting the answer, and the same correct answer can emerge from disjoint evidence sets (Appendix~\ref{app:answer}). When we split queries by whether the \emph{supporting} record changed, disagreement and correctness gains are both larger in the support-changed group (Table~\ref{tab:support-localization} in Appendix~\ref{app:additional-support-localization}).

\subsection{Does the Displaced Evidence Matter?}

We now apply the four-condition support-restoration test from Section~\ref{sec:support-answer}. We analyze pairs where exactly one supporting record is displaced, all four conditions can be matched at the same source count, and prompt length is equalized. Pairs that swap supporting identities and pairs that fail these structural requirements are analyzed separately (Appendix~\ref{app:additional-route-rescue}).
Restoring the displaced record recovers task performance (focal R--A = 0.671), while the matched non-support replacement offers little recovery despite equal source count and prompt length (Table~\ref{tab:restoration-main}). This holds in both the primary stratum and the broader ``All eligible'' set covering every annotated query that passes the four-condition rule.

\begin{figure}[htbp]
\centering
\includegraphics[width=\linewidth]{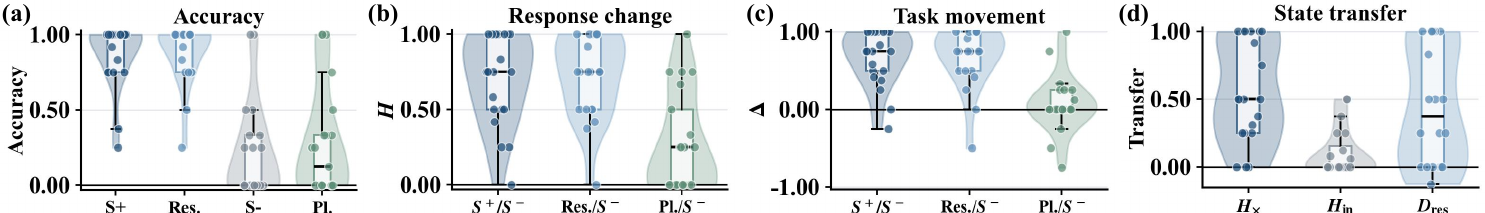}
\vspace{-6mm}
\caption{Support restoration (a--c) and state reuse (d). $S^+$, Res., $S^-$, and Pl. denote support present, rescue, support absent, and matched non-support replacement. Contrasts compare against $S^-$. Panel (d) uses repeated answers, with $H_{\mathrm{in}}=H_{\mathrm{within}}$. The full route profiles are in Appendix~\ref{app:answer-exact-token}.}
\label{fig:support-restoration}
\vspace{-5mm}
\end{figure}

\noindent\textbf{Across answer models.} We replicated these fixed-input experiments with all five answer models. Every focal pair changed its exposed source set, and the restoration-minus-absent contrast was positive for all five models. Qwen3-8B shows the smallest task-level effects, with its rescue-minus-placebo interval including zero; the other four models show robust recovery. Full estimates are in Appendix~\ref{app:answer-model-replication}. The consistency across models reinforces the interpretation that the effect is driven by the evidence change itself, not by an idiosyncratic sensitivity of any single answer model.

The evidence now shows that construction order changes sources \emph{and} that the displaced sources matter. The gap between restoration and replacement is large---over 60 percentage points under both focal and bounded-recent policies---and it holds across the broader ``All eligible'' stratum, ruling out the possibility that the primary consensus set is unrepresentatively sensitive. The natural next question is: which part of the construction process is responsible?
\begin{table}[htbp]
\vspace{-5mm}
\centering
\small
\caption{PersonaMem whole-record restoration. Primary is the semantic-consensus set; All eligible contains every fact-recall pair passing the structural gate. Pres., Abs., Res., and Repl. denote support present, support absent, restored support, and matched non-support replacement. In the contrast panel, R--A is restored minus absent and R--P is restored minus replacement. LongMemEval-S is reported in Section~\ref{sec:longer-histories}. Estimates and 95\% bootstrap limits weight users equally; full intervals appear in Tables~\ref{tab:support-rescue}, \ref{tab:recent-support-restoration}, \ref{tab:personamem-restoration-expanded}, and~\ref{tab:longmem-answers}.}
\label{tab:restoration-main}
\setlength{\tabcolsep}{1pt}
\begin{tabular*}{\linewidth}{@{\extracolsep{\fill}}lrrrrrrrrrrrr@{}}
\toprule
 & \multicolumn{6}{c}{Primary} & \multicolumn{6}{c}{All eligible}\\
\cmidrule(lr){2-7}\cmidrule(lr){8-13}
\textbf{Policy} & \textbf{Q} & \textbf{U} & \textbf{Pres.} & \textbf{Abs.} & \textbf{Res.} & \textbf{Repl.} & \textbf{Q} & \textbf{U} & \textbf{Pres.} & \textbf{Abs.} & \textbf{Res.} & \textbf{Repl.}\\
\midrule
Focal & 33 & 21 & 0.875 & 0.210 & 0.881 & 0.280 & 78 & 33 & 0.894 & 0.191 & 0.879 & 0.239\\
Recent-3 & 36 & 18 & 0.940 & 0.269 & 0.917 & 0.167 & 71 & 28 & 0.868 & 0.225 & 0.927 & 0.109\\
\bottomrule
\end{tabular*}
\par\smallskip
\begin{tabular*}{\linewidth}{@{\extracolsep{\fill}}lrrrrrrrrrrrr@{}}
\toprule
 & \multicolumn{3}{c}{Primary R$-$A} & \multicolumn{3}{c}{Primary R$-$P} & \multicolumn{3}{c}{All eligible R$-$A} & \multicolumn{3}{c}{All eligible R$-$P}\\
\cmidrule(lr){2-4}\cmidrule(lr){5-7}\cmidrule(lr){8-10}\cmidrule(lr){11-13}
\textbf{Policy} & \textbf{Estimate} & \textbf{Lower} & \textbf{Upper} & \textbf{Estimate} & \textbf{Lower} & \textbf{Upper} & \textbf{Estimate} & \textbf{Lower} & \textbf{Upper} & \textbf{Estimate} & \textbf{Lower} & \textbf{Upper}\\
\midrule
Focal & 0.671 & 0.496 & 0.817 & 0.601 & 0.442 & 0.756 & 0.688 & 0.576 & 0.790 & 0.640 & 0.487 & 0.771\\
Recent-3 & 0.648 & 0.519 & 0.773 & 0.750 & 0.639 & 0.856 & 0.702 & 0.607 & 0.789 & 0.818 & 0.743 & 0.889\\
\bottomrule
\end{tabular*}
\vspace{-2mm}
\end{table}

\subsection{Which Construction Step Causes the Change?}

The focal compactor has two order-dependent operations: \emph{similarity clustering} (which records are grouped together) and \emph{survivor selection} (which record in each cluster is kept). We isolate their contributions by varying each operation independently under three reordering schedules---reversal, half-swap, and odd-even---while keeping the source pool and answer interface fixed.

\begin{table}[htbp]
\vspace{-4mm}
\centering
\small
\caption{Mechanism localization by schedule using single-draw $H$ and $\Delta$. Estimates weight users equally; CIs are in Table~\ref{tab:semantic-mechanism-direct}. Changed counts pairs with different exposed source sets; the denominator is 63 in every row.}
\label{tab:mechanism-localization}
\setlength{\tabcolsep}{2pt}
\begin{tabular*}{\linewidth}{@{\extracolsep{\fill}}lrrrrrrrrr@{}}
\toprule
 & \multicolumn{3}{c}{Replay} & \multicolumn{3}{c}{Half-swap} & \multicolumn{3}{c}{Odd-even}\\
\cmidrule(lr){2-4}\cmidrule(lr){5-7}\cmidrule(lr){8-10}
\textbf{Operation} & \textbf{$H$} & \textbf{$\Delta$} & Changed & \textbf{$H$} & \textbf{$\Delta$} & Changed & \textbf{$H$} & \textbf{$\Delta$} & Changed\\
\midrule
Both operations & 0.603 & 0.390 & 63 & 0.350 & 0.011 & 63 & 0.425 & 0.182 & 43\\
Survivor selection & 0.558 & 0.347 & 63 & 0.367 & 0.028 & 63 & 0.420 & 0.202 & 41\\
Clustering only & 0.103 & 0.000 & 4 & 0.135 & -0.003 & 2 & 0.115 & 0.080 & 2\\
\bottomrule
\end{tabular*}
\vspace{-2mm}
\end{table}

Survivor selection accounts for most source changes (Table~\ref{tab:mechanism-localization}). Across all three schedules, varying which record each cluster keeps produces nearly as many source changes as varying both operations together. Changing clustering alone---while keeping the survivor rule fixed to source order---rarely affects the exposed set ($\leq$4 of 63 queries changed under any schedule).

Reversal helps this compactor because supporting records tend to appear early in the benchmark pools (median normalized position 0.25), while last-arrival retention discards them. Reversal turns early records into late arrivals, protecting them. First-arrival retention reverses the asymmetry; the pattern is stable across clustering thresholds (Appendix~\ref{app:order-structure}). Exhaustive enumeration over all $n_q!$ permutations of each pool confirms that $\sim$80\% of permutations change the survivor set under last-arrival retention, whereas fixing both clustering and selection to source order produces a permutation-invariant result (Table~\ref{tab:order-structure} in Appendix~\ref{app:order-structure}).

\noindent\textbf{State reuse across questions.} When we reuse each constructed state for other questions from the same user, the correctness advantage concentrates in questions whose exposed support differs between routes; questions lacking support on both paths show uncertain mean changes (Figure~\ref{fig:support-restoration}d; Table~\ref{tab:state-transfer-support}).

\subsection{Do Other Memory Policies Show the Same Pattern?}

The focal compactor is a controlled policy designed to isolate a failure mode. Real systems use different architectures---summarization, bounded windows, metadata graphs. We apply RoutePrism to four additional interfaces to localize where construction order changes behavior.

\noindent\textbf{MemoChat configuration.} MemoChat summarizes after six chunks, keeps two recent chunks and up to three topics, and retrieves up to four topic summaries plus two recent dialogs (each summary retains two dialogs and at most 320 characters; deterministic hashing, no learned topic segmentation).

\begin{table}[htbp]
\centering
\small
\sbox{0}{\begin{minipage}{0.62\linewidth}
\setlength{\tabcolsep}{1pt}
\begin{tabular*}{\linewidth}{@{\extracolsep{\fill}}lrrrrr@{}}
\toprule
\textbf{Policy} & \textbf{Queries} & \textbf{Users} & \textbf{$V$} & \textbf{$C$} & \textbf{Internal}\\
\midrule
Lexical reference & 589 & 37 & 0 & 0 & 589\\
Cross-fitted streaming & 589 & 37 & 40 & 40 & 51\\
MemoChat & 63 & 29 & 1 & 63 & 63\\
A-MEM & 63 & 29 & 0 & 0 & 63\\
\bottomrule
\end{tabular*}
\end{minipage}}
\makeatletter
\long\def\@makecaption#1#2{%
  \vbox to \dimexpr\ht0+\dp0\relax{%
    \baselineskip=10pt plus 1fill\relax
    \noindent #1: #2\par\vskip0pt}}
\makeatother
\sbox{2}{\begin{minipage}{0.37\linewidth}
\caption{Changed pairs at source ($V$), context ($C$), and internal-state ($I$) interfaces. Queries and Users give the analysis units; entries under $V$, $C$, and $I$ count changed pairs. A-MEM reuses 29 state pairs.}
\label{tab:policy-interfaces-main}
\end{minipage}}
\raisebox{\dp0}{\usebox{0}}\hfill
\raisebox{\dp2}{\usebox{2}}
\end{table}

\noindent\textbf{Cross-fitted streaming.} A held-out learned-retention arm uses a ridge score from term frequency--inverse document frequency (TF--IDF) features to make capacity-three online choices without revisiting discarded records. Its matched support-identity intervention is reported in Appendix~\ref{app:answer}.

\noindent\textbf{MemoChat display intervention.} The display experiment crosses two retained construction states with native or reversed block order within summary, retrieved-dialog, and recent-dialog groups. Block text and within-block order stay fixed, and two fresh draws in each cell use equal-length prompts. The construction contrast (M-R) averages layouts; the display contrast (M-D) averages states.

The signed contrasts use replay minus forward for construction and reversed minus native for display.
\begin{table}[htbp]
\centering
\small
\sbox{0}{\begin{minipage}{0.65\linewidth}
\setlength{\tabcolsep}{1pt}
\begin{tabular*}{\linewidth}{@{\extracolsep{\fill}}lrrrrrr@{}}
\toprule
 & \multicolumn{3}{c}{$D_{\mathrm{res}}$} & \multicolumn{3}{c}{$\Delta_{\mathrm{rep}}$}\\
\cmidrule(lr){2-4}\cmidrule(lr){5-7}
\textbf{Trace} & \textbf{Estimate} & \textbf{Lower} & \textbf{Upper} & \textbf{Estimate} & \textbf{Lower} & \textbf{Upper}\\
\midrule
M-R & 0.059 & 0.016 & 0.112 & 0.071 & -0.002 & 0.157\\
M-D & 0.013 & -0.010 & 0.040 & 0.042 & 0.004 & 0.080\\
\bottomrule
\end{tabular*}
\end{minipage}}
\makeatletter
\long\def\@makecaption#1#2{%
  \vbox to \dimexpr\ht0+\dp0\relax{%
    \baselineskip=10pt plus 1fill\relax
    \noindent #1: #2\par\vskip0pt}}
\makeatother
\sbox{2}{\begin{minipage}{0.32\linewidth}
\caption{MemoChat contrasts from the same 504 answers: 63 queries, 29 users. Lower and Upper give 95\% user-bootstrap limits.}
\label{tab:policy-context-effects}
\end{minipage}}
\raisebox{\dp0}{\usebox{0}}\hfill
\raisebox{\dp2}{\usebox{2}}
\end{table}

\noindent\textbf{Distinct interface patterns.} Table~\ref{tab:policy-interfaces-main} shows that the four policies produce order effects at different layers. The lexical reference changes nothing visible (internal index order shifts, but exposed sources and contexts stay identical). The cross-fitted streaming policy changes sources and contexts for a minority of queries (40 of 589). MemoChat changes the compiled context for \emph{every} query while barely changing the source identifiers---its construction residual exceeds the display-only residual (Appendix~\ref{app:external-memochat}). A-MEM changes internal metadata (tags and note descriptions) while the answer prompt---which reads note bodies via dense retrieval---stays identical. No single observation layer catches all four failure modes.

These results illustrate why RoutePrism records sources, contexts, and internal state separately: a diagnostic that checks only source identity would miss MemoChat's context changes entirely, while one that checks only the answer prompt would miss A-MEM's metadata drift. The MemoChat display experiment further separates the two sources of variation in summarization-based policies: the construction contrast ($D_{\mathrm{res}}=0.059$) exceeds the display-only contrast ($D_{\mathrm{res}}=0.013$), indicating that how MemoChat \emph{constructs} its summaries matters more than how it \emph{arranges} them in the answer prompt (Table~\ref{tab:policy-context-effects}). This distinction would be invisible to any evaluation that treats the compiled context as a monolithic input.

\subsection{Does the Pattern Hold on Longer Histories?}
\label{sec:longer-histories}

To test whether construction-order effects persist on longer histories, we use the cleaned LongMemEval-S release~\citep{wu2025longmemeval}: 470 questions with native histories of 38--62 sessions, where the benchmark provides evidence-session labels rather than model-generated support annotations. PersonaMem users have 4--7 records each. We reverse import order while preserving timestamps and within-session content, and apply BM25 ($k_1=1.5$, $b=0.75$) retrieval to expose the top three retained sessions.

\begin{table}[htbp]
\vspace{-4mm}
\centering\small
\caption{LongMemEval-S retention and exposure over 470 questions. Overlap is source-set Jaccard; recall is the fraction of published evidence sessions. F/R denote import order; intervals and source-order controls are in Appendix~\ref{app:longmem-scale}.}
\label{tab:longmem-scale-main}
\setlength{\tabcolsep}{3pt}
\begin{tabular*}{\linewidth}{@{\extracolsep{\fill}}lrrrrrr@{}}
\toprule
 & \multicolumn{3}{c}{Retained sessions} & \multicolumn{3}{c}{Exposed sessions}\\
\cmidrule(lr){2-4}\cmidrule(lr){5-7}
\textbf{Policy} & \textbf{Overlap} & \textbf{Recall F} & \textbf{Recall R} & \textbf{Overlap} & \textbf{Recall F} & \textbf{Recall R}\\
\midrule
Recent-8 & 0.000 & 0.215 & 0.146 & 0.000 & 0.209 & 0.140\\
Recent-16 & 0.000 & 0.381 & 0.319 & 0.000 & 0.350 & 0.295\\
Recent-32 & 0.349 & 0.688 & 0.619 & 0.287 & 0.598 & 0.539\\
Focal & 0.438 & 0.157 & 0.113 & 0.227 & 0.146 & 0.106\\
\bottomrule
\end{tabular*}
\vspace{-2mm}
\end{table}

Larger budgets retain more evidence, while retrieval continues to leave coverage gaps (Table~\ref{tab:longmem-scale-main}). Recent-8 and Recent-16 retain completely disjoint session sets across orders (overlap = 0.000), so every piece of evidence available to the agent changes when import order is reversed. Even Recent-32 only partially overlaps (Jaccard = 0.349).

\noindent\textbf{Answer accuracy by task type.} Forward import produces higher answer accuracy for both Luna and Qwen3-8B across nearly all task categories. Under Recent-8 with BM25, Luna scores 0.206 forward versus 0.081 reverse overall ($\Delta=-0.126$, 95\% CI $[-0.170,-0.083]$); Qwen3-8B shows a comparable gap (0.204 vs.\ 0.089). The largest asymmetry appears in knowledge-update questions ($\Delta=-0.389$), where gold answers align with more recent sessions that forward import retains. Preference questions also show a substantial drop ($\Delta=-0.200$), while multi-session questions---which require evidence from multiple sessions---show uniformly low accuracy under both orders (forward 0.058, reverse 0.017), reflecting a fundamental coverage limitation at budget $k=3$ (Appendix~\ref{app:longmem-coverage}).

\noindent\textbf{Restoration on longer histories.} The support-restoration test also transfers. Among the 24 eligible support-change pairs from single-session-user questions, Luna achieves 1.000 accuracy with evidence present and 0.000 when the evidence session is absent; restoration fully recovers performance (1.000), while the matched replacement does not (0.042). Qwen3-8B shows the same pattern with slightly lower ceilings (Table~\ref{tab:longmem-answers-main}). The gap between restored and replacement conditions ($>$0.87 for both models) exceeds the PersonaMem estimates, likely because the longer histories make the displaced session the sole source of relevant evidence.

\begin{table}[htbp]
\vspace{-3mm}
\centering\small
\caption{LongMemEval-S four-condition answer accuracy on 24 single-session-user questions. Both answer models are scored by the same masked Terra max judge. Lower and Upper are 95\% question-bootstrap limits.}
\label{tab:longmem-answers-main}
\setlength{\tabcolsep}{2pt}
\begin{tabular*}{\linewidth}{@{\extracolsep{\fill}}lrrrrrr@{}}
\toprule
\textbf{Condition} & \multicolumn{3}{c}{\textbf{Luna}} & \multicolumn{3}{c}{\textbf{Qwen3-8B}}\\
\cmidrule(lr){2-4}\cmidrule(lr){5-7}
 & \textbf{Estimate} & \textbf{Lower} & \textbf{Upper} & \textbf{Estimate} & \textbf{Lower} & \textbf{Upper}\\
\midrule
Present & 1.000 & 1.000 & 1.000 & 0.875 & 0.750 & 1.000\\
Absent & 0.000 & 0.000 & 0.000 & 0.000 & 0.000 & 0.000\\
Restored & 1.000 & 1.000 & 1.000 & 0.917 & 0.792 & 1.000\\
Replacement & 0.042 & 0.000 & 0.125 & 0.042 & 0.000 & 0.125\\
\bottomrule
\end{tabular*}
\vspace{-2mm}
\end{table}

\subsection{Robustness Checks}
\label{sec:semantic-controls}

Robustness checks cover the screening rule, answer model, and benchmark construction.

\noindent\textbf{Screening-rule sensitivity.} Replacing the Luna-based semantic screen with independent Terra or Sol readings, or using the benchmark's own taxonomy as the selection rule, preserves positive residuals and signed contrasts across all tested gates. The benchmark taxonomy retains all 146 fact-recall queries and gives $D_{\mathrm{res}}=0.536$ [$0.442, 0.628$]; the independent two-read screen on the same 146 cards gives $D_{\mathrm{res}}=0.626$ [$0.491, 0.756$] on 59 queries. A Terra-majority gate (70 queries, 31 users) gives $D_{\mathrm{res}}=0.469$; a Sol-majority gate (69 queries, 32 users) gives $D_{\mathrm{res}}=0.463$. Tightening to unanimity across all three readers shrinks the set but keeps the residual positive (Appendix~\ref{app:semantic}).

\noindent\textbf{Support annotation agreement.} Two independent support annotation passes agree on 125 of 146 direct-support sets ($\kappa=0.842$). Leave-one-user-out checks keep the rescue-minus-absent contrast between 0.654 and 0.729, confirming that no single user drives the result.

\noindent\textbf{Cross-model replication.} The core fixed-input matrices test the same source-level intervention across five answer models (Table~\ref{tab:model-replication-main}). Every focal pair changes its exposed source set under all five models. The four proprietary models show robust restoration effects, with focal R--A between 0.671 and 0.808. Qwen3-8B shows the smallest task-level effects ($R$--$A=0.214$, CI includes zero), consistent with lower context exploitation rather than a different source-level mechanism. The bounded-recent restoration pattern is similarly consistent, with $R$--$A$ between 0.648 and 0.685 for the proprietary models.

\begin{table}[htbp]
\vspace{-3mm}
\centering
\small
\caption{Core cross-model replication on fixed PersonaMem inputs. $H$ and $\Delta$ use the focal 63-pair route trace; $R$--$A$ (rescue minus absent) and $R$--$P$ (rescue minus replacement) use the support restoration sets. All entries are equal-user point estimates; 95\% intervals are in Appendix~\ref{app:answer-model-replication}.}
\label{tab:model-replication-main}
\setlength{\tabcolsep}{2pt}
\begin{tabular*}{\linewidth}{@{\extracolsep{\fill}}lrrrrrr@{}}
\toprule
\textbf{Model} & \textbf{$H_{\mathrm{focal}}$} & \textbf{$\Delta_{\mathrm{focal}}$} & \textbf{Focal $R$--$A$} & \textbf{Focal $R$--$P$} & \textbf{Recent $R$--$A$} & \textbf{Recent $R$--$P$}\\
\midrule
Astra & 0.581 & 0.442 & 0.808 & 0.756 & 0.671 & 0.819\\
Sol & 0.559 & 0.399 & 0.764 & 0.683 & 0.685 & 0.759\\
Terra & 0.558 & 0.376 & 0.774 & 0.681 & 0.681 & 0.810\\
Luna & 0.614 & 0.437 & 0.671 & 0.601 & 0.648 & 0.750\\
Qwen3-8B & 0.338 & 0.175 & 0.214 & 0.143 & 0.194 & 0.167\\
\bottomrule
\end{tabular*}
\vspace{-2mm}
\end{table}

\noindent\textbf{Benchmark artifacts.} A post hoc check finds that 2 of the 63 primary queries have duplicate answer-option text; excluding them leaves all focal estimates positive ($D_{\mathrm{res}}=0.524$, $\Delta_{\mathrm{rep}}=0.439$ on 61 pairs; Appendix~\ref{app:option-sensitivity}).

\FloatBarrier

\section{Discussion and Conclusion}

\noindent\textbf{Accuracy is not enough---audit the evidence.} Endpoint accuracy conflates source selection, context compilation, and answer sampling into a single number. RoutePrism's two-path trace separates these stages: $J$ catches source replacement, context comparison catches rewriting, and $D_{\mathrm{res}}$ removes answer-model noise. When paired runs retain identical sources and contexts ($J=1$, as with the lexical reference), canonical ingestion suffices; when they do not ($J=0.057$, as with the focal compactor), a second path and repeated answers reveal whether the difference exceeds sampling noise.

\noindent\textbf{Multi-layer observation.} Order effects manifest at entirely different layers across policies (Table~\ref{tab:policy-interfaces-main}): the focal compactor changes sources, MemoChat changes compiled contexts while preserving source identities, and A-MEM changes internal metadata without affecting the answer prompt. No single observation layer catches all failure modes.

\noindent\textbf{Practical design implications.} Three concrete actions follow: (i) compare survivor rules against task support before deployment; (ii) retain record-level provenance through merging and summarization; (iii) log which fields retrieval and answering actually consume. The paired-path methodology can also serve as a regression test for new order sensitivity.

\noindent\textbf{Answer-model dependence.} The source-level intervention transfers across all five models, but its task-level magnitude varies (Table~\ref{tab:model-replication-main}): Qwen3-8B shows smaller accuracy effects, likely reflecting differences in context exploitation. Route audits should report the answer model alongside the memory policy.

\noindent\textbf{Limitations.} The focal compactor is a controlled policy used to isolate one failure mode; production deployment lies outside the evaluated setting. PersonaMem histories are short (4--7 records), and LongMemEval-S restoration covers only single-record support-loss cases. Generalizing to production systems with learned policies and continuously growing histories remains future work, as does studying stochastic order variation from concurrent processing.

\bibliographystyle{iclr2027_conference}
\bibliography{references}

\appendix

\clearpage
\section*{Appendix Directory}
\begingroup
\small
\setlength{\parindent}{0pt}
\newcommand{\appendixentry}[3]{%
  \noindent\textbf{#1\quad\hyperref[#2]{#3}}\nobreak\dotfill\nobreak\pageref{#2}\par\vspace{2pt}}
\newcommand{\appendixsubentry}[3]{%
  \noindent\hspace*{1.8em}#1\quad\hyperref[#2]{#3}\nobreak\dotfill\nobreak\pageref{#2}\par}
\appendixentry{A}{app:setting}{Evaluation setting}
\appendixsubentry{A.1}{app:setting-benchmark}{Benchmark and pairing}
\appendixsubentry{A.2}{app:setting-transformation}{Declared transformation}
\appendixsubentry{A.3}{app:setting-answer-boundary}{Answer boundary and extensions}
\appendixsubentry{A.4}{app:setting-formal-interface}{Formal paired interface}
\appendixsubentry{A.5}{app:setting-estimands}{Estimands and answer-stage identities}
\appendixentry{B}{app:semantic}{Semantic audit}
\appendixsubentry{B.1}{app:semantic-partition}{Semantic partition and admissibility gate}
\appendixsubentry{B.2}{app:semantic-pool-audit}{Sensitivity audit over complete pools}
\appendixsubentry{B.3}{app:semantic-disagreements}{Screening disagreements}
\appendixsubentry{B.4}{app:semantic-independent-readings}{Independent semantic readings}
\appendixsubentry{B.5}{app:semantic-context-sensitivity}{Context and model sensitivity}
\appendixsubentry{B.6}{app:semantic-taxonomy-sensitivity}{Benchmark taxonomy sensitivity}
\appendixsubentry{B.7}{app:semantic-crossfit-restoration}{Cross-fitted support restoration}
\appendixentry{C}{app:policy}{Policy and transformation details}
\appendixsubentry{C.1}{app:policy-principal}{Principal policies}
\appendixsubentry{C.2}{app:policy-timing}{Policy timing}
\appendixsubentry{C.3}{app:policy-recent}{Bounded recent policy}
\appendixsubentry{C.4}{app:policy-mechanism}{Mechanism schedules}
\appendixsubentry{C.5}{app:policy-secondary}{Secondary policies and controls}
\appendixsubentry{C.6}{app:option-sensitivity}{Identical-option sensitivity}
\appendixentry{D}{app:construction}{Construction details}
\appendixsubentry{D.1}{app:construction-representation}{Record representation}
\appendixsubentry{D.2}{app:construction-state}{State objects}
\appendixsubentry{D.3}{app:construction-clustering}{Similarity clustering}
\appendixsubentry{D.4}{app:construction-focal-rule}{Executable focal rule}
\appendixsubentry{D.5}{app:construction-tie-audit}{Tie audit}
\appendixsubentry{D.6}{app:construction-retrieval}{Retrieval and budget boundary}
\appendixsubentry{D.7}{app:construction-schedules}{Operation-level schedules}
\appendixsubentry{D.8}{app:construction-crossfit}{Cross-fitted streaming policy}
\appendixsubentry{D.9}{app:construction-state-reuse}{State reuse across questions}
\appendixsubentry{D.10}{app:construction-persistent-continuation}{Persistent continuation}
\appendixsubentry{D.11}{app:order-structure}{Permutation and retention structure}
\appendixentry{E}{app:uncertainty}{Uncertainty and complete estimates}
\appendixsubentry{E.1}{app:uncertainty-bootstrap}{Bootstrap unit}
\appendixsubentry{E.2}{app:uncertainty-answer-stage}{Answer-stage diagnostic}
\appendixsubentry{E.3}{app:uncertainty-estimates}{Complete policy and provenance estimates}
\appendixsubentry{E.4}{app:uncertainty-contrasts}{Paired contrasts and answer-stage diagnostics}
\appendixentry{F}{app:additional}{Additional conditions}
\appendixsubentry{F.1}{app:additional-strata}{Semantic-stratum estimates}
\appendixsubentry{F.2}{app:additional-support-localization}{Answer-bearing support localization}
\appendixsubentry{F.3}{app:additional-route-rescue}{Route-support rescue}
\appendixsubentry{F.4}{app:additional-selection-breadth}{Breadth of selection rules}
\appendixsubentry{F.5}{app:additional-operation-dispersion}{Operation-level dispersion}
\appendixsubentry{F.6}{app:additional-selection-controls}{Selection and instruction controls}
\appendixentry{G}{app:related}{Extended related work}
\appendixsubentry{G.1}{app:related-long-horizon}{Long-horizon memory evaluation}
\appendixsubentry{G.2}{app:related-order-construction}{Order and construction effects}
\appendixsubentry{G.3}{app:related-structure-provenance}{Memory structure, updates, and provenance}
\appendixentry{H}{app:scope}{Comparison settings}
\appendixsubentry{H.1}{app:scope-adjacent-boundaries}{Adjacent evaluation boundaries}
\appendixsubentry{H.2}{app:scope-display-intervention}{Fixed-content display intervention (M-R and M-D)}
\appendixentry{I}{app:answer}{Answer-boundary controls}
\appendixsubentry{I.1}{app:answer-support-prompts}{Support-present and restored prompts}
\appendixsubentry{I.2}{app:answer-fresh-sensitivity}{Fresh-answer sensitivity}
\appendixsubentry{I.3}{app:answer-exact-token}{Exact-token focal trace (F-E)}
\appendixsubentry{I.4}{app:answer-endpoint-support}{Endpoint versus support}
\appendixsubentry{I.5}{app:answer-submission-balance}{Submission-order balance}
\appendixsubentry{I.6}{app:answer-policy-matrix}{Policy matrix for held-out users}
\appendixsubentry{I.7}{app:answer-support-audit}{Support-layer audit}
\appendixsubentry{I.8}{app:answer-canonical-serialization}{Canonical serialization control}
\appendixsubentry{I.9}{app:answer-focal-repeat}{Focal canonicalized repeat (F-C)}
\appendixsubentry{I.10}{app:answer-tokenizer-controls}{Declared tokenizer length controls}
\appendixsubentry{I.11}{app:answer-recent-aligned}{Recent-3: equal tokens and aligned offsets (R-E)}
\appendixsubentry{I.12}{app:answer-model-replication}{Core cross-model replication}
\appendixsubentry{I.13}{app:answer-crossfit-restoration}{Cross-fitted support restoration}
\appendixsubentry{I.14}{app:recent-support}{Support restoration under bounded recent retention}
\appendixsubentry{I.15}{app:personamem-restoration-expanded}{Expanded PersonaMem restoration coverage}
\appendixentry{J}{app:external-bm25}{Source-stable lexical control}
\appendixentry{K}{app:external-memochat}{MemoChat-style configuration}
\appendixsubentry{K.1}{app:memochat-policy-pairing}{Policy and pairing}
\appendixsubentry{K.2}{app:memochat-internal-state}{Internal state}
\appendixsubentry{K.3}{app:memochat-answer-boundary}{Answer boundary}
\appendixsubentry{K.4}{app:memochat-two-samples}{Two answer samples}
\appendixsubentry{K.5}{app:memochat-repeated-answers}{Repeated answers}
\appendixsubentry{K.6}{app:memochat-fixed-source}{Fixed-source sensitivity}
\appendixsubentry{K.7}{app:memochat-correctness}{Correctness projection}
\appendixentry{L}{app:external-amem}{Independent A-MEM policy}
\appendixsubentry{L.1}{app:amem-policy-pairing}{Policy and pairing}
\appendixsubentry{L.2}{app:amem-state-signature}{State signature}
\appendixsubentry{L.3}{app:amem-boundary-result}{Boundary result}
\appendixsubentry{L.4}{app:amem-retrieval-boundary}{Retrieval boundary}
\appendixentry{M}{app:measurements}{Measurement conventions and estimands}
\appendixsubentry{M.1}{app:measurements-index}{Experiment index}
\appendixsubentry{M.2}{app:measurements-metadata}{Answer-call metadata}
\appendixsubentry{M.3}{app:measurements-readouts}{Information lost by simpler readouts}
\appendixentry{N}{app:checks}{Admissibility and state observability}
\appendixsubentry{N.1}{app:checks-compiled-support}{Support after compilation}
\appendixsubentry{N.2}{app:checks-pool-screening}{Complete-pool screening}
\appendixsubentry{N.3}{app:checks-gold-reread}{Gold-visible re-reading and intersection analysis}
\appendixsubentry{N.4}{app:checks-crossfit}{Cross-fitted extension}
\appendixentry{O}{app:longmem-scale}{LongMemEval-S extension}
\appendixsubentry{O.1}{app:longmem-data}{Data and unit of analysis}
\appendixsubentry{O.2}{app:longmem-construction}{Construction and retrieval}
\appendixsubentry{O.3}{app:longmem-common-interface}{Common-interface comparison}
\appendixsubentry{O.4}{app:longmem-measurements}{Measurements}
\appendixsubentry{O.5}{app:longmem-restoration}{Matched answers on native histories}
\appendixsubentry{O.6}{app:longmem-coverage}{Answer coverage across task types}
\endgroup
\clearpage

\section{Evaluation Setting}
\label{app:setting}

\subsection{Benchmark and pairing}
\label{app:setting-benchmark}
We use the released PersonaMem-32K evaluation split of simulated user--language-model conversations, which contains 589 queries with four answer options, 195 session documents, and 37 user contexts. Each user has a fixed evidence pool of four to seven sessions, shared across that user's questions. The source identifiers define the forward and replay orders; the 589-query trace preserves one pool identifier per user across all questions and policies. Correctness is evaluated against the released answer keys. The original benchmark study reports human validation of sampled query--response pairs~\citep{personamem2025}.

\subsection{Declared transformation}
\label{app:setting-transformation}
For each pair, the forward order sorts the numeric suffix of the session identifier in increasing order and the replay order reverses the same list. The transformation uses none of the question, option, label, category, or evaluator-output fields. Memory is rebuilt from the controlled pool for each query, and the answer prompt constrains the model to one lowercase option label.

\subsection{Answer boundary and extensions}
\label{app:setting-answer-boundary}
The principal answer matrices use Luna max, with the fixed prompt and context budget used in the main comparison. One categorical answer is collected for each cell in the primary matrix. The cross-user extension evaluates the held-out users under four route schedules. The answer-repeat diagnostic collects two independent draws for each route and policy cell in the paired repeat matrix. The exact-token control equalizes the full prompt token length using the declared \texttt{o200k\_base} tokenizer. A separate balanced control gives both route realizations a common neutral-padding budget in a single draw. The aligned-offset extension adds a common prefix and neutral padding after the context so that the evidence and question start positions also coincide. Lexical weighting, bounded windows, utility buffers, and mechanism conditions use the same benchmark split and answer boundary. A fixed-input replication of the exact-token primary and support-rescue arms uses Sol max and is reported separately.

As a descriptive proxy for context length, we use $\lceil n/4\rceil$, where $n$ is the number of characters after serialization. Exact token counts are measured separately.

\subsection{Formal paired interface}
\label{app:setting-formal-interface}
For question $q$ with source pool $B_q$ of $n_q$ records, the forward and replay arrival orders are $\pi_f(i)=i$ and $\pi_r(i)=n_q-1-i$. Construction receives only $B_q$ and $\pi_o$; record text, dates, and within-record relations are unchanged. The route map is:
\begin{equation}
M_q^{(o)}=\operatorname{Build}(B_q;\pi_o),\quad (V_q^{(o)},C_q^{(o)})=\operatorname{Expose}(M_q^{(o)},q),\quad a_q^{(o)}\sim A(\cdot\mid C_q^{(o)},q,\mathcal{A}_q).
\label{eq:route-model}
\end{equation}
Exposure outputs are recorded before answer generation. Admissibility requires $g_q(B_q,\pi_f)=g_q(B_q,\pi_r)=y_q$; the primary set is the intersection of two independent admissible readings.

\subsection{Estimands and answer-stage identities}
\label{app:setting-estimands}
For a query-level quantity $z_q$, let $\mathcal{Q}_u$ be the eligible questions for user $u$ on which $z_q$ is defined, and let $\mathcal{U}$ be the users with $|\mathcal{Q}_u|>0$. The equal-user estimator used throughout is
\begin{equation}
\overline{z}
=\frac{1}{|\mathcal{U}|}\sum\nolimits_{u\in\mathcal{U}}
\left(\frac{1}{|\mathcal{Q}_u|}\sum\nolimits_{q\in\mathcal{Q}_u}z_q\right).
\label{eq:user-aggregation}
\end{equation}
For $p_{qa}^{(o)}=\Pr(a_q^{(o)}=a)$ and independent draws from fixed route distributions, the residual in the main text has population value
\begin{equation}
\mathbb{E}[D_{q,\mathrm{res}}]=D_q^\star=\frac{1}{2}\sum\nolimits_{a\in\mathcal{A}_q}\bigl(p_{qa}^{(f)}-p_{qa}^{(r)}\bigr)^2.
\label{eq:answer-distance}
\end{equation}
$D_q^\star$ is half the squared Euclidean distance between the two answer distributions; finite-sample residuals may be negative. Repeated runs use $\Delta_{q,\mathrm{rep}}=\overline{c}_q^{(r)}-\overline{c}_q^{(f)}$ with bootstrap over users.

\section{Semantic Audit}
\label{app:semantic}

\subsection{Semantic partition and admissibility gate}
\label{app:semantic-partition}

A semantic partition is fixed before route aggregation. Exchangeable facts cover recall of user-shared facts and facts mentioned by the user; temporal updates cover preference evolution and the reasons for earlier updates; recommendation, ideation, and new-scenario questions form the residual group. The three groups contain 146, 238, and 205 queries, respectively, and their counts sum to the full 589-query split. The label rule uses the benchmark's question taxonomy and does not inspect answers, route contrasts, selected source identities, or uncertainty estimates. The exchangeable group uses the storage-permutation estimand. For the temporal group, Appendix~\ref{app:additional} reports a chronology-aware provenance comparison; raw reversal is retained as a chronology diagnostic.

\subsection{Sensitivity audit over complete pools}
\label{app:semantic-pool-audit}
An independent audit tests the semantic premise on the complete source pool used by each paired route; the benchmark taxonomy defines the resulting 146-query coverage stratum. Each audit card contains the question, its four answer options, and all source blocks in $B_q$, with source positions preserved; the gold label, route-selected sources, answers, correctness labels, and effect estimates are omitted. Among the 146 query cards, 18, 68, 54, and 6 contain four, five, six, and seven source blocks, respectively. Two independent readers evaluate every complete pool under the same three-category rule. A pool is labeled admissible when reordering the complete source pool cannot alter the intended answer, ambiguous when the card does not determine whether chronology is part of that answer, and inadmissible when the question or evidence requires order, evolution, recency, or an update relation. The complete label marginals, confusion matrix, and agreement statistic are reported in Tables~\ref{tab:record-semantic} and~\ref{tab:fullpool-semantic-confusion}. The intersection of the two admissible sets selects the primary semantic stratum from the original paired traces.

\subsection{Screening disagreements}
\label{app:semantic-disagreements}
The archived reasons expose differences in event reference, preference scope, and temporal dependence. For query \texttt{0034a537}, the first reading treats the coin-sorting question as recall of a particular experience, whereas the second interprets the word ``anymore'' as an evolving preference. For \texttt{0d2259ae}, the readings differ on whether leaving competitive cooking classes changes the broader interest in recipes. For \texttt{87cfa0aa}, both readings distinguish an earlier disappointing film club from a later satisfying one, but differ on whether the question resolves which club it recalls. These examples come from the stored reading rationales, with query identifiers abbreviated. The later target-visible rubric in Appendix~\ref{app:checks} explicitly distinguishes temporal relations already stated in the records from relations supplied by block order; it is analyzed as a separate sensitivity rule.

\subsection{Independent semantic readings}
\label{app:semantic-independent-readings}
The complete pool cards and rubric were fixed before either reading. Each reading covered the same 146 cards; within each reading, independent readers processed disjoint batches under the same three-category rule, while route outcomes, answer outputs, and effect estimates were withheld.

\subsection{Context and model sensitivity}
\label{app:semantic-context-sensitivity}
A third blind Luna max read (same cards/rubric, no route outputs or answers) marked 25 admissible, 2 ambiguous, 119 inadmissible; it agreed with reads A and B on 94 and 80 pools, respectively. Strict plurality retains 64 pools and agrees with the two-reading mask on 145/146 decisions. Route estimates on the plurality set: $(H_{\times},D_{\mathrm{res}},\Delta_{\mathrm{rep}})=(0.595,0.470,0.411)$; on the two-reading consensus: $(0.586,0.459,0.403)$; on the 24-query all-three intersection: $(0.521,0.431,0.410)$.

An independent Terra max reading marks 58 admissible, 4 ambiguous, 84 inadmissible. Strict-majority aggregation with the two Luna readings retains 70 queries (31 users): $(H_{\times},D_{\mathrm{res}},\Delta_{\mathrm{rep}})=(0.594,0.469,0.390)$; the all-three intersection (48 queries, 26 users): $(0.529,0.433,0.346)$.

A Sol max reading retains 58 admissible and 88 inadmissible (no ambiguous). Strict-majority set (69 queries, 32 users): $(0.599,0.463,0.386)$; all-three set (50 queries, 27 users): $(0.585,0.474,0.403)$. All intervals are 95\% user-bootstrap.

\subsection{Benchmark taxonomy sensitivity}
\label{app:semantic-taxonomy-sensitivity}
The benchmark taxonomy retains the two fact-recall labels (\texttt{recall\_user\_shared\_facts}, \texttt{recalling\_facts\_mentioned\_by\_the\_user}), fixed before route aggregation and independent of answers. On the 146-query set (37 users), every pair changes its selected state: $1{-}J=0.955\;[0.919,0.991]$, $(H_{\times},H_{\mathrm{within}},D_{\mathrm{res}},\Delta_{\mathrm{rep}})=(0.659,0.123,0.536,0.312)$.

A separate record-grounded screen (two independent passes, no route outputs or answers) retains 59 of 146 cards (29 users); all change their source state: $(H_{\times},H_{\mathrm{within}},D_{\mathrm{res}},\Delta_{\mathrm{rep}})=(0.708,0.082,0.626,0.307)$. All intervals are 95\% user-bootstrap; full estimates appear in Table~\ref{tab:semantic-record-grounded}.

\begin{table}[ht]
\centering
\small
\caption{Semantic sensitivity under independent selection rules. Panel A reports screen counts; panels B and C report route coordinates with 95\% bootstrap limits, using equal weight for users. Route coordinates use the retained answer traces. NA denotes not applicable.}
\label{tab:semantic-record-grounded}
\setlength{\tabcolsep}{2pt}
\textbf{A. Semantic screen}\par\vspace{2pt}
\begin{tabular*}{\linewidth}{@{\extracolsep{\fill}}lllrrrrr@{}}
\toprule
 &  &  &  &  & \multicolumn{3}{c}{Label counts}\\
\cmidrule(lr){6-8}
\textbf{Read} & \textbf{Rule} & \textbf{Model} & \textbf{Queries} & \textbf{Users} & \textbf{Admissible} & \textbf{Ambiguous} & \textbf{Inadmissible}\\
\midrule
Record pass A & Single & Independent & 146 & 37 & 66 & 59 & 21\\
Record pass B & Single & Independent & 146 & 37 & 115 & 23 & 8\\
Admissible intersection & Two reads & Independent & 59 & 29 & \multicolumn{1}{l}{NA} & \multicolumn{1}{l}{NA} & \multicolumn{1}{l}{NA}\\
Benchmark fact recall & Taxonomy & Benchmark & 146 & 37 & \multicolumn{1}{l}{NA} & \multicolumn{1}{l}{NA} & \multicolumn{1}{l}{NA}\\
\bottomrule
\end{tabular*}
\par\smallskip
\textbf{B. Response disagreement}\par\vspace{2pt}
\begin{tabular*}{\linewidth}{@{\extracolsep{\fill}}lrrrrrrrr@{}}
\toprule
 &  &  & \multicolumn{3}{c}{$H_{\times}$} & \multicolumn{3}{c}{$H_{\mathrm{within}}$}\\
\cmidrule(lr){4-6}\cmidrule(lr){7-9}
\textbf{Rule} & \textbf{Queries} & \textbf{Users} & \textbf{Estimate} & \textbf{Lower} & \textbf{Upper} & \textbf{Estimate} & \textbf{Lower} & \textbf{Upper}\\
\midrule
Two reads & 59 & 29 & 0.708 & 0.589 & 0.822 & 0.082 & 0.030 & 0.144\\
Taxonomy & 146 & 37 & 0.659 & 0.577 & 0.741 & 0.123 & 0.082 & 0.168\\
\bottomrule
\end{tabular*}
\par\smallskip
\textbf{C. Residual and correctness change}\par\vspace{2pt}
\begin{tabular*}{\linewidth}{@{\extracolsep{\fill}}lrrrrrrrr@{}}
\toprule
 &  &  & \multicolumn{3}{c}{$D_{\mathrm{res}}$} & \multicolumn{3}{c}{$\Delta_{\mathrm{rep}}$}\\
\cmidrule(lr){4-6}\cmidrule(lr){7-9}
\textbf{Rule} & \textbf{Queries} & \textbf{Users} & \textbf{Estimate} & \textbf{Lower} & \textbf{Upper} & \textbf{Estimate} & \textbf{Lower} & \textbf{Upper}\\
\midrule
Two reads & 59 & 29 & 0.626 & 0.491 & 0.756 & 0.307 & 0.132 & 0.486\\
Taxonomy & 146 & 37 & 0.536 & 0.442 & 0.628 & 0.312 & 0.195 & 0.432\\
\bottomrule
\end{tabular*}
\end{table}

The targeted batch X-T contains two support-loss pairs held out from policy fitting. All conditions retain two sources in canonical order, with prompt lengths matched within each pair and two answers per padding variant. Restoration recovers the supported response, whereas replacing the missing record without direct support does not consistently recover correctness (Table~\ref{tab:crossfit-support-rescue}). These pairs also occur in the later all-change batch X-I. Joining query, condition, padding, and draw identifiers verifies identical serialized inputs for all 32 corresponding calls. One support-absent response differs between the separately collected batches, accounting for their different observed counts; the artifact preserves both answer matrices.
\begin{table}[ht]
\centering
\small
\caption{Matched support rescue for two support-loss pairs from the cross-fitted policy. Each condition has eight answer draws; source count and full-prompt token count are matched within each pair.}
\label{tab:crossfit-support-rescue}
\begin{tabular}{lrrrr}
\toprule
\textbf{Condition} & \multicolumn{2}{c}{\textbf{Correctness}} & \multicolumn{2}{c}{\textbf{Answer relation}}\\
\cmidrule(lr){2-3}\cmidrule(lr){4-5}
 & \textbf{Correct} & \textbf{Accuracy} & \textbf{Same as present} & \textbf{Change from absent}\\
\midrule
Support present & 8 & 1.000 & 8 & 5\\
Support absent & 3 & 0.375 & 3 & 0\\
Support rescue & 8 & 1.000 & 8 & 5\\
Replacement without direct support & 4 & 0.500 & 4 & 1\\
\bottomrule
\end{tabular}
\end{table}

\subsection{Cross-fitted support restoration}
\label{app:semantic-crossfit-restoration}
The complete analysis includes all nine support-identity changes selected without answer outcomes, including the two support-loss cases above. Conditions share the source count, canonical serialization, and matched prompt boundary. The main text reports the comparison, and the released matrices contain the source-level labels and selection rule.

\begin{table}[ht]
\centering
\caption{Readings across contexts and models. Every reading sees the same 146 complete source pools under a rubric that withholds outcome information. Labels are reported as admissible, ambiguous, and inadmissible counts; L denotes Luna max, T denotes Terra max, and S denotes Sol max.}
\label{tab:semantic-third-pass}
\begin{tabular*}{0.98\textwidth}{@{\extracolsep{\fill}}lllrrr@{}}
\toprule
\multicolumn{3}{c}{\textbf{Reading specification}} & \multicolumn{3}{c}{\textbf{Label counts}}\\
\cmidrule(lr){1-3}\cmidrule(lr){4-6}
\textbf{Read} & \textbf{Rule} & \textbf{Model} & \textbf{Admissible} & \textbf{Ambiguous} & \textbf{Inadmissible}\\
\midrule
Pass A & Single read & Luna max & 70 & 8 & 68\\
Pass B & Single read & Luna max & 77 & 12 & 57\\
Third context & Single read & Luna max & 25 & 2 & 119\\
Strict plurality & Majority & L/L/L & 64 & 10 & 72\\
\midrule
Third model & Single read & Terra max & 58 & 4 & 84\\
Majority & Majority & L/L/T & 70 & 10 & 66\\
All three & Unanimity & L/L/T & 48 & 0 & 98\\
\midrule
Sol context & Single read & Sol max & 58 & 0 & 88\\
Majority & Majority & L/L/S & 69 & 10 & 67\\
All three & Unanimity & L/L/S & 50 & 0 & 96\\
\bottomrule
\end{tabular*}
\end{table}

\begin{table}[ht]
\centering
\small
\caption{Sensitivity of the semantic gate across models. Route quantities use the retained Luna traces; intervals are 95\% user-bootstrap intervals. The consensus from two Luna readings is the prespecified primary semantic stratum; L, T, and S denote Luna, Terra, and Sol max readings.}
\label{tab:semantic-crossmodel-route}
\setlength{\tabcolsep}{2pt}
\textbf{A. Response coordinates}\par\vspace{2pt}
\begin{tabular*}{\linewidth}{@{\extracolsep{\fill}}llrrrrrrrr@{}}
\toprule
 &  &  &  & \multicolumn{3}{c}{$H_{\times}$} & \multicolumn{3}{c}{$D_{\mathrm{res}}$}\\
\cmidrule(lr){5-7}\cmidrule(lr){8-10}
\textbf{Readers} & \textbf{Rule} & \textbf{Queries} & \textbf{Users} & \textbf{Estimate} & \textbf{Lower} & \textbf{Upper} & \textbf{Estimate} & \textbf{Lower} & \textbf{Upper}\\
\midrule
L/L/T & Majority & 70 & 31 & 0.594 & 0.470 & 0.713 & 0.469 & 0.339 & 0.602\\
L/L/T & Unanimity & 48 & 26 & 0.529 & 0.380 & 0.678 & 0.433 & 0.288 & 0.577\\
L/L/S & Majority & 69 & 32 & 0.599 & 0.478 & 0.715 & 0.463 & 0.332 & 0.598\\
L/L/S & Unanimity & 50 & 27 & 0.585 & 0.441 & 0.725 & 0.474 & 0.319 & 0.623\\
\bottomrule
\end{tabular*}
\par\smallskip
\textbf{B. Correctness change}\par\vspace{2pt}
\begin{tabular*}{\linewidth}{@{\extracolsep{\fill}}llrrrrr@{}}
\toprule
 &  &  &  & \multicolumn{3}{c}{$\Delta_{\mathrm{rep}}$}\\
\cmidrule(lr){5-7}
\textbf{Readers} & \textbf{Rule} & \textbf{Queries} & \textbf{Users} & \textbf{Estimate} & \textbf{Lower} & \textbf{Upper}\\
\midrule
L/L/T & Majority & 70 & 31 & 0.390 & 0.224 & 0.556\\
L/L/T & Unanimity & 48 & 26 & 0.346 & 0.135 & 0.543\\
L/L/S & Majority & 69 & 32 & 0.386 & 0.191 & 0.565\\
L/L/S & Unanimity & 50 & 27 & 0.403 & 0.191 & 0.600\\
\bottomrule
\end{tabular*}
\end{table}

\section{Policy and Transformation Details}
\label{app:policy}

\subsection{Principal policies}
\label{app:policy-principal}
Under the source-order comparator, lexical similarity clustering and survivor resolution both use the fixed benchmark source order. The paired comparison route applies both operations in the declared arrival order. For the comparison route, each cluster retains the member with the largest arrival position. The two routes share the same conflict-cluster representation and answer interface. The raw condition passes the controlled evidence pool to the answer boundary without compaction; the preserved-provenance condition passes its selected context unchanged to the answer model. Single-operation arms keep one of clustering or survivor resolution in source order and apply arrival order to the other, making the two locations where order enters separately inspectable.

\subsection{Policy timing}
\label{app:policy-timing}
Construction is determined by the source representation and source-order convention before answer generation. We fix the focal compactor's overlap threshold at $0.055$ as part of the construction configuration and reuse this configuration for the primary route, operation schedules, semantic-screen reanalyses, and compactor controls; no route-specific tuning is performed. Construction reads only source text and route positions; question text, answer options, gold labels, generated answers, and effect summaries are unavailable to it.

\subsection{Bounded recent policy}
\label{app:policy-recent}
The bounded-recent construction retains the three records with the largest arrival positions in each query's source pool and then presents them in canonical source order. Its paired route reverses the arrival order before applying the same retention rule, so the selected source set can change while the canonical ordering rule remains fixed. The offset-aligned exact-token trace contains 252 answer rows over 63 queries and 29 users; every paired source set changes, and full-prompt token counts plus evidence and question offsets are matched within each pair.

\subsection{Mechanism schedules}
\label{app:policy-mechanism}
The mechanism arms either hold conflict clusters fixed while selecting survivors in arrival order, hold survivor selection fixed while forming clusters in arrival order, or let both operations follow arrival order. The replay schedule reverses the complete pool. The half-swap schedule places the latter half before the earlier half, and the odd-even schedule groups sessions at even positions before sessions at odd positions. These schedules do not see the query and preserve the same multiset of source identifiers.

\subsection{Secondary policies and controls}
\label{app:policy-secondary}
A recent-window condition selects a fixed recent window before the context budget, and its fixed-source comparator selects that same window in source order. A utility-buffer condition ranks records by a deterministic salience score before the budget, and its fixed-source comparator applies the same ranking, using source order to break ties. The provenance-gated, neutral, and placebo variants append distinct instructions to the selected context.

\subsection{Identical-Option Sensitivity}
\label{app:option-sensitivity}

The benchmark scores option identifiers. An exact-text check after whitespace normalization finds duplicate option texts in 16 of the 146 fact-recall queries, including two of the 63 primary queries. Different identifiers can therefore denote the same response, affecting both disagreement and gold-option accuracy. We retain the original scoring rule and perform a post hoc exclusion analysis using the existing answer traces. Removing these two queries leaves 61 focal pairs from 29 users and 31 matched restoration pairs from 21 users. The focal estimates remain positive (Table~\ref{tab:option-sensitivity}). In the reduced restoration set, support-present, support-absent, rescue, and placebo accuracies are 0.873, 0.218, 0.875, and 0.296, respectively; rescue minus absent is 0.657 with a 95\% user-bootstrap interval of $[0.480,0.808]$. This check addresses identical option strings; it does not adjudicate semantic equivalence between differently worded answers.

\begin{table}[ht]
\centering
\caption{Focal estimates after excluding identical-option queries. Each row contains 61 pairs from 29 users and combines one formal answer with two fresh draws per route. A and B use the original padding variants. Lower and Upper are 95\% percentile limits from 10,000 user-bootstrap resamples.}
\label{tab:option-sensitivity}
\begin{tabular*}{\linewidth}{@{\extracolsep{\fill}}lrrrrrr@{}}
\toprule
 & \multicolumn{3}{c}{$D_{\mathrm{res}}$} & \multicolumn{3}{c}{$\Delta_{\mathrm{rep}}$}\\
\cmidrule(lr){2-4}\cmidrule(lr){5-7}
\textbf{Variant} & \textbf{Estimate} & \textbf{Lower} & \textbf{Upper} & \textbf{Estimate} & \textbf{Lower} & \textbf{Upper}\\
\midrule
A & 0.524 & 0.361 & 0.686 & 0.439 & 0.254 & 0.611\\
B & 0.507 & 0.361 & 0.649 & 0.345 & 0.144 & 0.534\\
\bottomrule
\end{tabular*}
\end{table}

\section{Construction Details}
\label{app:construction}

\subsection{Record representation}
\label{app:construction-representation}
Each admissible session is represented by its source identifier, benchmark source position, arrival position under the declared schedule, and serialized text. The compactor's lexical decisions use a session's user-turn text after role markers are removed; the raw condition retains the complete session serialization. The query, answer choices, gold label, question category, retrieval score, and generated answer are excluded from every policy decision.

\subsection{State objects}
\label{app:construction-state}
For route $o$, the similarity partition $K_q^{(o)}$ groups $B_q$ into clusters with representatives $r_{qk}^{(o)}$; the survivor set is $S_q^{(o)}=\{r_{qk}^{(o)}\}_k$. Retrieval selects candidates $E_q^{(o)}\subseteq S_q^{(o)}$, and the budget yields the exposed set $V_q^{(o)}\subseteq E_q^{(o)}$, serialized as $C_q^{(o)}$. All state comparisons use $V_q^{(o)}$ (the set visible at the answer boundary); the operation audit confirms $E_q^{(o)}=V_q^{(o)}$ for every analyzed row.

\subsection{Similarity clustering}
\label{app:construction-clustering}
Within the focal compactor, stopwords and tokens of length at most two are removed from the user-turn text. A new session is then compared with one representative from each existing cluster using Jaccard overlap. A session joins the cluster with the greatest overlap when that overlap is at least the fixed threshold $0.055$; otherwise, it begins a new cluster. Exact overlap ties are assigned to the earliest-created cluster. The representative used for a subsequent comparison is the member with the highest deterministic preference score, defined as the count of lowercased user-turn tokens matching the focal compactor's fixed preference lexicon; exact ties retain the earliest member in processing order. Survivor resolution is a separate step. The canonical arm forms clusters in benchmark source order and retains the member with the largest source position, whereas the arrival route forms clusters in the declared arrival order and retains the member with the largest arrival position, using the stable source identifier as the final tie-break key.

\subsection{Executable focal rule}
\label{app:construction-focal-rule}
The artifact includes \texttt{focal\_compactor.py} under \texttt{data/decisive/focal\_source/}. The following pseudocode separates the comparison representative from the final survivor; \texttt{order} is benchmark order for the canonical comparator and the declared arrival order for the arrival arm.
\begin{quote}\begin{minipage}{\linewidth}\small
\begin{verbatim}
clusters = []
for record in order:
    k, best = None, 0
    for j, cluster in enumerate(clusters):
        rep = first_argmax(cluster, preference_score)
        score = jaccard(tokens(record), tokens(rep))
        if score > best:
            k, best = j, score
    if k is None or best < 0.055:
        clusters.append([record])
    else:
        clusters[k].append(record)
key = source_position if canonical else arrival_position
survivors = [argmax(c, (key, source_id)) for c in clusters]
\end{verbatim}
\end{minipage}\end{quote}
Tokenization extracts \texttt{[USER]} spans case-insensitively, joins them with two newlines, and uses stripped input text when no role marker is present. The token pattern is \verb|[A-Za-z][A-Za-z0-9']+|. Similarity uses the set of lowercased tokens after removing tokens of length at most two and the following stopwords:
\begin{quote}\small
\texttt{a an and are as at be been by can could do for from had has have he her his i if in is it its me my of on or our she that the their them there they this to was we were what when which who will with you your}.
\end{quote}
The preference score counts token occurrences in the lowercased user text matching this lexicon:
\begin{quote}\small
\texttt{avoid appreciate bother boring choice comfortable desire dislike enjoy excited favorite frustrated fulfilling hate interest interested joy like love motivated passion prefer preference relaxed resent satisfy stifled tired want worry enjoyed liked loved preferred disliked hated wanted}.
\end{quote}
An empty token-set union has Jaccard similarity zero. The first maximum implements the representative tie rule, and the strict improvement test implements the earliest-cluster tie rule. Retrieval and serialization then follow the answer-boundary configuration below.

\subsection{Tie audit}
\label{app:construction-tie-audit}
The complete 146-pool audit contains no exact ties in the maximum cluster-assignment overlap score. Ties in the maximum preference score used to choose a comparison representative occur 15 times under source-order construction and 12 times under reversed-arrival construction. On the primary consensus subset of 63 pools, the corresponding counts are 7 and 7. These events concern representative selection for the next comparison, not the final survivor, and are resolved by processing order. The final survivor keys remain source position for the source-order comparator and arrival position for the reversed route, with the source identifier as the deterministic final tie-break key.

\subsection{Retrieval and budget boundary}
\label{app:construction-retrieval}
After memory construction, the default lexical retriever retains up to ten candidates. The ordinary retrieval path resolves equal scores by their position in the constructed list. The provenance and matched-context controls use a source-position tie-break that is invariant to order when the selected source set must be held fixed; they may then serialize that fixed set in arrival order. Context construction admits complete records in the selected budget order until the 32,768-character cap is reached. The final context is serialized from the selected source identities, and the descriptive length proxy is $\lceil n/4\rceil$ for a context containing $n$ characters. This proxy is reported only as a compact diagnostic; exact token-count claims use the tokenizer-matched control described below.

\subsection{Operation-level schedules}
\label{app:construction-schedules}
Across mechanism arms, the evidence pool and answer boundary remain unchanged, while dependence on order is assigned to different construction operations. The both-operations path lets arrival order determine both cluster formation and survivor choice. The survivor selection path forms clusters in source order and lets arrival order choose the survivor. The clustering-only path forms clusters in arrival order and chooses the survivor by source position. Each schedule then applies the same permutation independent of the query. The schedules use full reversal, half-swap, or odd-even grouping. The complete schedule matrix is reported in Table~\ref{tab:full-mechanism-schedule}, and the state-layer diagnostic is reported in Table~\ref{tab:mechanism-state}.

\subsection{Cross-fitted streaming policy}
\label{app:construction-crossfit}
For each held-out-user split, a TF--IDF ridge predictor fitted on other users' documents scores each document by its mean pairwise lexical overlap. An online rule with capacity three maximizes summed scores plus $0.35\log(1+v)$, where $v$ counts distinct non-stopword tokens in the subset; discarded sessions are not reconsidered and ties use a stable source identifier. The arrival-order route processes the held-out sequence in schedule order; the comparator uses canonical source order.

The internal-state signature comprises fitted model parameters and retained records canonicalized by source identity. Reconstructing both routes reproduces retained identities, parameters, exposed sources, and context digests in all 1,178 archived rows; 51 query pairs show state changes, of which 11 share the same answer context.

\subsection{State reuse across questions}
\label{app:construction-state-reuse}
The consensus stratum contributes 63 queries from 29 users; the lexicographically smallest query per user defines the construction anchor. Construction uses only the anchor's source pool (no question text, answer choices, or gold labels). Two neutral padding variants and two fresh draws per route yield 272 answer rows. Under formal aggregation: $H_{\times}=0.555\;[0.394,0.715]$, $H_{\mathrm{within}}=0.107\;[0.040,0.185]$, $D_{\mathrm{res}}=0.448\;[0.263,0.633]$, $\Delta_{\mathrm{rep}}=0.405\;[0.208,0.592]$ (95\% CIs). Support-stratified estimates appear in Table~\ref{tab:state-transfer-support}.

\begin{table}[ht]
\centering
\caption{Multiple-question consequences of fixed states under overall and support-constrained strata. Estimates average padding variants before assigning equal weight to users; Lower and Upper columns give 95\% bootstrap limits. Targets in the support-present stratum differ in direct support across the paired states; targets in the support-absent stratum expose no direct support under either route.}
\label{tab:state-transfer-support}
\label{tab:transfer-main}
\setlength{\tabcolsep}{2pt}
\textbf{A. Response disagreement}\par\vspace{2pt}
\begin{tabular*}{\linewidth}{@{\extracolsep{\fill}}lrrrrrrrr@{}}
\toprule
 &  &  & \multicolumn{3}{c}{$H_{\times}$} & \multicolumn{3}{c}{$H_{\mathrm{within}}$}\\
\cmidrule(lr){4-6}\cmidrule(lr){7-9}
\textbf{Stratum} & \textbf{Queries} & \textbf{Users} & \textbf{Estimate} & \textbf{Lower} & \textbf{Upper} & \textbf{Estimate} & \textbf{Lower} & \textbf{Upper}\\
\midrule
All targets & 34 & 20 & 0.555 & 0.394 & 0.715 & 0.107 & 0.040 & 0.185\\
Support present & 18 & 13 & 0.712 & 0.538 & 0.865 & 0.135 & 0.048 & 0.240\\
Support absent & 16 & 11 & 0.409 & 0.182 & 0.636 & 0.057 & 0.000 & 0.136\\
\bottomrule
\end{tabular*}
\par\smallskip
\textbf{B. Residual and correctness change}\par\vspace{2pt}
\begin{tabular*}{\linewidth}{@{\extracolsep{\fill}}lrrrrrrrr@{}}
\toprule
 &  &  & \multicolumn{3}{c}{$D_{\mathrm{res}}$} & \multicolumn{3}{c}{$\Delta_{\mathrm{rep}}$}\\
\cmidrule(lr){4-6}\cmidrule(lr){7-9}
\textbf{Stratum} & \textbf{Queries} & \textbf{Users} & \textbf{Estimate} & \textbf{Lower} & \textbf{Upper} & \textbf{Estimate} & \textbf{Lower} & \textbf{Upper}\\
\midrule
All targets & 34 & 20 & 0.448 & 0.263 & 0.633 & 0.405 & 0.208 & 0.592\\
Support present & 18 & 13 & 0.577 & 0.365 & 0.788 & 0.673 & 0.519 & 0.827\\
Support absent & 16 & 11 & 0.352 & 0.102 & 0.625 & 0.114 & $-0.114$ & 0.386\\
\bottomrule
\end{tabular*}
\par\smallskip
\textbf{C. Support-stratum contrast (aligned minus absent)}\par\vspace{2pt}
\begin{tabular*}{\linewidth}{@{\extracolsep{\fill}}lrrr@{}}
\toprule
 & \multicolumn{3}{c}{$\Delta_{\mathrm{rep}}$}\\
\cmidrule(lr){2-4}
\textbf{Contrast} & \textbf{Estimate} & \textbf{Lower} & \textbf{Upper}\\
\midrule
Aligned minus absent & 0.559 & 0.239 & 0.841\\
\bottomrule
\end{tabular*}
\end{table}

\subsection{Persistent continuation}
\label{app:construction-persistent-continuation}
MemoChat constructs one route-specific state per user and reuses it for held-out target queries (20 users, 34 targets; 8 users with $\ge$2 targets). The policy changes the compiled context in all 68 target--padding groups but source lineage in only 2, yielding $H_{\times}=0.084\;[0.038,0.141]$, $H_{\mathrm{within}}=0.098\;[0.038,0.163]$, $D_{\mathrm{res}}=-0.014\;[-0.030,0.000]$, and $\Delta_{\mathrm{rep}}=-0.002\;[-0.050,0.048]$ (95\% CIs).

\begin{table}[ht]
\centering
\small
\caption{Persistent MemoChat continuation. State reuse reports users, targets, users with at least two targets (Multi), changed source lineage ($\Delta S$), and changed context ($\Delta C$). Profile estimates give equal weight to users; Lower and Upper columns give 95\% bootstrap limits for target--padding-group summaries, with users as the resampling unit.}
\label{tab:external-memochat-transfer}
\setlength{\tabcolsep}{2pt}
\textbf{A. State reuse}\par\vspace{2pt}
\begin{tabular*}{\linewidth}{@{\extracolsep{\fill}}lrrrrr@{}}
\toprule
 & \multicolumn{3}{c}{Coverage} & \multicolumn{2}{c}{Changed pairs}\\
\cmidrule(lr){2-4}\cmidrule(lr){5-6}
\textbf{Policy} & \textbf{Users} & \textbf{Targets} & \textbf{Multi} & \textbf{$\Delta S$} & \textbf{$\Delta C$}\\
\midrule
MemoChat & 20 & 34 & 8 & 2 & 68\\
\bottomrule
\end{tabular*}
\par\smallskip
\textbf{B. Response profile}\par\vspace{2pt}
\begin{tabular*}{\linewidth}{@{\extracolsep{\fill}}rrrrrrrrr@{}}
\toprule
\multicolumn{3}{c}{$J$} & \multicolumn{3}{c}{$H_{\times}$} & \multicolumn{3}{c}{$H_{\mathrm{within}}$}\\
\cmidrule(lr){1-3}\cmidrule(lr){4-6}\cmidrule(lr){7-9}
\textbf{Estimate} & \textbf{Lower} & \textbf{Upper} & \textbf{Estimate} & \textbf{Lower} & \textbf{Upper} & \textbf{Estimate} & \textbf{Lower} & \textbf{Upper}\\
\midrule
0.993 & 0.979 & 1.000 & 0.084 & 0.038 & 0.141 & 0.098 & 0.038 & 0.163\\
\bottomrule
\end{tabular*}
\par\smallskip
\textbf{C. Residual and correctness change}\par\vspace{2pt}
\begin{tabular*}{\linewidth}{@{\extracolsep{\fill}}rrrrrr@{}}
\toprule
\multicolumn{3}{c}{$D_{\mathrm{res}}$} & \multicolumn{3}{c}{$\Delta_{\mathrm{rep}}$}\\
\cmidrule(lr){1-3}\cmidrule(lr){4-6}
\textbf{Estimate} & \textbf{Lower} & \textbf{Upper} & \textbf{Estimate} & \textbf{Lower} & \textbf{Upper}\\
\midrule
$-0.014$ & $-0.030$ & 0.000 & $-0.002$ & $-0.050$ & 0.048\\
\bottomrule
\end{tabular*}
\par\smallskip
\textbf{D. Accuracy}\par\vspace{2pt}
\begin{tabular*}{\linewidth}{@{\extracolsep{\fill}}rrrrrr@{}}
\toprule
\multicolumn{3}{c}{Chronological} & \multicolumn{3}{c}{Replay}\\
\cmidrule(lr){1-3}\cmidrule(lr){4-6}
\textbf{Estimate} & \textbf{Lower} & \textbf{Upper} & \textbf{Estimate} & \textbf{Lower} & \textbf{Upper}\\
\midrule
0.876 & 0.754 & 0.972 & 0.874 & 0.775 & 0.956\\
\bottomrule
\end{tabular*}
\end{table}

\subsection{Permutation and Retention Structure}
\label{app:order-structure}
We enumerate all $n_q!$ arrival permutations of each source pool (primary: 29 pools, 18,336 permutations; broader 146-query set: 37 pools, 20,400 permutations). For each threshold in $\{0.025,0.055,0.100\}$, clustering follows arrival order and three survivor rules are compared: last arrival, first arrival, or largest source position. A control fixing both operations to source order is permutation-invariant.

For each permutation $\pi$ we record whether $S_q^{(\pi)}$ differs from its forward counterpart and whether it intersects the annotated support set. Uniform-permutation quantities average over permutations, then questions within users, then users equally. These are survivor measurements before retrieval; no answers are generated.

\begin{table}[htbp]
\centering\small
\caption{Exhaustive survivor analysis on 63 queries from 29 users. Change is the probability of a survivor-set difference from forward construction under uniform permutations. Forward, Reverse, and Uniform give direct-support retention under the respective arrival distributions. All entries weight users equally.}
\label{tab:order-structure}
\setlength{\tabcolsep}{3pt}
\begin{tabular*}{\linewidth}{@{\extracolsep{\fill}}rlrrrr@{}}
\toprule
\textbf{Threshold} & \textbf{Survivor rule} & \textbf{Change} & \textbf{Forward} & \textbf{Reverse} & \textbf{Uniform}\\
\midrule
0.025 & Last arrival & 0.806 & 0.105 & 0.635 & 0.304\\
0.025 & First arrival & 0.806 & 0.635 & 0.070 & 0.301\\
0.025 & Largest source position & 0.019 & 0.105 & 0.070 & 0.099\\
0.055 & Last arrival & 0.815 & 0.116 & 0.635 & 0.316\\
0.055 & First arrival & 0.810 & 0.635 & 0.116 & 0.319\\
0.055 & Largest source position & 0.040 & 0.116 & 0.116 & 0.118\\
0.100 & Last arrival & 0.797 & 0.116 & 0.635 & 0.316\\
0.100 & First arrival & 0.797 & 0.635 & 0.116 & 0.316\\
0.100 & Largest source position & 0.000 & 0.116 & 0.116 & 0.116\\
\bottomrule
\end{tabular*}
\end{table}

The endpoint asymmetry follows the interaction between source position and retention. Across 94 query--support-record instances, normalized benchmark position (zero-based source position divided by $n_q-1$) has median $0.250$. The four intervals $[0,0.25)$, $[0.25,0.5)$, $[0.5,0.75)$, and $[0.75,1]$ contain 41, 12, 15, and 26 instances, respectively. For the 33 queries retaining support only under reversal, the mean of query-level mean support positions is $0.110$. Reversal therefore preferentially preserves early evidence under last-arrival selection. This explains an endpoint advantage on these pools without estimating an accuracy effect over arbitrary permutations.

Cluster structure also varies with the threshold. Across the 18,336 primary pool permutations, mean cluster counts are $1.332$, $1.739$, and $2.099$ at thresholds $0.025$, $0.055$, and $0.100$, respectively. These summaries weight permutations equally, so larger pools receive more weight; the artifact retains exact cluster-count and cluster-size distributions for each pool. For a Recent-3 control before budget application, the probability of retaining at least one of $k_q$ direct-support records under a uniform permutation is $1-\binom{n_q-k_q}{3}/\binom{n_q}{3}$, with the numerator zero when fewer than three non-support records exist. The released script evaluates this control on the same fixed annotations.

\section{Uncertainty and Complete Estimates}
\label{app:uncertainty}

\subsection{Bootstrap unit}
\label{app:uncertainty-bootstrap}
All intervals are percentile bootstrap intervals resampling users with equal weight, conditioning on the fixed benchmark split, policy rules, route transformations, and semantic/support labels. Fresh-answer quantities use the repeated-draw estimands below. For provenance, $\Gamma_H=H_{\mathrm{gated}}-H_{\mathrm{preserved}}$ and $\Gamma_{\Delta}=\Delta_{\mathrm{gated}}-\Delta_{\mathrm{preserved}}$.

\subsection{Answer-stage diagnostic}
\label{app:uncertainty-answer-stage}
For $m$ forward draws $F_1,\ldots,F_m$ and $m$ replay draws $R_1,\ldots,R_m$, the query-level residual route distance is
\begin{equation}
\begin{aligned}
 D_{q,\mathrm{res}}^{(m)}={}&\frac{1}{m^2}\sum\nolimits_{i=1}^{m}\sum\nolimits_{j=1}^{m}\mathbf{1}[F_i\ne R_j]\\
 &-\frac{1}{2m(m-1)}\left(\sum\nolimits_{\substack{1\le i,j\le m\\i\ne j}}\mathbf{1}[F_i\ne F_j]+\sum\nolimits_{\substack{1\le i,j\le m\\i\ne j}}\mathbf{1}[R_i\ne R_j]\right).
\end{aligned}
\end{equation}
The focal repeat uses $m=2$ (fresh draws); extensions combining the formal answer with fresh draws use $m=3$. The population target is $D_q^\star=\tfrac12\sum_{a\in\mathcal A_q}(p_{qa}^{(f)}-p_{qa}^{(r)})^2$; under stationarity $D_{q,\mathrm{res}}^{(m)}$ is unbiased for $D_q^\star$ (though finite-sample values may be negative). Policy comparisons use $\Gamma_{D_{\mathrm{res}}}(\tau)=\overline{D}_{\mathrm{res},\mathrm{ord}}(\tau)-\overline{D}_{\mathrm{res},\mathrm{can}}(\tau)$. Complete condition estimates appear in Table~\ref{tab:full-primary}.

\subsection{Complete policy and provenance estimates}
\label{app:uncertainty-estimates}

\begin{table}[ht]
\centering
\caption{Complete policy and provenance condition estimates. Lower and Upper columns give 95\% bootstrap limits with equal weight for users; retained-state and length diagnostics are in Table~\ref{tab:full-pathways}.}
\label{tab:full-primary}
\begin{tabular*}{0.98\textwidth}{@{\extracolsep{\fill}}lrrrrrr@{}}
\toprule
\textbf{Condition} & \multicolumn{3}{c}{\textbf{$H$}} & \multicolumn{3}{c}{\textbf{$\Delta$}}\\
\cmidrule(lr){2-4}\cmidrule(lr){5-7}
 & \textbf{Estimate} & \textbf{Lower} & \textbf{Upper} & \textbf{Estimate} & \textbf{Lower} & \textbf{Upper}\\
\midrule
Raw & 0.149 & 0.115 & 0.184 & $-0.023$ & $-0.044$ & $-0.002$\\
Canonical & 0.166 & 0.128 & 0.208 & 0.016 & $-0.018$ & 0.048\\
Arrival route & 0.509 & 0.449 & 0.569 & 0.127 & 0.074 & 0.183\\
Preserved & 0.150 & 0.121 & 0.181 & $-0.007$ & $-0.041$ & 0.027\\
Gated & 0.160 & 0.125 & 0.200 & $-0.020$ & $-0.057$ & 0.014\\
Neutral & 0.126 & 0.099 & 0.155 & 0.000 & $-0.025$ & 0.025\\
Placebo & 0.167 & 0.140 & 0.195 & $-0.009$ & $-0.046$ & 0.027\\
\bottomrule
\end{tabular*}
\end{table}

Table~\ref{tab:paired-contrasts} reports the paired contrasts and diagnostics at the answer boundary.

\subsection{Paired contrasts and answer-stage diagnostics}
\label{app:uncertainty-contrasts}

\begin{table}[ht]
\centering
\caption{Primary paired contrasts and diagnostics at the answer boundary. Accuracy contrasts are gated minus preserved. Lower and Upper columns give 95\% bootstrap limits with equal weight for users.}
\label{tab:paired-contrasts}
\setlength{\tabcolsep}{2pt}
\textbf{A. Policy contrasts}\par\vspace{2pt}
\begin{tabular*}{\linewidth}{@{\extracolsep{\fill}}lrrrrrr@{}}
\toprule
 & \multicolumn{3}{c}{$\Gamma_H$} & \multicolumn{3}{c}{$\Gamma_\Delta$}\\
\cmidrule(lr){2-4}\cmidrule(lr){5-7}
\textbf{Comparison} & \textbf{Estimate} & \textbf{Lower} & \textbf{Upper} & \textbf{Estimate} & \textbf{Lower} & \textbf{Upper}\\
\midrule
Exchangeable (storage permutation) & 0.528 & 0.426 & 0.624 & 0.302 & 0.192 & 0.414\\
Omnibus (reversal diagnostic) & 0.343 & 0.275 & 0.411 & 0.111 & 0.045 & 0.179\\
\bottomrule
\end{tabular*}
\par\smallskip
\textbf{B. Answer-stage diagnostics}\par\vspace{2pt}
\begin{tabular*}{\linewidth}{@{\extracolsep{\fill}}rrrrrrrrr@{}}
\toprule
\multicolumn{3}{c}{Same-prompt flip rate} & \multicolumn{3}{c}{Chronological accuracy contrast} & \multicolumn{3}{c}{Replay accuracy contrast}\\
\cmidrule(lr){1-3}\cmidrule(lr){4-6}\cmidrule(lr){7-9}
\textbf{Estimate} & \textbf{Lower} & \textbf{Upper} & \textbf{Estimate} & \textbf{Lower} & \textbf{Upper} & \textbf{Estimate} & \textbf{Lower} & \textbf{Upper}\\
\midrule
0.124 & 0.087 & 0.164 & $-0.015$ & $-0.054$ & 0.026 & $-0.028$ & $-0.059$ & 0.004\\
\bottomrule
\end{tabular*}
\end{table}

\section{Additional Conditions}
\label{app:additional}

\subsection{Semantic-stratum estimates}
\label{app:additional-strata}

Table~\ref{tab:question-strata-main} reports the complete estimates for each semantic stratum.

\begin{table}[ht]
\centering
\caption{Semantic-stratum summaries. Panel A reports storage-policy contrasts $\Gamma_H$ and $\Gamma_\Delta$ for exchangeable and residual queries. Panel B reports raw reversal coordinates $H,\Delta$ and chronology-gated provenance contrasts $\Gamma_H^{\mathrm{prov}},\Gamma_\Delta^{\mathrm{prov}}$ for the same 238 temporal queries from 33 users; the superscript $\mathrm{prov}$ denotes gated minus preserved. Lower and Upper give 95\% user-bootstrap limits. Estimates weight users equally; raw temporal reversal is retained as a chronology diagnostic.}
\label{tab:question-strata-main}
\setlength{\tabcolsep}{2pt}
\textbf{A. Storage-policy contrasts}\par\vspace{2pt}
\begin{tabular*}{\linewidth}{@{\extracolsep{\fill}}lrrrrrrrr@{}}
\toprule
 &  &  & \multicolumn{3}{c}{$\Gamma_H$} & \multicolumn{3}{c}{$\Gamma_\Delta$}\\
\cmidrule(lr){4-6}\cmidrule(lr){7-9}
\textbf{Stratum} & \textbf{Queries} & \textbf{Users} & \textbf{Estimate} & \textbf{Lower} & \textbf{Upper} & \textbf{Estimate} & \textbf{Lower} & \textbf{Upper}\\
\midrule
Exchangeable & 146 & 37 & 0.528 & 0.426 & 0.624 & 0.302 & 0.192 & 0.414\\
Residual & 205 & 37 & 0.317 & 0.229 & 0.410 & 0.102 & 0.017 & 0.185\\
\bottomrule
\end{tabular*}
\par\smallskip
\textbf{B. Temporal diagnostics}\par\vspace{2pt}
\setlength{\tabcolsep}{1.6pt}
\begin{tabular*}{\linewidth}{@{\extracolsep{\fill}}rrrrrrrrrrrr@{}}
\toprule
\multicolumn{6}{c}{Raw reversal} & \multicolumn{6}{c}{Gated minus preserved}\\
\cmidrule(lr){1-6}\cmidrule(lr){7-12}
\multicolumn{3}{c}{$H$} & \multicolumn{3}{c}{$\Delta$} & \multicolumn{3}{c}{$\Gamma_H^{\mathrm{prov}}$} & \multicolumn{3}{c}{$\Gamma_\Delta^{\mathrm{prov}}$}\\
\cmidrule(lr){1-3}\cmidrule(lr){4-6}\cmidrule(lr){7-9}\cmidrule(lr){10-12}
Estimate & Lower & Upper & Estimate & Lower & Upper & Estimate & Lower & Upper & Estimate & Lower & Upper\\
\midrule
0.242 & 0.142 & 0.345 & $-0.089$ & $-0.197$ & 0.001 & $-0.018$ & $-0.106$ & 0.057 & 0.017 & $-0.083$ & 0.118\\
\bottomrule
\end{tabular*}
\end{table}

Coverage from the taxonomy is supplemented by a consensus audit over complete source pools. For each exchangeable fact query, two independent outcome-blinded readings inspected the fixed evidence pool $B_q$. The question, four answer options, and four to seven source blocks appeared in fixed positions; the gold label, route outputs, selected identities, answers, correctness, and effect estimates were withheld. Exact agreement was observed for 116 of 146 query pools ($0.795$; Cohen's $\kappa=0.634$). The intersection of the two admissible labels retains 63 queries from 29 users. On this set, the order-following route has $\overline{H}=0.571$ (95\% CI $[0.425,0.711]$) and $\overline{\Delta}=0.366$ (95\% CI $[0.178,0.546]$); the source-order comparator has $\overline{H}=0.076$ (95\% CI $[0.007,0.169]$) and $\overline{\Delta}=0.028$ (95\% CI $[-0.048,0.121]$). The corresponding policy contrasts are $\Gamma_H=0.495$ (95\% CI $[0.330,0.656]$) and $\Gamma_\Delta=0.338$ (95\% CI $[0.143,0.529]$).

\begin{table}[ht]
\centering
\caption{Semantic sensitivity audit over complete pools. Panel A reports the outcome-blind screen, Panel B reports inter-reader agreement, and Panel C reports storage-policy contrasts $\Gamma_H$ and $\Gamma_\Delta$ on the retained intersection. Lower and upper columns are 95\% bootstrap limits with equal weight for users. NA means not applicable; NR means not reported.}
\label{tab:record-semantic}
\setlength{\tabcolsep}{2pt}
\textbf{A. Semantic screen}\par\vspace{2pt}
\begin{tabular*}{\linewidth}{@{\extracolsep{\fill}}lrrrrr@{}}
\toprule
 &  &  & \multicolumn{3}{c}{Label counts}\\
\cmidrule(lr){4-6}
\textbf{Audit quantity} & \textbf{Queries} & \textbf{Users} & \textbf{Admissible} & \textbf{Ambiguous} & \textbf{Inadmissible}\\
\midrule
Complete pool & 146 & 37 & \multicolumn{1}{l}{NA} & \multicolumn{1}{l}{NA} & \multicolumn{1}{l}{NA}\\
Pass A & 146 & 37 & 70 & 8 & 68\\
Pass B & 146 & 37 & 77 & 12 & 57\\
Unanimous admissible & 63 & 29 & 63 & \multicolumn{1}{l}{NA} & \multicolumn{1}{l}{NA}\\
\bottomrule
\end{tabular*}
\par\smallskip
\textbf{B. Agreement}\par\vspace{2pt}
\begin{tabular*}{\linewidth}{@{\extracolsep{\fill}}lrrrrr@{}}
\toprule
 &  &  & \multicolumn{3}{c}{Agreement}\\
\cmidrule(lr){4-6}
\textbf{Quantity} & \textbf{Queries} & \textbf{Users} & \textbf{Estimate} & \textbf{Lower} & \textbf{Upper}\\
\midrule
Exact agreement & 116 & 37 & 0.795 & \multicolumn{1}{l}{NR} & \multicolumn{1}{l}{NR}\\
Cohen's $\kappa$ & 146 & 37 & 0.634 & \multicolumn{1}{l}{NR} & \multicolumn{1}{l}{NR}\\
\bottomrule
\end{tabular*}
\par\smallskip
\textbf{C. Policy contrasts}\par\vspace{2pt}
\begin{tabular*}{\linewidth}{@{\extracolsep{\fill}}rrrrrrrr@{}}
\toprule
 &  & \multicolumn{3}{c}{$\Gamma_H$} & \multicolumn{3}{c}{$\Gamma_\Delta$}\\
\cmidrule(lr){3-5}\cmidrule(lr){6-8}
\textbf{Queries} & \textbf{Users} & \textbf{Estimate} & \textbf{Lower} & \textbf{Upper} & \textbf{Estimate} & \textbf{Lower} & \textbf{Upper}\\
\midrule
63 & 29 & 0.495 & 0.330 & 0.656 & 0.338 & 0.143 & 0.529\\
\bottomrule
\end{tabular*}
\end{table}

Table~\ref{tab:fullpool-semantic-confusion} shows the cell-level agreement pattern underlying the audit counts.

\begin{table}[ht]
\centering
\caption{Confusion matrix for two independent readings of complete pools. Rows are reading A labels and columns are reading B labels; the unanimous-admissible cell defines the retained sensitivity subset.}
\label{tab:fullpool-semantic-confusion}
\begin{tabular}{lrrr}
\toprule
\textbf{Reading A} & \multicolumn{3}{c}{\textbf{Reading B}}\\
\cmidrule(lr){2-4}
 & \textbf{Admissible} & \textbf{Ambiguous} & \textbf{Inadmissible}\\
\midrule
\textbf{Admissible} & 63 & 2 & 5\\
\textbf{Ambiguous} & 7 & 1 & 0\\
\textbf{Inadmissible} & 7 & 9 & 52\\
\bottomrule
\end{tabular}
\end{table}

Table~\ref{tab:semantic-sensitivity} applies the semantic masks to the canonicalized two-draw repeat matrix. Every retained query changes its exposed sources. Residual disagreement and correctness change have positive user-weighted estimates; $H_{\mathrm{eq}}$ counts answer changes with unchanged correctness.

\begin{table}[ht]
\centering
\caption{Sensitivity of the canonicalized two-draw route profile to the fixed semantic screen. The $H_{\mathrm{eq}}$ panel reports the equal-user-weighted rate of answer changes with unchanged correctness, where $H_{\mathrm{eq},q}=\mathbf{1}[a_q^{(f)}\ne a_q^{(r)}\ \land\ c_q^{(f)}=c_q^{(r)}]$. Lower and upper columns give 95\% bootstrap limits with equal weight for users.}
\label{tab:semantic-sensitivity}
\setlength{\tabcolsep}{2pt}
\textbf{A. Route coordinates}\par\vspace{2pt}
\begin{tabular*}{\linewidth}{@{\extracolsep{\fill}}lrrrrrrrr@{}}
\toprule
 &  &  & \multicolumn{3}{c}{$D_{\mathrm{res}}$} & \multicolumn{3}{c}{$\Delta_{\mathrm{rep}}$}\\
\cmidrule(lr){4-6}\cmidrule(lr){7-9}
\textbf{Mask} & \textbf{Queries} & \textbf{Users} & \textbf{Estimate} & \textbf{Lower} & \textbf{Upper} & \textbf{Estimate} & \textbf{Lower} & \textbf{Upper}\\
\midrule
A admissible & 70 & 31 & 0.496 & 0.370 & 0.632 & 0.441 & 0.259 & 0.605\\
B admissible & 77 & 33 & 0.484 & 0.347 & 0.619 & 0.351 & 0.172 & 0.518\\
Consensus & 63 & 29 & 0.459 & 0.322 & 0.598 & 0.403 & 0.213 & 0.580\\
\bottomrule
\end{tabular*}
\par\smallskip
\textbf{B. Same correctness}\par\vspace{2pt}
\begin{tabular*}{\linewidth}{@{\extracolsep{\fill}}lrrrrr@{}}
\toprule
 &  &  & \multicolumn{3}{c}{$H_{\mathrm{eq}}$}\\
\cmidrule(lr){4-6}
\textbf{Mask} & \textbf{Queries} & \textbf{Users} & \textbf{Estimate} & \textbf{Lower} & \textbf{Upper}\\
\midrule
A admissible & 70 & 31 & 0.040 & 0.006 & 0.083\\
B admissible & 77 & 33 & 0.072 & 0.021 & 0.133\\
Consensus & 63 & 29 & 0.048 & 0.008 & 0.100\\
\bottomrule
\end{tabular*}
\end{table}

A frozen rubric is applied to each of the 146 queries using the question, answer options, gold option, and complete source pool. A source is labeled \emph{direct} when its user-authored content supplies the fact, preference, event, or state needed to select the gold option; \emph{indirect} when it provides relevant background without entailing that option; and \emph{none} when it provides no support. Each card records a short textual basis and confidence. Route outputs, source identities selected by each route, correctness, and effect estimates are withheld. Two independent fresh Luna readings at maximum reasoning effort agree on 697 of 778 source labels ($0.896$; Cohen's $\kappa=0.842$) and on 125 of 146 direct-support sets ($0.856$). The support-change assignment agrees for 140 of 146 coverage queries and for 60 of the 63 consensus queries; the second reading identifies 40 changed pairs out of 63, compared with 39 out of 63 in the prespecified primary reading. We retain the prespecified primary read for the reported estimates. Leave-one-user-out checks keep the rescue-minus-absent accuracy contrast between 0.654 and 0.729 and the rescue-minus-present contrast between $-0.019$ and 0.019. Among the 13 users represented in both support groups, the changed-minus-preserved contrasts remain 0.306--0.472 for $H$ and 0.507--0.632 for $\Delta$.

\subsection{Answer-bearing support localization}
\label{app:additional-support-localization}
All 63 consensus pairs change the exposed source identity set; 39 also change the direct-support set, while 24 preserve it. The corresponding split in the 146-query coverage layer contains 92 changed pairs and 54 preserved pairs. Estimates and intervals are reported in Table~\ref{tab:support-localization}. 

\begin{table}[ht]
\centering
\caption{Answer-bearing support localization. Rows give estimates with equal weight for users; lower and upper columns are 95\% bootstrap limits. Support status records whether the source set that directly supports the gold option changes. Difference reports the within-user changed-minus-preserved contrast among users represented in both groups: 13 for Consensus and 26 for Coverage. NA means not applicable.}
\label{tab:support-localization}
\setlength{\tabcolsep}{2pt}
\begin{tabular*}{\linewidth}{@{\extracolsep{\fill}}llrrrrrrrr@{}}
\toprule
 &  &  &  & \multicolumn{3}{c}{$H$} & \multicolumn{3}{c}{$\Delta$}\\
\cmidrule(lr){5-7}\cmidrule(lr){8-10}
\textbf{Stratum} & \textbf{Support} & \textbf{Queries} & \textbf{Users} & \textbf{Estimate} & \textbf{Lower} & \textbf{Upper} & \textbf{Estimate} & \textbf{Lower} & \textbf{Upper}\\
\midrule
Consensus & Changed & 39 & 26 & 0.724 & 0.571 & 0.865 & 0.583 & 0.349 & 0.792\\
Consensus & Preserved & 24 & 16 & 0.284 & 0.112 & 0.479 & 0.125 & 0.000 & 0.281\\
Coverage & Changed & 92 & 37 & 0.720 & 0.628 & 0.813 & 0.494 & 0.354 & 0.631\\
Coverage & Preserved & 54 & 26 & 0.500 & 0.351 & 0.649 & 0.056 & $-0.028$ & 0.163\\
Consensus & Difference & \multicolumn{1}{l}{NA} & 13 & 0.359 & 0.013 & 0.686 & 0.545 & 0.250 & 0.814\\
Coverage & Difference & \multicolumn{1}{l}{NA} & 26 & 0.195 & $-0.004$ & 0.394 & 0.553 & 0.407 & 0.692\\
\bottomrule
\end{tabular*}
\end{table}

The intervention uses the 39 consensus pairs identified by the outcome-blinded support audit, whose rubric has access to neither route outputs nor answer outputs and uses the gold option as its target. In 35 pairs, one directly supporting source is replaced by an eligible source without direct support under the fixed source order; four additional pairs swap two directly supporting identities and are retained as a separate diagnostic. The replacement identity is determined from the source labels and the identities selected by each route before answer outputs are read. Each condition keeps the selected source count, canonical ordering rule, evidence representation, question, options, common prefix, total token count, and answer position fixed. Two padding forms applied symmetrically across routes and two fresh answer draws per condition provide repeated observations at the same answer boundary. The table reports user-weighted summaries and paired contrasts; Figure~\ref{fig:support-restoration} shows user-level distributions for the source-count-matched rescue and state-reuse analyses.

\begin{table}[ht]
\centering
\caption{Controlled support intervention at the answer boundary. The support-presence estimates use 35 eligible pairs from 23 users; the identity-swap diagnostic uses 4 additional pairs from 4 users. The Lower and Upper columns give 95\% bootstrap confidence limits with equal user weight; identity-swap rows are a separate diagnostic.}
\label{tab:support-intervention}
\setlength{\tabcolsep}{2pt}
\textbf{A. Accuracy}\par\vspace{2pt}
\begin{tabular*}{\linewidth}{@{\extracolsep{\fill}}rrrrrrrrr@{}}
\toprule
\multicolumn{3}{c}{Support present accuracy} & \multicolumn{3}{c}{Support removed accuracy} & \multicolumn{3}{c}{Present minus removed accuracy}\\
\cmidrule(lr){1-3}\cmidrule(lr){4-6}\cmidrule(lr){7-9}
\textbf{Estimate} & \textbf{Lower} & \textbf{Upper} & \textbf{Estimate} & \textbf{Lower} & \textbf{Upper} & \textbf{Estimate} & \textbf{Lower} & \textbf{Upper}\\
\midrule
0.902 & 0.833 & 0.960 & 0.188 & 0.069 & 0.326 & 0.714 & 0.569 & 0.844\\
\bottomrule
\end{tabular*}
\par\smallskip
\textbf{B. Paired diagnostics}\par\vspace{2pt}
\begin{tabular*}{\linewidth}{@{\extracolsep{\fill}}lrrrrrr@{}}
\toprule
 & \multicolumn{3}{c}{$H$} & \multicolumn{3}{c}{$\Delta$}\\
\cmidrule(lr){2-4}\cmidrule(lr){5-7}
\textbf{Intervention} & \textbf{Estimate} & \textbf{Lower} & \textbf{Upper} & \textbf{Estimate} & \textbf{Lower} & \textbf{Upper}\\
\midrule
Matched & 0.714 & 0.572 & 0.844 & 0.714 & 0.569 & 0.841\\
Identity swap & 0.063 & 0.000 & 0.188 & $-0.063$ & $-0.188$ & 0.000\\
\bottomrule
\end{tabular*}
\end{table}

\subsection{Route-support rescue}
\label{app:additional-route-rescue}
Using the gold target and no generated answers, the support audit identifies 35 single-source candidates; matching the selected source count across all four conditions retains 33 eligible pairs, with two structurally mismatched candidates kept as a separate diagnostic. Each eligible pair contains support-present, support-absent, support-rescue, and matched-placebo states without direct support. Support-present and support-rescue accuracies are nearly identical (0.875 and 0.881), whereas support-absent and support-placebo accuracies are lower (0.210 and 0.280). The paired response contrasts are 0.694 for the original route, 0.724 for rescue, and 0.274 for placebo; the corresponding task contrasts are 0.665, 0.671, and 0.069. The four conditions share the question, answer interface, selected source count, canonical ordering rule, evidence representation, and token boundary. The complete rescue estimates are reported in Table~\ref{tab:support-rescue}.

\begin{table}[ht]
\centering
\caption{Route-support rescue estimates for 33 eligible pairs from 21 users with equal selected-source counts. Lower and upper columns give 95\% bootstrap limits with equal weight for users.}
\label{tab:support-rescue}
\setlength{\tabcolsep}{2pt}
\textbf{A. Support accuracy}\par\vspace{2pt}
\begin{tabular*}{\linewidth}{@{\extracolsep{\fill}}lrrr@{}}
\toprule
 & \multicolumn{3}{c}{Accuracy}\\
\cmidrule(lr){2-4}
\textbf{Condition} & \textbf{Estimate} & \textbf{Lower} & \textbf{Upper}\\
\midrule
Support present & 0.875 & 0.778 & 0.952\\
Support absent & 0.210 & 0.087 & 0.349\\
Support rescue & 0.881 & 0.790 & 0.952\\
Support placebo & 0.280 & 0.141 & 0.440\\
\bottomrule
\end{tabular*}
\par\smallskip
\textbf{B. Paired contrasts}\par\vspace{2pt}
\begin{tabular*}{\linewidth}{@{\extracolsep{\fill}}lrrrrrr@{}}
\toprule
 & \multicolumn{3}{c}{$H$} & \multicolumn{3}{c}{$\Delta$}\\
\cmidrule(lr){2-4}\cmidrule(lr){5-7}
\textbf{Contrast} & \textbf{Estimate} & \textbf{Lower} & \textbf{Upper} & \textbf{Estimate} & \textbf{Lower} & \textbf{Upper}\\
\midrule
Route contrast & 0.694 & 0.563 & 0.813 & 0.665 & 0.506 & 0.802\\
Rescue contrast & 0.724 & 0.605 & 0.833 & 0.671 & 0.496 & 0.817\\
Placebo contrast & 0.274 & 0.143 & 0.417 & 0.069 & $-0.077$ & 0.226\\
Rescue to present & 0.101 & 0.040 & 0.177 & 0.006 & $-0.044$ & 0.067\\
\bottomrule
\end{tabular*}
\end{table}

Table~\ref{tab:secondary-appendix} collects the remaining boundary comparisons.

\begin{table}[ht]
\centering
\caption{Secondary policy contrasts on complete 589-query matrices from 37 users. The overlap check uses inverse document frequency (IDF) weighting. Each contrast subtracts the route statistic for the second representation from that for the first: compressed minus canonical-compressed (overlap-IDF identity check), bounded-recent-3 minus bounded-source-3, utility-arrival-3 minus utility-source-3, and provenance-gated minus provenance-preserved. The columns report $\Gamma_H$ and $\Gamma_\Delta$ for disagreement and replay-minus-forward correctness change, respectively. For every row, the Estimate, Lower, and Upper columns report the equal-user-weighted point estimate and its 95\% user-clustered bootstrap limits.}
\label{tab:secondary-appendix}
\begin{tabular*}{0.98\textwidth}{@{\extracolsep{\fill}}lrrrrrr@{}}
\toprule
\textbf{Condition} & \multicolumn{3}{c}{\textbf{$\Gamma_H$}} & \multicolumn{3}{c}{\textbf{$\Gamma_\Delta$}}\\
\cmidrule(lr){2-4}\cmidrule(lr){5-7}
 & \textbf{Estimate} & \textbf{Lower} & \textbf{Upper} & \textbf{Estimate} & \textbf{Lower} & \textbf{Upper}\\
\midrule
Overlap-IDF check & 0.360 & 0.311 & 0.409 & 0.124 & 0.055 & 0.190\\
Recent window versus source & 0.302 & 0.256 & 0.349 & 0.071 & 0.013 & 0.128\\
Utility buffer versus source & 0.043 & $-0.002$ & 0.089 & $-0.004$ & $-0.039$ & 0.030\\
Provenance gate & 0.010 & $-0.032$ & 0.052 & $-0.013$ & $-0.049$ & 0.025\\
\bottomrule
\end{tabular*}
\end{table}

\subsection{Breadth of selection rules}
\label{app:additional-selection-breadth}
The additional conditions are an overlap-IDF identity check, bounded-window selection, utility-buffer selection, and a provenance instruction. The overlap-IDF identity-check row repeats the compressed-versus-canonical contrast under an alternative IDF-weighted retriever whose realized prompts, compiled contexts, retrieved-source lists, and budget-selected source identities match the corresponding BM25 rows; the two selection comparisons change selection before the capacity budget is applied; and the final comparison changes only the answer-stage instruction. 

Table~\ref{tab:full-pathways} reports source-set overlap and compiled-context length for each condition.

\begin{table}[ht]
\centering
\caption{Complete retained-context diagnostics on the complete 589-query matrix from 37 users. The Estimate, Lower, and Upper columns give the Jaccard similarity $J$ and its 95\% bootstrap confidence limits with equal user weight; length columns report character-derived context-length proxies under forward and replay order.}
\label{tab:full-pathways}
\begin{tabular*}{0.98\textwidth}{@{\extracolsep{\fill}}lrrrrr@{}}
\toprule
\textbf{Condition} & \multicolumn{3}{c}{\textbf{Jaccard similarity $J$}} & \multicolumn{2}{c}{\textbf{Context length}}\\
\cmidrule(lr){2-4}\cmidrule(lr){5-6}
 & \textbf{Estimate} & \textbf{Lower} & \textbf{Upper} & \textbf{Forward} & \textbf{Replay}\\
\midrule
Raw & 1.000 & 1.000 & 1.000 & 6562 & 6562\\
Canonical & 1.000 & 1.000 & 1.000 & 2766 & 2766\\
Arrival route & 0.045 & 0.009 & 0.081 & 2766 & 4181\\
Preserved & 1.000 & 1.000 & 1.000 & 6992 & 6992\\
Bounded recent (3) & 0.121 & 0.088 & 0.153 & 6716 & 6967\\
Bounded source (3) & 1.000 & 1.000 & 1.000 & 6868 & 6868\\
Utility by arrival (3) & 0.828 & 0.745 & 0.906 & 6363 & 6449\\
Utility by source (3) & 1.000 & 1.000 & 1.000 & 6428 & 6428\\
\bottomrule
\end{tabular*}
\end{table}

For each arm $b\in\{\mathrm{op},\mathrm{can}\}$, $z_b(\tau)$ is the equal-user single-draw route statistic comparing schedule $\tau$ with forward order within that arm. Here $z\in\{H,\Delta\}$, $\mathrm{op}$ is the varied operation arm, and $\mathrm{can}$ is source-order compaction. The operation contrast is $\delta_z^{\mathrm{op}}(\tau)=z_{\mathrm{op}}(\tau)-z_{\mathrm{can}}(\tau)$. Table~\ref{tab:full-mechanism-schedule} reports these operation-versus-canonical contrasts.

\begin{table}[ht]
\centering
\caption{Operation-versus-canonical contrasts $\delta_H^{\mathrm{op}}$ and $\delta_\Delta^{\mathrm{op}}$ by schedule on the complete 589-query matrix from 37 users. The Lower and Upper columns give 95\% bootstrap confidence limits with equal user weight.}
\label{tab:full-mechanism-schedule}
\begin{tabular*}{0.98\textwidth}{@{\extracolsep{\fill}}llrrrrrr@{}}
\toprule
\textbf{Path} & \textbf{Schedule} & \multicolumn{3}{c}{\textbf{$\delta_H^{\mathrm{op}}$}} & \multicolumn{3}{c}{\textbf{$\delta_\Delta^{\mathrm{op}}$}}\\
\cmidrule(lr){3-5}\cmidrule(lr){6-8}
\cmidrule(lr){3-3}\cmidrule(lr){4-5}\cmidrule(lr){6-6}\cmidrule(lr){7-8}
 &  & \textbf{Estimate} & \textbf{Lower} & \textbf{Upper} & \textbf{Estimate} & \textbf{Lower} & \textbf{Upper}\\
\midrule
Both operations & Replay & 0.360 & 0.303 & 0.420 & 0.171 & 0.094 & 0.251\\
Both operations & Half-swap & 0.261 & 0.202 & 0.327 & 0.132 & 0.054 & 0.211\\
Both operations & Odd-even & 0.203 & 0.129 & 0.279 & 0.077 & $-0.001$ & 0.162\\
\midrule
Survivor selection & Replay & 0.312 & 0.255 & 0.370 & 0.151 & 0.081 & 0.223\\
Survivor selection & Half-swap & 0.238 & 0.175 & 0.309 & 0.099 & 0.032 & 0.167\\
Survivor selection & Odd-even & 0.173 & 0.104 & 0.247 & 0.056 & $-0.018$ & 0.137\\
\midrule
Clustering only & Replay & $-0.013$ & $-0.051$ & 0.024 & 0.005 & $-0.032$ & 0.046\\
Clustering only & Half-swap & $-0.030$ & $-0.068$ & 0.005 & $-0.002$ & $-0.042$ & 0.040\\
Clustering only & Odd-even & $-0.024$ & $-0.064$ & 0.016 & $-0.010$ & $-0.043$ & 0.024\\
\bottomrule
\end{tabular*}
\end{table}

Table~\ref{tab:mechanism-state} reports the state-matching diagnostics for the operation matrix.

\begin{table}[ht]
\centering
\caption{State-layer diagnostics for the operation matrix on the complete 589-query matrix from 37 users. Survivor selection varies representatives with arrival order under a fixed partition; clustering only varies clustering, with representatives fixed to source order. Entries give equal-user-weighted estimates over 37 users; $J$ is source-set overlap and $C_{\mathrm{chg}}$ is the fraction of paired serialized contexts that differ, with $C_{\mathrm{chg},q}=\mathbf{1}[C_q^{(f)}\ne C_q^{(r)}]$.}
\label{tab:mechanism-state}
\setlength{\tabcolsep}{2pt}
\begin{tabular*}{\linewidth}{@{\extracolsep{\fill}}lrrrrrr@{}}
\toprule
 & \multicolumn{2}{c}{Replay} & \multicolumn{2}{c}{Half-swap} & \multicolumn{2}{c}{Odd-even}\\
\cmidrule(lr){2-3}\cmidrule(lr){4-5}\cmidrule(lr){6-7}
\textbf{Path} & \textbf{$J$} & \textbf{$C_{\mathrm{chg}}$} & \textbf{$J$} & \textbf{$C_{\mathrm{chg}}$} & \textbf{$J$} & \textbf{$C_{\mathrm{chg}}$}\\
\midrule
Survivor selection & 0.045 & 1.000 & 0.054 & 1.000 & 0.414 & 0.622\\
Clustering only & 0.973 & 0.054 & 0.986 & 0.027 & 0.986 & 0.027\\
\bottomrule
\end{tabular*}
\end{table}

\subsection{Operation-level dispersion}
\label{app:additional-operation-dispersion}

Reanalysis on the 63-query primary stratum yields survivor-selection contrasts $(\delta_H^{\mathrm{op}},\delta_\Delta^{\mathrm{op}})$ of $(0.531,0.356)$ under replay, $(0.310,0.037)$ under half-swap, and $(0.355,0.210)$ under odd-even; clustering-only contrasts are $(0.056,0.013)$, $(0.019,0.010)$, and $(-0.036,0.093)$. Table~\ref{tab:semantic-mechanism-dd} gives bootstrap intervals.
On the full 589-query matrix, the survivor-selection $\times$ clustering interaction in $H$ is $0.061\;[0.001,0.127]$, $0.053\;[-0.009,0.120]$, and $0.055\;[-0.001,0.111]$ under the three schedules; corresponding $\Delta$ interactions are $0.014\;[-0.045,0.072]$, $0.035\;[-0.011,0.081]$, and $0.032\;[-0.014,0.078]$.
Survivor selection changes the exposed source set for 63, 63, and 41 of 63 primary queries under the three schedules; clustering changes 4, 2, and 2.

\subsection{Selection and instruction controls}
\label{app:additional-selection-controls}
The bounded-window selection changes the retained records before the capacity budget is applied across the two orders, whereas the fixed source-order comparator selects the same records. The utility-buffer condition changes the retained set for 191 of 589 pools and the serialized context for 149; its fixed source-order comparator produces no corresponding changes. Provenance-preserved and gated-provenance contexts contain the same selected records and serialized context within each schedule.

Descriptive estimates by question stratum are reported in Table~\ref{tab:question-strata-main}.

Each gate appends a conflict instruction that names the benchmark index. The neutral arm asks the answer model to compare evidence without inferring recency from order. The placebo arm uses a displayed index that is unrelated to the benchmark index. The clean-accuracy comparison used a reference margin of two percentage points; the confidence interval for that comparison is reported directly. Under this comparison, the gated arm does not support an answer-stage repair.

\begin{table}[ht]
\centering
\caption{Operation-versus-canonical contrasts on the primary semantic stratum of 63 queries from 29 users. Lower and upper columns give 95\% bootstrap limits with equal weight for users.}
\label{tab:semantic-mechanism-dd}
\begin{tabular*}{0.98\textwidth}{@{\extracolsep{\fill}}llrrrrrr@{}}
\toprule
\textbf{Operation} & \textbf{Schedule} & \multicolumn{3}{c}{\textbf{$\delta_H^{\mathrm{op}}$}} & \multicolumn{3}{c}{\textbf{$\delta_\Delta^{\mathrm{op}}$}}\\
\cmidrule(lr){3-5}\cmidrule(lr){6-8}
 &  & \textbf{Estimate} & \textbf{Lower} & \textbf{Upper} & \textbf{Estimate} & \textbf{Lower} & \textbf{Upper}\\
\midrule
Survivor selection & Replay & 0.531 & 0.382 & 0.679 & 0.356 & 0.149 & 0.553\\
Survivor selection & Half-swap & 0.310 & 0.164 & 0.466 & 0.037 & $-0.138$ & 0.198\\
Survivor selection & Odd-even & 0.355 & 0.180 & 0.531 & 0.210 & 0.021 & 0.399\\
Clustering only & Replay & 0.056 & $-0.037$ & 0.172 & 0.013 & $-0.097$ & 0.117\\
Clustering only & Half-swap & 0.019 & $-0.092$ & 0.129 & 0.010 & $-0.106$ & 0.121\\
Clustering only & Odd-even & $-0.036$ & $-0.139$ & 0.069 & 0.093 & 0.007 & 0.203\\
\bottomrule
\end{tabular*}
\end{table}

Coordinates for the individual operations are shown in Table~\ref{tab:semantic-mechanism-direct}.

\begin{table}[ht]
\centering
\caption{Coordinates for each operation on 63 queries from 29 users. Both denotes the two operations following arrival order; Survivor denotes survivor selection following arrival order; and Cluster denotes clustering following arrival order with survivor choice fixed to source order. Changed is an integer count of pairs with different exposed source sets; the denominator is 63 in every row. Entries weight users equally; lower and upper columns are 95\% bootstrap limits.}
\label{tab:semantic-mechanism-direct}
\setlength{\tabcolsep}{2.2pt}
\begin{tabular*}{\textwidth}{@{\extracolsep{\fill}}llrrrrrrr@{}}
\toprule
\textbf{Path} & \textbf{Schedule} & \multicolumn{3}{c}{\textbf{$H$}} & \multicolumn{3}{c}{\textbf{$\Delta$}} & \textbf{Changed}\\
\cmidrule(lr){3-5}\cmidrule(lr){6-8}
 &  & \textbf{Estimate} & \textbf{Lower} & \textbf{Upper} & \textbf{Estimate} & \textbf{Lower} & \textbf{Upper} & \\
\midrule
Both & Replay & 0.603 & 0.460 & 0.740 & 0.390 & 0.183 & 0.583 & 63\\
Both & Half-swap & 0.350 & 0.210 & 0.496 & 0.011 & $-0.161$ & 0.180 & 63\\
Both & Odd-even & 0.425 & 0.278 & 0.578 & 0.182 & 0.005 & 0.357 & 43\\
Survivor & Replay & 0.558 & 0.407 & 0.706 & 0.347 & 0.140 & 0.540 & 63\\
Survivor & Half-swap & 0.367 & 0.231 & 0.511 & 0.028 & $-0.149$ & 0.197 & 63\\
Survivor & Odd-even & 0.420 & 0.278 & 0.567 & 0.202 & 0.020 & 0.383 & 41\\
Cluster & Replay & 0.103 & 0.017 & 0.224 & 0.000 & $-0.103$ & 0.103 & 4\\
Cluster & Half-swap & 0.135 & 0.034 & 0.250 & $-0.003$ & $-0.115$ & 0.109 & 2\\
Cluster & Odd-even & 0.115 & 0.029 & 0.224 & 0.080 & $-0.017$ & 0.195 & 2\\
\bottomrule
\end{tabular*}
\end{table}

\section{Extended Related Work}
\label{app:related}

\subsection{Long-horizon memory evaluation}
\label{app:related-long-horizon}
MemGPT~\citep{packer2023memgpt} and MemoryBank~\citep{zhong2023memorybank} make persistent memory storage explicit. LoCoMo~\citep{maharana-etal-2024-evaluating} evaluates long-term conversational memory, while LoCoMo-Plus~\citep{li-etal-2026-locomo} evaluates cognitive memory under semantic cue--trigger disconnect. LongMemEval~\citep{wu2025longmemeval}, MemoryAgentBench~\citep{hu2026memoryagentbench}, and AMemGym~\citep{ji2026amemgym} collectively evaluate long-horizon memory through extraction, updating, forgetting, and interactive state, with the latter two emphasizing incremental or on-policy interactions. MemoryArena~\citep{he2026memoryarena} couples memory to interdependent actions. AMA-Bench~\citep{ama-bench2026} and STATE-Bench~\citep{statebench2026} extend memory evaluation to agentic applications and stateful workflows. LightMem~\citep{fang2026lightmem} studies lightweight and efficient memory-augmented generation, whereas EMBER~\citep{li2026ember} studies budgeted evidence retention. MemSkill~\citep{zhang2026memskill} examines evolving memory skills; the experience-following study~\citep{wang2026experience} examines how memory management affects agent behavior; and scale-conditioned memory evaluation~\citep{shao2026scaleconditioned} examines evidence usability across model scales. The Agent Memory Benchmark~\citep{amb2026} covers tools, documents, and preferences, while PersonaMem~\citep{personamem2025} studies user profiles.

\subsection{Order and construction effects}
\label{app:related-order-construction}
Adaptive Chameleon~\citep{xie2024adaptive} studies how language models respond to external evidence under knowledge conflict. Lost in the Middle~\citep{liu2024lostmiddle} and Ordered Prompts~\citep{lu2022ordered} document sensitivity to presentation order. Stable-RAG~\citep{zhang-etal-2026-stable} extends the analysis to retrieval permutations, while SegTreeMem~\citep{liu2026segtreemem} studies temporal order during memory construction. Useful Memories Become Faulty~\citep{zhang2026faultymemory} examines consolidation failures and source coverage in memory construction.

\subsection{Memory structure, updates, and provenance}
\label{app:related-structure-provenance}
MemTree~\citep{rezazadeh2024memtree} uses hierarchical structure; Hindsight~\citep{latimer2026hindsight} separates source-backed facts from evolving beliefs; Compiled Memory~\citep{rhodes2026compiled} promotes verified experience into instructions; A-MEM~\citep{xu2025amem} maintains evolving links; and MemCon~\citep{jiang2026memcon} learns adaptive memory-management policies. GovMem~\citep{qi2026govmem} and MemTxn~\citep{cui2026memtxn} make governance and recovery explicit, while Louck's origin-bound memory-security work~\citep{liu2026tmanm} treats source authority as a security property. Ground Truth First~\citep{spencer2026groundtruth} studies evaluation grounded in verifiable evidence, \citet{wang2026tracetrust} survey evidence tracing and execution provenance, and Selection Integrity~\citep{fei2026selectionintegrityllmgraph} studies authenticated source selection. Remembering More, Risking More~\citep{altawaha2026remembering}, STALE~\citep{chao2026stale}, Supersede~\citep{patel2026supersede}, and MemEvoBench~\citep{zhang2026memevobench} address memory-related risk, stale evidence, supersession, and memory evolution. \citet{reddy2026freshness} instead examine post-retrieval assembly by separating evidence extraction from policy execution. AttriMem~\citep{li2026attrimem} uses answer attribution to provide process feedback for memory construction.

\section{Comparison Settings}
\label{app:scope}

Table~\ref{tab:claim-boundaries} lists the objects held fixed in each comparison. Table~\ref{tab:comparison-planes} compares the interventions and observations used in related evaluations.

\begin{table}[ht]
\centering
\caption{Objects held fixed and interpretations of the paired comparisons. Restoration changes source identities while preserving the pool and exposed source count.}
\label{tab:claim-boundaries}
\renewcommand{\arraystretch}{0.84}
\begin{tabular*}{0.98\textwidth}{@{\extracolsep{\fill}}>{\raggedright\arraybackslash}p{0.55in} >{\raggedright\arraybackslash}p{0.42in} >{\raggedright\arraybackslash}p{0.42in} >{\raggedright\arraybackslash}p{1.05in} >{\raggedright\arraybackslash}p{2.15in}@{}}
\toprule
\textbf{Question} & \multicolumn{3}{c}{\textbf{Fixed object}} & \textbf{Interpretation}\\
\cmidrule(lr){2-4}
 & \textbf{User} & \textbf{Query} & \textbf{Evidence} & \\
\midrule
Order & same & same & Source pool & Changes in exposed sources and contexts\\
Operation & same & same & Source pool & Contributions of clustering and survivor selection\\
Support & same & same & Pool and source count & Task value of the displaced support record\\
Answers & same & same & Prompt and context & Variation within the answer model\\
Reuse & same & new & Constructed state & Effects on another question from the same user\\
\bottomrule
\end{tabular*}
\end{table}

Storage permutation, temporal reversal, and retrieval perturbation act on different objects. A temporal reversal can change the intended answer, while a storage permutation of independent records preserves it. A comparison that records answers alone measures response stability but cannot locate changes in the memory state.

\begin{table}[ht]
\centering
\caption{Comparison planes for adjacent order and memory evaluations. Columns separate the fixed object, route boundary, observable state, state--task link, and endpoint. StateAud., Selection Int., and Faulty denote StateAuditor, Selection Integrity, and Useful Memories Become Faulty, respectively.}
\label{tab:comparison-planes}
\setlength{\tabcolsep}{2.5pt}
\begin{tabular*}{0.98\textwidth}{@{\extracolsep{\fill}}>{\raggedright\arraybackslash}p{0.74in} >{\raggedright\arraybackslash}p{0.52in} >{\raggedright\arraybackslash}p{0.78in} >{\raggedright\arraybackslash}p{0.95in} >{\raggedright\arraybackslash}p{1.15in} >{\raggedright\arraybackslash}p{0.86in}@{}}
\toprule
\textbf{Study} & \multicolumn{2}{c}{\textbf{Comparison}} & \multicolumn{2}{c}{\textbf{Observation}} & \textbf{Endpoint}\\
\cmidrule(lr){2-3}\cmidrule(lr){4-5}
 & \textbf{Fixed object} & \textbf{Route boundary} & \textbf{Observable state} & \textbf{State--task link} & \\
\midrule
Stable-RAG & Retrieved set & Prompt order & Hidden state & Prompt and answer & Stability\\
SegTreeMem & Events & Build order & Segment tree & State and utility & Utility\\
StateAud. & Trace & Update path & Recorded state & State and repair & Repair\\
Faulty & Trace & Consolidation & Text state & Text and utility & Consolidation\\
AttriMem & Policy & Attribution & Token attribution & Score and learning & Learning\\
\midrule
AcquaBench & Task & Target substitution & Gold support & Support and utility & Accuracy\\
Selection Int. & Graph state & Selection rule & Source lineage & Accumulability & Selection\\
\midrule
RoutePrism & Source pool & Task-preserving build order & Sources and contexts & Matched support restoration & Task effect\\
\bottomrule
\end{tabular*}
\end{table}

\subsection{Adjacent evaluation boundaries}
\label{app:scope-adjacent-boundaries}
Stable-RAG~\citep{zhang-etal-2026-stable} permutes retrieved documents before generation, SegTreeMem~\citep{liu2026segtreemem} changes the construction order of a temporal segment structure, and StateAuditor~\citep{sun2026stateauditor} checks whether a recorded state can support repair during updates. AcquaBench~\citep{luo2026acquabench} holds the task fixed while comparing clean, gold, and sham substitutions.

\subsection{Fixed-content display intervention (M-R and M-D)}
\label{app:scope-display-intervention}
Each of the 126 retained MemoChat contexts contains four summary blocks, four retrieved-dialog blocks, and two recent-dialog blocks. Reversing blocks within these three groups preserves their text, headings, and internal sentence order. Native prompts are recovered exactly before padding, and all four construction--layout cells for each question have equal token counts under \texttt{o200k\_base}. Two fresh Luna max answers per cell give 504 answers. For each construction state, we compute cross-layout disagreement minus mean within-layout disagreement, then average states within questions and questions within users. Construction contrasts compare the two states at each fixed layout; M-R averages these contrasts over the two layouts. Its signed correctness contrast averages the replay-minus-forward difference in the same cells, with user aggregation and intervals computed jointly from the four-cell records. All intervals use 10,000 paired user-bootstrap resamples with seed 20260910.

\begin{table}[htbp]
\centering\small
\caption{Fresh MemoChat residual contrasts. Every row uses 63 queries and 29 users; Lower and Upper are 95\% limits. The final row is a paired difference, computed before bootstrap resampling.}
\label{tab:memochat-display}
\begin{tabular*}{\linewidth}{@{\extracolsep{\fill}}lrrr@{}}
\toprule
\textbf{Comparison} & \textbf{Estimate} & \textbf{Lower} & \textbf{Upper}\\
\midrule
Display & 0.013 & -0.010 & 0.040\\
Construction, native layout & 0.089 & 0.008 & 0.195\\
Construction, reversed layout & 0.029 & 0.000 & 0.072\\
Construction, mean layouts & 0.059 & 0.016 & 0.112\\
Construction minus display & 0.046 & 0.004 & 0.101\\
\bottomrule
\end{tabular*}
\end{table}

Across construction states, both summary-content and dialog-content multisets change for every query, even though block counts agree. Source identifiers do not identify unchanged compiled fragments. The fixed-content intervention tests this specific display reversal; the construction contrast also includes changed fragments and summaries and does not identify a separate additive semantic-content effect. These fresh answers are analyzed separately from the seven-draw M-C matrix.

\section{Answer Boundary Controls}
\label{app:answer}

\subsection{Support-present and restored prompts}
\label{app:answer-support-prompts}
We compare the complete prompt texts after deduplicating repeated answer draws (Table~\ref{tab:restoration-prompt-identity}). For most focal pairs, returning the displaced record recreates the support-present input exactly. Present and Restored then provide repeated measurements at the same input; the comparison with Absent and Replacement measures the effect of changing the supplied record. Recent-3's support-present and restored prompts differ throughout this matrix, so its support-present condition is a separate input reference.
\begin{table}[htbp]
\centering\small
\caption{Exact full-prompt identity between Present and Restored. Prompt pairs counts query--padding combinations: two per focal query and one per recent-policy query. Identical counts equal prompt texts; repeated answer draws do not add prompt pairs.}
\label{tab:restoration-prompt-identity}
\begin{tabular*}{\linewidth}{@{\extracolsep{\fill}}lrrrr@{}}
\toprule
\textbf{Policy} & \textbf{Queries} & \textbf{Users} & \textbf{Prompt pairs} & \textbf{Identical}\\
\midrule
Focal & 33 & 21 & 66 & 62\\
Recent-3 & 36 & 18 & 36 & 0\\
\bottomrule
\end{tabular*}
\end{table}

\subsection{Fresh-answer sensitivity}
\label{app:answer-fresh-sensitivity}
We exclude the original formal answers and recompute the residual and correctness contrast using the remaining fresh calls (Table~\ref{tab:fresh-only}). Focal inputs retain canonical serialization and their original padding; MemoChat retains its compiled contexts. Removing the formal answers preserves the positive focal residual and correctness contrast. MemoChat retains a positive residual with an uncertain signed correctness change. This comparison addresses the mixture of collection batches while keeping the realized prompts fixed.
\begin{table}[htbp]
\centering\small
\caption{Fresh-answer estimates for 63 queries from 29 users in every row. $m$ is the number of fresh draws per route and prompt. F-EA/F-EB and R-EA/R-EB denote focal/recent traces with padding A/B; A and B retain the original padding variants. Lower and Upper are 95\% intervals from 10,000 user-bootstrap resamples with seed 2026091011.}
\label{tab:fresh-only}
\setlength{\tabcolsep}{3pt}
\begin{tabular*}{\linewidth}{@{\extracolsep{\fill}}lrrrrrrr@{}}
\toprule
 & & \multicolumn{3}{c}{$D_{\mathrm{res}}$} & \multicolumn{3}{c}{$\Delta_{\mathrm{rep}}$}\\
\cmidrule(lr){3-5}\cmidrule(lr){6-8}
\textbf{Trace} & \textbf{$m$} & \textbf{Estimate} & \textbf{Lower} & \textbf{Upper} & \textbf{Estimate} & \textbf{Lower} & \textbf{Upper}\\
\midrule
F-EA & 2 & 0.517 & 0.346 & 0.683 & 0.426 & 0.241 & 0.606\\
F-EB & 2 & 0.538 & 0.393 & 0.686 & 0.327 & 0.112 & 0.533\\
M-C & 6 & 0.039 & 0.004 & 0.092 & 0.038 & -0.017 & 0.112\\
\bottomrule
\end{tabular*}
\end{table}

\begin{table}[htbp]
\centering
\small
\caption{Profiles for exact-token traces F-E (focal) and R-E (Recent-3). A and B denote padding with \texttt{one} and \texttt{x}, respectively. $H_1$ and $\Delta_1$ use one formal answer per route; repeated estimates combine that answer with two fresh draws per route. Estimates weight users equally. Every row contains 63 query pairs from 29 users.}
\label{tab:direct-route-details}
\setlength{\tabcolsep}{2pt}
\begin{tabular*}{\linewidth}{@{\extracolsep{\fill}}lrrrrrrr@{}}
\toprule
 & \multicolumn{3}{c}{Direct profile} & \multicolumn{4}{c}{Repeated answers}\\
\cmidrule(lr){2-4}\cmidrule(lr){5-8}
\textbf{Trace} & \textbf{$J$} & \textbf{$H_1$} & \textbf{$\Delta_1$} & \textbf{$H_{\times}$} & \textbf{$H_{\mathrm{within}}$} & \textbf{$D_{\mathrm{res}}$} & \textbf{$\Delta_{\mathrm{rep}}$}\\
\midrule
F-EA & 0.057 & 0.631 & 0.480 & 0.630 & 0.101 & 0.529 & 0.444\\
F-EB & 0.057 & 0.528 & 0.394 & 0.593 & 0.080 & 0.513 & 0.349\\
R-EA & 0.166 & 0.564 & 0.071 & 0.586 & 0.093 & 0.493 & 0.028\\
R-EB & 0.166 & 0.534 & $-0.017$ & 0.527 & 0.144 & 0.383 & $-0.002$\\
\bottomrule
\end{tabular*}
\end{table}

\begin{table}[ht]
\centering
\caption{Exact-token matching on the primary semantic stratum of 63 queries from 29 users. Estimates give equal weight to users; lower and upper columns give 95\% user-bootstrap limits.}
\label{tab:exact-token-control}
\setlength{\tabcolsep}{2pt}
\begin{tabular*}{\linewidth}{@{\extracolsep{\fill}}rrrrrr@{}}
\toprule
\multicolumn{3}{c}{$H$} & \multicolumn{3}{c}{$\Delta$}\\
\cmidrule(lr){1-3}\cmidrule(lr){4-6}
\textbf{Estimate} & \textbf{Lower} & \textbf{Upper} & \textbf{Estimate} & \textbf{Lower} & \textbf{Upper}\\
\midrule
0.538 & 0.387 & 0.683 & 0.336 & 0.138 & 0.528\\
\bottomrule
\end{tabular*}
\end{table}

\subsection{Exact-token focal trace (F-E)}
\label{app:answer-exact-token}
An exact-token-matched trace with canonical display provides a direct answer-level control for the primary semantic stratum. It contains 252 rows over 63 queries and 29 users, with two route orders and two padding forms; every route pair has equal token counts, the same question and answer suffix, preserved evidence bodies, and different selected source sets. All paired rows pass the structural audit, with Luna max held fixed across the matrix. The two padding forms give $(H,\Delta)=(0.631,0.480)$ and $(0.528,0.394)$; corresponding intervals with equal user weight are reported in Table~\ref{tab:direct-trace}. The main exact-token control gives $(H,\Delta)=(0.538,0.336)$ on the same semantic stratum. Both padding variants enforce pairwise token equality under the declared tokenizer; because their filler realizations differ, we treat them as two matched controls rather than one pooled estimate. Repeated answer estimands for this trace appear in Table~\ref{tab:exact-repeat-estimands}.

\begin{table}[ht]
\centering
\caption{Directly retained equal token route trace with canonical display. Lower and upper columns give 95\% bootstrap limits with equal weight for users; both padding rows contain 63 query pairs from 29 users, and all 63 pairs have different exposed source sets.}
\label{tab:direct-trace}
\setlength{\tabcolsep}{2.0pt}
\begin{tabular*}{\textwidth}{@{\extracolsep{\fill}}lrrrrrr@{}}
\toprule
\textbf{Padding} & \multicolumn{3}{c}{\textbf{$H$}} & \multicolumn{3}{c}{\textbf{$\Delta$}}\\
\cmidrule(lr){2-4}\cmidrule(lr){5-7}
 & \textbf{Estimate} & \textbf{Lower} & \textbf{Upper} & \textbf{Estimate} & \textbf{Lower} & \textbf{Upper} \\
\midrule
F-EA & 0.631 & 0.484 & 0.772 & 0.480 & 0.286 & 0.657\\
F-EB & 0.528 & 0.380 & 0.670 & 0.394 & 0.214 & 0.564\\
\bottomrule
\end{tabular*}
\end{table}

\begin{table}[ht]
\centering
\caption{Repeated-answer estimands for focal exact-token trace F-E (one formal and two fresh answers per route). Lower and upper columns give 95\% bootstrap limits with equal weight for users.}
\label{tab:exact-repeat-estimands}
\setlength{\tabcolsep}{2pt}
\textbf{A. Response disagreement}\par\vspace{2pt}
\begin{tabular*}{\linewidth}{@{\extracolsep{\fill}}lrrrrrrrrr@{}}
\toprule
 & \multicolumn{3}{c}{$H_{\times}$} & \multicolumn{3}{c}{$H_{\mathrm{within}}$} & \multicolumn{3}{c}{$D_{\mathrm{res}}$}\\
\cmidrule(lr){2-4}\cmidrule(lr){5-7}\cmidrule(lr){8-10}
\textbf{Trace} & \textbf{Estimate} & \textbf{Lower} & \textbf{Upper} & \textbf{Estimate} & \textbf{Lower} & \textbf{Upper} & \textbf{Estimate} & \textbf{Lower} & \textbf{Upper}\\
\midrule
F-EA & 0.630 & 0.497 & 0.754 & 0.101 & 0.039 & 0.175 & 0.529 & 0.361 & 0.690\\
F-EB & 0.593 & 0.459 & 0.720 & 0.080 & 0.034 & 0.137 & 0.513 & 0.374 & 0.652\\
\bottomrule
\end{tabular*}
\par\smallskip
\textbf{B. Distribution and correctness change}\par\vspace{2pt}
\begin{tabular*}{\linewidth}{@{\extracolsep{\fill}}lrrrrrr@{}}
\toprule
 & \multicolumn{3}{c}{$\widehat D^\star$} & \multicolumn{3}{c}{$\Delta_{\mathrm{rep}}$}\\
\cmidrule(lr){2-4}\cmidrule(lr){5-7}
\textbf{Trace} & \textbf{Estimate} & \textbf{Lower} & \textbf{Upper} & \textbf{Estimate} & \textbf{Lower} & \textbf{Upper}\\
\midrule
F-EA & 0.563 & 0.414 & 0.714 & 0.444 & 0.261 & 0.612\\
F-EB & 0.540 & 0.402 & 0.673 & 0.349 & 0.153 & 0.539\\
\bottomrule
\end{tabular*}
\end{table}

Correctness transitions for the two matched realizations are listed in Table~\ref{tab:exact-transitions}.

\begin{figure}[htbp]
\centering
\includegraphics[width=\linewidth]{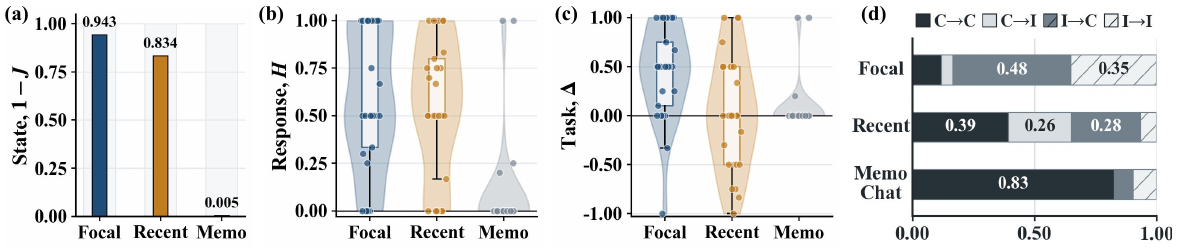}
\caption{Single-draw route profiles for the focal compactor, bounded recent policy, and MemoChat. Panels (a)--(c) show source change, $H_1$, and $\Delta_1$; panel (d) shows correctness transitions, with C denoting correct and I incorrect. Focal and Recent combine the two padding variants within users. Dots denote user means; boxes show the median and interquartile range, and whiskers extend to observations within 1.5 interquartile ranges.}
\label{fig:primary-profile}
\end{figure}

\subsection{Endpoint versus support}
\label{app:answer-endpoint-support}
Applying the support annotations, collected before route outputs and answer outcomes, to the same retained trace identifies 15 route and padding comparisons from 10 users in which both routes returned the same correct option. All 15 also replaced the source set that directly supports the gold option, so the direct-support overlap was zero. This diagnostic on fixed traces shows why endpoint correctness does not determine answer-relevant provenance; it is descriptive evidence for the state distinction rather than a population estimate.

Totals weighted by query in the same retained trace are $\Delta=0.429$ and $\Delta=0.397$ for padding forms A and B, respectively; including answer changes that preserve correctness gives $H=0.556$ and $H=0.508$. The main table uses the macro estimand with equal weight for users, so these totals provide an aggregation-sensitivity check rather than an additional estimate.

\subsection{Submission-order balance}
\label{app:answer-submission-balance}
Balanced submission order supplies a separate Luna max control. It contains 126 pairs and 252 fresh answers, with each route submitted first in 63 pairs. Combined estimates are $H=0.569$ and $\Delta=0.374$. The dispatch-order interaction between chronological-first and replay-first runs is $-0.007$ [$-0.162,0.156$] for $H$ and $0.108$ [$-0.052,0.273$] for $\Delta$, using bootstrap intervals with equal user weight on the 22 users represented in both strata. The route contrast persists after submission position is balanced, whereas the position interaction is compatible with zero.

\begin{table}[ht]
\centering
\caption{Correctness transitions in the directly retained equal token trace with canonical display. Column groups fix forward correctness; subcolumns give replay correctness.}
\label{tab:exact-transitions}
\setlength{\tabcolsep}{2pt}
\begin{tabular*}{\linewidth}{@{\extracolsep{\fill}}lrrrr@{}}
\toprule
 & \multicolumn{2}{c}{Forward correct} & \multicolumn{2}{c}{Forward incorrect}\\
\cmidrule(lr){2-3}\cmidrule(lr){4-5}
\textbf{Padding} & \textbf{Replay correct} & \textbf{Replay incorrect} & \textbf{Replay correct} & \textbf{Replay incorrect}\\
\midrule
Matched A & 8 & 2 & 29 & 24\\
Matched B & 7 & 2 & 27 & 27\\
\bottomrule
\end{tabular*}
\end{table}

All retained answer rows use gpt-5.6-luna at max effort with the fixed categorical answer schema. The principal matrix contains 8,246 answer cells from 589 queries, seven representations, and two orders. The matched prompt control contains 400 prompt pairs, 800 answer rows, 100 queries selected by a rule stratified by user, and 37 users. The query list is fixed before the fresh answer draws and uses no route response, correctness label, or state change statistic. It records matched-prompt answer flips and binds schedule disagreement and same-prompt answer variation to the same user contexts; the corresponding rate is reported in Table~\ref{tab:paired-contrasts}.

A matched context realization check contains 2,356 answer rows. A paired-input check confirms that prompts, contexts, and source identities are identical in all 2,356 rows compared with the primary execution. The recovered $H$ and $\Delta$ contrasts provide an independent check of answer realization under those fixed contexts. This check leaves retrieval unchanged. Policy contrasts on users held out from fitting are reported in Table~\ref{tab:crossfit-policy}.

For schedule $\tau$, define the policy contrast as $\Gamma_z(\tau)=z_{\mathrm{ord}}(\tau)-z_{\mathrm{can}}(\tau)$ for $z\in\{H,\Delta,D_{\mathrm{res}}\}$, where ord denotes the arrival-order streaming policy and can its source-order comparator.

\begin{table}[ht]
\centering
\caption{Policy contrasts on users held out from fitting. Lower and upper columns give 95\% bootstrap limits with equal weight for users.}
\label{tab:crossfit-policy}
\begin{tabular*}{0.98\textwidth}{@{\extracolsep{\fill}}lrrrrrr@{}}
\toprule
\textbf{Schedule} & \multicolumn{3}{c}{\textbf{$\Gamma_H$}} & \multicolumn{3}{c}{\textbf{$\Gamma_{\Delta}$}}\\
\cmidrule(lr){2-4}\cmidrule(lr){5-7}
 & \textbf{Estimate} & \textbf{Lower} & \textbf{Upper} & \textbf{Estimate} & \textbf{Lower} & \textbf{Upper}\\
\midrule
Replay & $-0.017$ & $-0.055$ & 0.020 & $-0.034$ & $-0.068$ & $-0.003$\\
Half-swap & $-0.007$ & $-0.044$ & 0.029 & $-0.047$ & $-0.080$ & $-0.015$\\
Odd-even & 0.000 & $-0.032$ & 0.037 & $-0.049$ & $-0.077$ & $-0.020$\\
\bottomrule
\end{tabular*}
\end{table}

\begin{figure}[ht]
\centering
\includegraphics[width=0.92\textwidth]{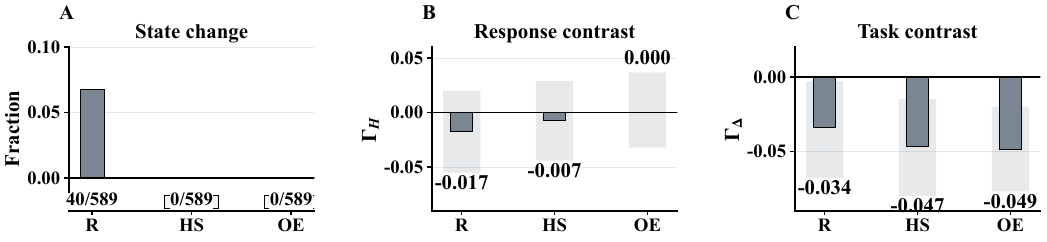}
\caption{Route policy contrasts across replay (R), half-swap (HS), and odd-even (OE) for users held out from fitting. Panels show the fraction of changed selected states, $\Gamma_H$, and $\Gamma_{\Delta}$; shaded intervals give 95\% limits with equal weight for users.}
\label{fig:crossfit-policy-app}
\end{figure}

\subsection{Policy matrix for held-out users}
\label{app:answer-policy-matrix}
The cross-fit extension contains $589\times2\times4=4{,}712$ valid Luna max answer rows. It combines two policy representations with chronological, replay, half-swap, and odd-even schedules. The arrival arm changes its retained state on 51 of 589 query pairs and its source set exposed to the answer on 40, whereas the source stable arm changes neither. The retained-state count records the construction boundary; the exposed-state count records the state that can affect the answer. The full replay contrast is $\Gamma_H=-0.017$ [$-0.055,0.020$] and $\Gamma_{\Delta}=-0.034$ [$-0.068,-0.003$], with the schedule decomposition in Table~\ref{tab:crossfit-policy}.

\begin{figure}[ht]
\centering
\includegraphics[width=0.88\textwidth]{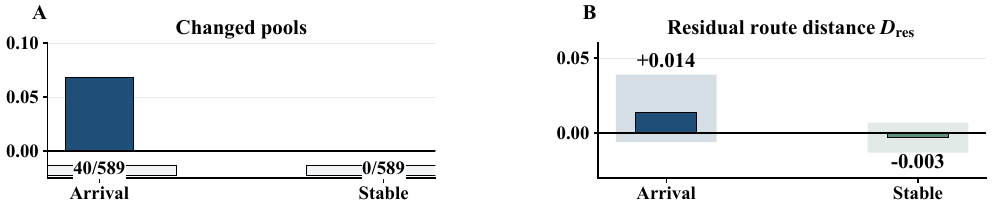}
\caption{Repeated answer decomposition for route policies evaluated on held-out users. Bars show the fraction of pools with changed selected states and the residual route estimate after removing answer variation within a prompt; intervals are in Table~\ref{tab:repeat-decomposition}.}
\label{fig:crossfit-repeat}
\end{figure}

The cross-fit traces also enable an answer-stage check. The source stable arm presents the same context across its paired schedules, so its answer variation provides the fixed state reference for the arrival arm. After subtracting that variation within a prompt, the residual estimates are $0.014$ and $-0.003$ for the arrival and source-stable arms. Figure~\ref{fig:crossfit-policy-app} and Figure~\ref{fig:crossfit-repeat} display the exposed state rate by schedule and the residual decomposition; Table~\ref{tab:repeat-decomposition} gives the corresponding intervals.

\subsection{Support-layer audit}
\label{app:answer-support-audit}
On the 146-query fact-recall set, the arrival policy changes the source set exposed to the answer in 15 pairs. Nine of these changes also alter the identity of the directly supporting source, and two remove direct support from the exposed set; one of those two pairs changes the answer and correctness. This analysis includes pairs selected for a source-set change.

\begin{table}[ht]
\centering
\small
\caption{Support-layer localization for the cross-fit policy on held-out users. Counts use the same 146 query cards and support labels collected without outcome information.}
\label{tab:crossfit-support}
\begin{tabular}{lrrrrrr}
\toprule
\textbf{Case} & \multicolumn{2}{c}{\textbf{Coverage}} & \multicolumn{3}{c}{\textbf{State change}} & \textbf{Task}\\
\cmidrule(lr){2-3}\cmidrule(lr){4-6}\cmidrule(lr){7-7}
 & \textbf{Pools} & \textbf{Users} & \textbf{Visible} & \textbf{Support} & \textbf{Loss} & \textbf{Answer}\\
\midrule
All audited & 146 & 37 & 15 & 9 & 2 & 1\\
Visible change & 15 & 3 & 15 & 9 & 2 & 1\\
Direct support change & 9 & 3 & 9 & 9 & 2 & 1\\
Support loss & 2 & 2 & 2 & 2 & 2 & 1\\
\bottomrule
\end{tabular}
\end{table}

\begin{table}[ht]
\centering
\caption{Repeated answer route decomposition. Lower and upper columns give 95\% bootstrap limits with equal weight for users.}
\label{tab:repeat-decomposition}
\begin{tabular*}{0.98\textwidth}{@{\extracolsep{\fill}}lrrrr@{}}
\toprule
\textbf{Policy} & \textbf{Changed fraction} & \textbf{Estimate} & \textbf{Lower} & \textbf{Upper}\\
\midrule
Arrival route & 0.068 & 0.014 & $-0.006$ & 0.039\\
Source stable & 0.000 & $-0.003$ & $-0.013$ & 0.007\\
Arrival minus source & 0.068 & 0.017 & $-0.004$ & 0.043\\
\bottomrule
\end{tabular*}
\end{table}

\subsection{Canonical serialization control}
\label{app:answer-canonical-serialization}
Canonical serialization presents the selected source identities in a fixed source order for both construction routes, preserving the source multiset and evidence bodies. Its 1,178 Luna max answers give $H=0.145$ [$0.111,0.181$] and $\Delta=-0.029$ [$-0.057,-0.002$] over all 589 paired queries. The contrast concentrates on the 40 pools whose selected source sets differ, where $H=0.301$ [$0.118,0.400$] and $\Delta=-0.092$ [$-0.200,0.000$]; these pools come from three users with 17, 13, and 10 changed queries. The 549 pools with unchanged selection span all 37 users and yield $H=0.125$ [$0.093,0.158$] and $\Delta=-0.023$ [$-0.050,0.003$]. The unchanged-source group uses identical prompts after canonicalization.

To test whether this contrast depends on a single answer draw, we drew two additional answers per route for the 40 pairs with changed selection while keeping every canonicalized prompt fixed. Including the formal answer as draw 0 gives three draws per route. Across this finite stratum, the raw disagreement across routes is $0.368$, the answer disagreement within a route is $0.139$, and the residual route estimate is $0.229$; the corresponding signed correctness change is $-0.058$. The 31 changed pairs without a temporal-order requirement retain a residual distance using only fresh draws of $0.234$ [$0.111,0.300$], and the 15 exchangeable changed pairs retain $0.164$ [$0.125,0.200$]. These are conditional sensitivities from three users; the full intervals with equal weight for users and the formal single-draw comparison are given in Table~\ref{tab:canonical-repeat}.

Canonical serialization route contrasts are reported in Table~\ref{tab:canonical-serialization}.

\begin{table}[ht]
\centering
\caption{Canonical serialization route contrasts. Lower and upper columns give 95\% bootstrap limits with equal weight for users.}
\label{tab:canonical-serialization}
\begin{tabular*}{0.98\textwidth}{@{\extracolsep{\fill}}lrrrrrr@{}}
\toprule
\textbf{Subset} & \multicolumn{3}{c}{\textbf{$H$}} & \multicolumn{3}{c}{\textbf{$\Delta$}}\\
\cmidrule(lr){2-4}\cmidrule(lr){5-7}
 & \textbf{Estimate} & \textbf{Lower} & \textbf{Upper} & \textbf{Estimate} & \textbf{Lower} & \textbf{Upper}\\
\midrule
All queries & 0.145 & 0.111 & 0.181 & $-0.029$ & $-0.057$ & $-0.002$\\
Selection changed (40) & 0.301 & 0.118 & 0.400 & $-0.092$ & $-0.200$ & 0.000\\
Selection unchanged (549) & 0.125 & 0.093 & 0.158 & $-0.023$ & $-0.050$ & 0.003\\
\bottomrule
\end{tabular*}
\end{table}

\begin{table}[ht]
\centering
\caption{Fresh answer extension on the 40 canonical pairs whose selected source sets differ. Lower and upper columns give 95\% bootstrap limits with equal weight for users.}
\label{tab:canonical-repeat}
\setlength{\tabcolsep}{2pt}
\textbf{A. Response disagreement}\par\vspace{2pt}
\begin{tabular*}{\linewidth}{@{\extracolsep{\fill}}rrrrrr@{}}
\toprule
\multicolumn{3}{c}{$H_{\times}^{(m)}$} & \multicolumn{3}{c}{$H_{\mathrm{within}}^{(m)}$}\\
\cmidrule(lr){1-3}\cmidrule(lr){4-6}
\textbf{Estimate} & \textbf{Lower} & \textbf{Upper} & \textbf{Estimate} & \textbf{Lower} & \textbf{Upper}\\
\midrule
0.368 & 0.144 & 0.533 & 0.139 & 0.088 & 0.200\\
\bottomrule
\end{tabular*}
\par\smallskip
\textbf{B. Residual and correctness change}\par\vspace{2pt}
\begin{tabular*}{\linewidth}{@{\extracolsep{\fill}}rrrrrr@{}}
\toprule
\multicolumn{3}{c}{$D_{\mathrm{res}}$} & \multicolumn{3}{c}{$\Delta_{\mathrm{rep}}$}\\
\cmidrule(lr){1-3}\cmidrule(lr){4-6}
\textbf{Estimate} & \textbf{Lower} & \textbf{Upper} & \textbf{Estimate} & \textbf{Lower} & \textbf{Upper}\\
\midrule
0.229 & 0.056 & 0.333 & $-0.058$ & $-0.167$ & 0.020\\
\bottomrule
\end{tabular*}
\end{table}

\subsection{Focal canonicalized repeat (F-C)}
\label{app:answer-focal-repeat}
The focal repeat uses the 146-query fact-recall set and its 63-query consensus subset; the preceding held-out-policy repeat uses 40 queries with changed selections.
On this set, the focal repeat keeps source order fixed in both route prompts and adds two fresh Luna max draws per route, giving 584 answer rows over 37 users. Source identities and evidence text are preserved, and every paired route has a different selected source set before canonicalization. Raw disagreement across routes is $0.659$ [$0.577,0.740$], disagreement within a route is $0.123$ [$0.083,0.168$], residual route distance is $0.536$ [$0.442,0.628$], and correctness change over repeated draws is $\Delta_{\mathrm{rep}}=0.312$ [$0.193,0.428$]. Selected source composition and context length both vary in this experiment. On the consensus subset of 63 queries, $D_{\mathrm{res}}=0.459$ [$0.322,0.598$] and $\Delta_{\mathrm{rep}}=0.403$ [$0.213,0.580$]. The 52 route-row display changes and 146 query-level source-set changes are distinct predicates.

\subsection{Declared tokenizer length controls}
\label{app:answer-tokenizer-controls}
Exact length matching aligns the complete paired prompts under the declared \texttt{o200k\_base} tokenizer, adding neutral padding only inside the memory field while holding the question, answer suffix, and evidence fixed. The control contains $589\times2=1{,}178$ fresh Luna max answers and leaves source selection and serialization specific to each route intact. On the consensus stratum of 63 queries, the matched control gives $H=0.538$ [$0.387,0.683$] and $\Delta=0.336$ [$0.138,0.528$], so the route contrast remains after token matching. A common neutral extension gives $H=0.621$ [$0.483,0.753$] and $\Delta=0.390$ [$0.184,0.579$] for padding form A, and $H=0.627$ [$0.478,0.767$] and $\Delta=0.356$ [$0.132,0.562$] for padding form B.

\subsection{Recent-3: equal tokens and aligned offsets (R-E)}
\label{app:answer-recent-aligned}
For Recent-3, a route-neutral common prefix and suffix align the question start as well as total prompt length. The control covers 252 answers from 63 queries and 29 users; every route pair has different selected source sets, while evidence and question offsets agree within each pair. The two prompt realizations give $(H,\Delta)=(0.564,0.071)$ with 95\% limits $[0.422,0.697]$ and $[-0.133,0.279]$, respectively, and $(H,\Delta)=(0.534,-0.017)$ with limits $[0.384,0.678]$ and $[-0.229,0.195]$, respectively. The response contrast therefore remains after both length and the principal prompt offset are aligned; correctness change remains sensitive to answer realization.

\subsection{Core cross-model replication}
\label{app:answer-model-replication}
The three core fixed-input matrices reuse the same PersonaMem states, source identities, question and option strings, token controls, and support labels for all five answer models. The focal exact route contains 63 paired queries from 29 users; focal support restoration contains 33 queries from 21 users; and Recent-3 support restoration contains 36 queries from 18 users. Table~\ref{tab:answer-model-replication} gives point estimates, and Table~\ref{tab:answer-model-intervals} gives the corresponding user-bootstrap intervals.

\begin{table}[ht]
\centering
\small
\caption{Core answer-model replication on fixed PersonaMem inputs. $R$--$A$ is rescue minus absent and $R$--$P$ is rescue minus matched non-support replacement. Entries are equal-user point estimates.}
\label{tab:answer-model-replication}
\setlength{\tabcolsep}{2pt}
\begin{tabular*}{\linewidth}{@{\extracolsep{\fill}}lrrrrrr@{}}
\toprule
\textbf{Model} & \textbf{$H_{\mathrm{focal}}$} & \textbf{$\Delta_{\mathrm{focal}}$} & \textbf{Focal $R$--$A$} & \textbf{Focal $R$--$P$} & \textbf{Recent $R$--$A$} & \textbf{Recent $R$--$P$}\\
\midrule
Astra & 0.581 & 0.442 & 0.808 & 0.756 & 0.671 & 0.819\\
Sol & 0.559 & 0.399 & 0.764 & 0.683 & 0.685 & 0.759\\
Terra & 0.558 & 0.376 & 0.774 & 0.681 & 0.681 & 0.810\\
Luna & 0.614 & 0.437 & 0.671 & 0.601 & 0.648 & 0.750\\
Qwen3-8B & 0.338 & 0.175 & 0.214 & 0.143 & 0.194 & 0.167\\
\bottomrule
\end{tabular*}
\end{table}

\begin{table}[ht]
\centering
\scriptsize
\caption{95\% user-bootstrap intervals for the core cross-model replication. Point estimates are in Table~\ref{tab:answer-model-replication}; each entry here reports Lower and Upper only. $R$--$A$ and $R$--$P$ use the definitions in that table.}
\label{tab:answer-model-intervals}
\setlength{\tabcolsep}{1.2pt}
\begin{tabular*}{\linewidth}{@{\extracolsep{\fill}}lrrrrrrrrrr@{}}
\toprule
 & \multicolumn{2}{c}{$\Delta_{\mathrm{focal}}$} & \multicolumn{2}{c}{Focal $R$--$A$} & \multicolumn{2}{c}{Focal $R$--$P$} & \multicolumn{2}{c}{Recent $R$--$A$} & \multicolumn{2}{c}{Recent $R$--$P$}\\
\cmidrule(lr){2-3}\cmidrule(lr){4-5}\cmidrule(lr){6-7}\cmidrule(lr){8-9}\cmidrule(lr){10-11}
\textbf{Model} & \textbf{Lower} & \textbf{Upper} & \textbf{Lower} & \textbf{Upper} & \textbf{Lower} & \textbf{Upper} & \textbf{Lower} & \textbf{Upper} & \textbf{Lower} & \textbf{Upper}\\
\midrule
Astra & 0.260 & 0.614 & 0.649 & 0.948 & 0.593 & 0.897 & 0.509 & 0.819 & 0.704 & 0.926\\
Sol & 0.218 & 0.570 & 0.607 & 0.903 & 0.524 & 0.829 & 0.551 & 0.810 & 0.653 & 0.861\\
Terra & 0.182 & 0.557 & 0.619 & 0.917 & 0.524 & 0.823 & 0.537 & 0.815 & 0.690 & 0.917\\
Luna & 0.265 & 0.598 & 0.500 & 0.821 & 0.440 & 0.754 & 0.519 & 0.773 & 0.639 & 0.856\\
Qwen3-8B & 0.043 & 0.319 & 0.071 & 0.381 & -0.071 & 0.341 & 0.083 & 0.333 & 0.042 & 0.319\\
\bottomrule
\end{tabular*}
\end{table}

\FloatBarrier
\paragraph{Extended answer collection.} The collected matrix contains 22,242 completed direct-answer rows for each of Luna and Qwen3-8B across 18 answer-bearing extensions, including mechanism schedules, policy interfaces, state transfer, PersonaMem restoration, and LongMemEval inputs. The manuscript reports Luna estimates for these extensions unless a model comparison is stated. For LongMemEval, both answer models are scored by the same masked Terra rubric; the model-level coverage and restoration comparisons appear in Appendix~\ref{app:longmem-scale}.

Within the provenance-policy family, each schedule keeps the selected records and serialized context fixed across its paired conditions. The gated arm uses the benchmark session index to resolve conflicts. The neutral arm asks the answer model to compare evidence without a recency rule. The placebo arm displays a non-benchmark index unrelated to the benchmark index. The accuracy comparison used a reference margin of two percentage points; its interval is reported directly. Principal, boundary, secondary, and retained context comparisons are reported in Tables~\ref{tab:full-primary}--\ref{tab:full-pathways}.

All principal policy arms use the same answer prompt.
\begin{quote}
You are a helpful assistant. Based only on the user's memory/history, choose the best answer to the question. Output only one lowercase letter. The valid outputs are a, b, c, or d.
\end{quote}

For the provenance gated arm, the appended instruction reads as follows. \textquotedblleft When two memory statements conflict, use the statement with the higher benchmark session index as the current state, and treat lower index statements as historical context. Do not let arrival order override the benchmark index.\textquotedblright{} The neutral and placebo instructions follow the same fixed answer schema.

\subsection{Cross-fitted support restoration}
\label{app:answer-crossfit-restoration}
Among the nine cross-fitted pairs whose direct-support source sets change, the supported reference and restoration each produce 31 correct answers out of 36 draws per condition, compared with 29 for the route alternative and 22 for matched replacement without direct support. Within the two pairs that lose all direct support, the reference and restoration each yield eight correct answers out of eight draws, while the alternative and matched replacement each yield four. These nine eligible pairs come from three evaluation users.

\subsection{Support Restoration under Bounded Recent Retention}
\label{app:recent-support}

The bounded recent policy keeps the last three arriving records, independently of the question and support labels. Reversing construction therefore selects the first three records in benchmark source order. Of the 63 primary pairs, 40 expose exactly one annotated direct-support record on one route and none on the other. Four lack an independent non-support replacement in the original pool, leaving 36 pairs from 18 users. The remaining 23 primary pairs do not meet the single-support-loss condition. Eligibility uses the existing source labels without answer outcomes.

The supported and absent conditions retain their original selected sources. Restoration replaces the smallest source-order record in the absent state with the sole direct-support record from the supported state. The non-support condition replaces the same record with the smallest source-order record outside the absent set that is not labeled direct support. Each condition contains three distinct records, displayed in canonical source order. The inserted support and non-support records occupy the same display position in all 36 pairs. Full source bodies and the question/options interface reproduce the archived recent-policy prompts. Neutral padding matches total token count and evidence and question start positions within each pair using \texttt{o200k\_base}.

The eligibility set, replacement rule, three fresh Luna max draws per condition, and analysis were fixed before collecting the 432 answers. All calls completed. Accuracy is averaged over draws within questions, over questions within users, and then equally across users. Table~\ref{tab:recent-support-restoration} reports all four conditions and the three specified contrasts, using 10,000 user-bootstrap resamples with seed 2026090911. These estimates apply to the eligible support-loss stratum of this policy.

\begin{table}[htbp]
\centering
\small
\caption{Bounded-recent-3 support restoration on 36 pairs from 18 users. Each condition has three fresh answers per pair. Lower and Upper are 95\% user-bootstrap limits; estimates give users equal weight. Replacement inserts an independent record without direct support.}
\label{tab:recent-support-restoration}
\setlength{\tabcolsep}{2pt}
\textbf{A. Accuracy}\par\vspace{2pt}
\begin{tabular*}{\linewidth}{@{\extracolsep{\fill}}rrrrrrrrrrrr@{}}
\toprule
\multicolumn{3}{c}{Support present} & \multicolumn{3}{c}{Support absent} & \multicolumn{3}{c}{Restoration} & \multicolumn{3}{c}{Replacement}\\
\cmidrule(lr){1-3}\cmidrule(lr){4-6}\cmidrule(lr){7-9}\cmidrule(lr){10-12}
\textbf{Estimate} & \textbf{Lower} & \textbf{Upper} & \textbf{Estimate} & \textbf{Lower} & \textbf{Upper} & \textbf{Estimate} & \textbf{Lower} & \textbf{Upper} & \textbf{Estimate} & \textbf{Lower} & \textbf{Upper}\\
\midrule
0.940 & 0.866 & 1.000 & 0.269 & 0.144 & 0.407 & 0.917 & 0.833 & 0.981 & 0.167 & 0.074 & 0.273\\
\bottomrule
\end{tabular*}
\par\smallskip
\textbf{B. Paired accuracy contrasts}\par\vspace{2pt}
\begin{tabular*}{\linewidth}{@{\extracolsep{\fill}}rrrrrrrrr@{}}
\toprule
\multicolumn{3}{c}{Restoration $-$ absent} & \multicolumn{3}{c}{Restoration $-$ replacement} & \multicolumn{3}{c}{Restoration $-$ present}\\
\cmidrule(lr){1-3}\cmidrule(lr){4-6}\cmidrule(lr){7-9}
\textbf{Estimate} & \textbf{Lower} & \textbf{Upper} & \textbf{Estimate} & \textbf{Lower} & \textbf{Upper} & \textbf{Estimate} & \textbf{Lower} & \textbf{Upper}\\
\midrule
0.648 & 0.519 & 0.773 & 0.750 & 0.639 & 0.856 & $-0.023$ & $-0.051$ & 0.000\\
\bottomrule
\end{tabular*}
\end{table}
\FloatBarrier

\subsection{Expanded PersonaMem Restoration Coverage}
\label{app:personamem-restoration-expanded}

The full fact-recall plane contains 146 queries from 37 users. We apply the same outcome-blind structural gate used for the primary intervention to every query: the two routes must differ by one direct-support record, the absent route must have a removable non-direct record, all four conditions must preserve source count, and the original pool must contain an independent non-direct replacement. This yields 78 focal-compactor queries from 33 users and 71 Recent-3 queries from 28 users. The primary rows below reproduce the frozen matrices; All eligible adds the queries outside the semantic-consensus screen, and Outside primary isolates that added stratum. The expanded rows use fresh Luna max answers for the added queries, with two padding variants and two draws per condition for Focal and three draws per condition for Recent-3. Accuracy averages draws within queries, queries within users, and users equally. Intervals use 10,000 user-bootstrap resamples.

\begin{table}[htbp]
\centering
\scriptsize
\caption{Expanded PersonaMem restoration accuracies. Primary reproduces the frozen semantic-consensus analysis; All eligible includes every structurally eligible fact-recall query; Outside primary contains the added queries only. Estimates and Lower/Upper are 95\% user-bootstrap limits.}
\label{tab:personamem-restoration-expanded}
\setlength{\tabcolsep}{1.5pt}
\textbf{A. Focal compactor}\par\vspace{2pt}
\begin{tabular*}{\linewidth}{@{\extracolsep{\fill}}lrr*{12}{r}@{}}
\toprule
 &  &  & \multicolumn{3}{c}{Present} & \multicolumn{3}{c}{Absent} & \multicolumn{3}{c}{Restored} & \multicolumn{3}{c}{Replacement}\\
\cmidrule(lr){4-6}\cmidrule(lr){7-9}\cmidrule(lr){10-12}\cmidrule(lr){13-15}
\textbf{Scope} & \textbf{Queries} & \textbf{Users} & \textbf{Estimate} & \textbf{Lower} & \textbf{Upper} & \textbf{Estimate} & \textbf{Lower} & \textbf{Upper} & \textbf{Estimate} & \textbf{Lower} & \textbf{Upper} & \textbf{Estimate} & \textbf{Lower} & \textbf{Upper}\\
\midrule
Primary & 33 & 21 & 0.875 & 0.778 & 0.952 & 0.210 & 0.087 & 0.349 & 0.881 & 0.790 & 0.952 & 0.280 & 0.141 & 0.440\\
All eligible & 78 & 33 & 0.894 & 0.831 & 0.948 & 0.191 & 0.113 & 0.279 & 0.879 & 0.800 & 0.942 & 0.239 & 0.134 & 0.354\\
Outside primary & 45 & 27 & 0.906 & 0.832 & 0.968 & 0.150 & 0.074 & 0.234 & 0.881 & 0.785 & 0.955 & 0.149 & 0.058 & 0.260\\
\bottomrule
\end{tabular*}
\par\smallskip
\textbf{B. Bounded Recent-3}\par\vspace{2pt}
\begin{tabular*}{\linewidth}{@{\extracolsep{\fill}}lrr*{12}{r}@{}}
\toprule
 &  &  & \multicolumn{3}{c}{Present} & \multicolumn{3}{c}{Absent} & \multicolumn{3}{c}{Restored} & \multicolumn{3}{c}{Replacement}\\
\cmidrule(lr){4-6}\cmidrule(lr){7-9}\cmidrule(lr){10-12}\cmidrule(lr){13-15}
\textbf{Scope} & \textbf{Queries} & \textbf{Users} & \textbf{Estimate} & \textbf{Lower} & \textbf{Upper} & \textbf{Estimate} & \textbf{Lower} & \textbf{Upper} & \textbf{Estimate} & \textbf{Lower} & \textbf{Upper} & \textbf{Estimate} & \textbf{Lower} & \textbf{Upper}\\
\midrule
Primary & 36 & 18 & 0.940 & 0.866 & 1.000 & 0.269 & 0.144 & 0.407 & 0.917 & 0.833 & 0.981 & 0.167 & 0.074 & 0.273\\
All eligible & 71 & 28 & 0.868 & 0.802 & 0.929 & 0.225 & 0.143 & 0.318 & 0.927 & 0.865 & 0.979 & 0.109 & 0.052 & 0.176\\
Outside primary & 35 & 22 & 0.778 & 0.626 & 0.909 & 0.216 & 0.091 & 0.356 & 0.970 & 0.924 & 1.000 & 0.073 & 0.005 & 0.182\\
\bottomrule
\end{tabular*}
\par\smallskip
\textbf{C. Paired correctness contrasts}\par\vspace{2pt}
\begin{tabular*}{\linewidth}{@{\extracolsep{\fill}}lrrrrrrrrrr@{}}
\toprule
 & \multicolumn{3}{c}{Restored $-$ Absent} & \multicolumn{3}{c}{Restored $-$ Replacement} & \multicolumn{3}{c}{Restored $-$ Present}\\
\cmidrule(lr){2-4}\cmidrule(lr){5-7}\cmidrule(lr){8-10}
\textbf{Policy and scope} & \textbf{Estimate} & \textbf{Lower} & \textbf{Upper} & \textbf{Estimate} & \textbf{Lower} & \textbf{Upper} & \textbf{Estimate} & \textbf{Lower} & \textbf{Upper}\\
\midrule
Focal (Primary) & 0.671 & 0.496 & 0.817 & 0.601 & 0.442 & 0.756 & 0.006 & -0.044 & 0.067\\
Focal (All eligible) & 0.688 & 0.576 & 0.790 & 0.640 & 0.487 & 0.771 & -0.015 & -0.047 & 0.013\\
Focal (Outside primary) & 0.731 & 0.611 & 0.843 & 0.732 & 0.559 & 0.874 & -0.025 & -0.063 & 0.005\\
Recent-3 (Primary) & 0.648 & 0.519 & 0.773 & 0.750 & 0.639 & 0.856 & -0.023 & -0.051 & 0.000\\
Recent-3 (All eligible) & 0.702 & 0.607 & 0.789 & 0.818 & 0.743 & 0.889 & 0.059 & -0.013 & 0.133\\
Recent-3 (Outside primary) & 0.754 & 0.614 & 0.879 & 0.896 & 0.785 & 0.975 & 0.192 & 0.048 & 0.359\\
\bottomrule
\end{tabular*}
\end{table}
\FloatBarrier

\section{Source Stable Lexical Control}
\label{app:external-bm25}

An external lexical control follows the Agent Memory Benchmark's memory layout for the 32K PersonaMem-v1 files. It is a format-level reference based on BM25 ranking. Sessions are split at system turns, formatted as memory documents, segmented into 512-token windows using \texttt{cl100k\_base}, and ranked with BM25 at $k=10$ using lowercase tokens matched by \texttt{[A-Za-z][A-Za-z0-9']+}. The chronological and replay arms ingest exactly the same document multiset in opposite orders. Tie resolution follows insertion order, so the score function is held fixed while the construction route varies.

The internal-state measurement rebuilds 74 indexes for the 37 user histories and reproduces all 1,178 archived retrieval contexts. Its signature includes the ordered chunk identifiers and bodies, document term counts and lengths, inverse document frequencies, and BM25 parameters. Index positions are retained because they resolve score ties. The signatures differ for all 589 query pairs, while sorting indexed records by stable identifiers yields identical bodies and term counts in every pair.

The same top-$k$ source set is retrieved under both ingestion orders for all 589 queries. The compiled contexts also agree exactly, so the two route prompts are identical; differences between independently generated answers arise at the answer stage. The \texttt{gpt-5.6-luna} max-effort readout is reported as a baseline accuracy estimate and is not pooled with the compactor estimates.

For the chronological route, accuracy is $0.682$ with a 95\% bootstrap interval of $[0.632,0.727]$ over 37 users.

\section{MemoChat-Style Configuration}
\label{app:external-memochat}

\subsection{Policy and pairing}
\label{app:memochat-policy-pairing}
We evaluate the MemoChat-style sequential-memory runtime supplied by MemoryData on the same PersonaMem source pools. It follows the memo-based design of \citet{lu2023memochat}, with the simplified configuration specified below. Its sequential buffer, topic summaries, retention of recent dialogue, and topic retrieval are evaluated on the semantic-consensus stratum of 63 queries from 29 users and the complete fact-recall set of 146 queries. Within each pair, the routes differ only in the order in which identical source records enter the policy; source text, timestamps, task, answer model, and answer interface remain fixed. The returned state is mapped to its contributing dialogue records before answer generation.

The fixed runtime triggers topic summarization after six chunks, retains two recent chunks, keeps at most three topics in a window, and retrieves four topic summaries and two recent dialogs, with each topic summary retaining two dialogs. Summaries are limited to 320 characters, keyword extraction to at most 16 terms, and the embedding uses a deterministic 1024-dimensional hashing representation; learned topic segmentation is disabled. The 62 pairs with unchanged selected sources and changed compiled contexts locate a route effect after source selection; the experiment does not separate changes in summary content from changes in final context order.

\subsection{Internal state}
\label{app:memochat-internal-state}
A separate reconstruction captures the topic records, summaries, stored dialogs, and recent-dialog buffer after construction and before retrieval. Signatures exclude generated identifiers and timestamps, canonicalize record maps, and retain dialog order within records and the recent buffer. All 63 pairs have different signatures. Reconstruction uses the frozen inputs and pinned runtime without model calls. Selected source identities match all 126 archived exposures; 125 contexts match exactly, and one differs only in the order of tied retrieval blocks. The original answer contexts are retained for the response analysis.

\subsection{Answer boundary}
\label{app:memochat-answer-boundary}
At the answer boundary, each returned context receives neutral padding after the memory content until the two full prompts have equal length under \texttt{o200k\_base}; within each pair, the question and answer suffix are identical. The original route matrix gives one answer per route and query, and a paired repeat matrix reuses each prompt for six additional answers. Together they provide seven answer draws per route for each of 63 query pairs from 29 users. Each route prompt is reused unchanged across its seven draws, and the two routes have equal full-prompt token counts. The route-level estimates are summarized in Table~\ref{tab:external-memochat}.

\begin{table}[ht]
\centering
\caption{Direct MemoChat route trace on the semantic consensus stratum. This table is separate from the all-pool coverage matrix; estimates weight users equally, the Lower and Upper columns give 95\% bootstrap limits, and \textit{Queries} gives the denominator for all change counts.}
\label{tab:external-memochat}
\setlength{\tabcolsep}{2pt}
\textbf{A. Changed pairs}\par\vspace{2pt}
\begin{tabular*}{\linewidth}{@{\extracolsep{\fill}}lrrr@{}}
\toprule
 & \multicolumn{3}{c}{Changed pairs}\\
\cmidrule(lr){2-4}
\textbf{Policy} & \textbf{Source changes} & \textbf{Context changes} & \textbf{Queries}\\
\midrule
MemoChat & 1 & 63 & 63\\
\bottomrule
\end{tabular*}
\par\smallskip
\textbf{B. Route coordinates}\par\vspace{2pt}
\begin{tabular*}{\linewidth}{@{\extracolsep{\fill}}rrrrrrrrr@{}}
\toprule
\multicolumn{3}{c}{$J$} & \multicolumn{3}{c}{$H$} & \multicolumn{3}{c}{$\Delta$}\\
\cmidrule(lr){1-3}\cmidrule(lr){4-6}\cmidrule(lr){7-9}
\textbf{Estimate} & \textbf{Lower} & \textbf{Upper} & \textbf{Estimate} & \textbf{Lower} & \textbf{Upper} & \textbf{Estimate} & \textbf{Lower} & \textbf{Upper}\\
\midrule
0.995 & 0.985 & 1.000 & 0.084 & 0.007 & 0.188 & 0.076 & 0.000 & 0.179\\
\bottomrule
\end{tabular*}
\par\smallskip
\textbf{C. Accuracy}\par\vspace{2pt}
\begin{tabular*}{\linewidth}{@{\extracolsep{\fill}}rrrrrr@{}}
\toprule
\multicolumn{3}{c}{Chronological accuracy} & \multicolumn{3}{c}{Replay accuracy}\\
\cmidrule(lr){1-3}\cmidrule(lr){4-6}
\textbf{Estimate} & \textbf{Lower} & \textbf{Upper} & \textbf{Estimate} & \textbf{Lower} & \textbf{Upper}\\
\midrule
0.828 & 0.697 & 0.940 & 0.904 & 0.805 & 0.982\\
\bottomrule
\end{tabular*}
\end{table}

\subsection{Two answer samples}
\label{app:memochat-two-samples}
The 63-pair route trace and the 146-query coverage matrix use the same policy, source pools, paired prompts, and \texttt{gpt-5.6-luna} max-effort answer boundary. They are independent answer realizations: the former supplies the direct route profile in Table~\ref{tab:external-memochat}, whereas the latter covers all pools and applies the semantic strata after generation. On their shared 63-pair consensus stratum, the direct trace gives $H=0.084,\Delta=0.076$, whereas the coverage realization gives $H=0.182,\Delta=-0.010$. The structural observations also agree: $J=0.995$, one pair changes its selected source set, and all 63 pairs change their compiled contexts. The seven-draw route matrix gives $H_{\times}=0.106$, $H_{\mathrm{within}}=0.069$, and $D_{\mathrm{res}}=0.037$, so the coverage estimate is retained as a separate answer-sensitivity estimate.

Repeated answer decomposition for the same policy pair is reported in Table~\ref{tab:external-memochat-repeat}.

\begin{table}[ht]
\centering
\caption{Repeated answer decomposition for the MemoChat route pair. The Lower and Upper columns give 95\% bootstrap limits with equal user weight.}
\label{tab:external-memochat-repeat}
\setlength{\tabcolsep}{2pt}
\textbf{A. Response disagreement}\par\vspace{2pt}
\begin{tabular*}{\linewidth}{@{\extracolsep{\fill}}rrrrrr@{}}
\toprule
\multicolumn{3}{c}{$H_{\times}$} & \multicolumn{3}{c}{$H_{\mathrm{within}}$}\\
\cmidrule(lr){1-3}\cmidrule(lr){4-6}
\textbf{Estimate} & \textbf{Lower} & \textbf{Upper} & \textbf{Estimate} & \textbf{Lower} & \textbf{Upper}\\
\midrule
0.106 & 0.042 & 0.186 & 0.069 & 0.030 & 0.114\\
\bottomrule
\end{tabular*}
\par\smallskip
\textbf{B. Residual and distribution distance}\par\vspace{2pt}
\begin{tabular*}{\linewidth}{@{\extracolsep{\fill}}rrrrrr@{}}
\toprule
\multicolumn{3}{c}{$D_{\mathrm{res}}$} & \multicolumn{3}{c}{$\widehat D^\star$}\\
\cmidrule(lr){1-3}\cmidrule(lr){4-6}
\textbf{Estimate} & \textbf{Lower} & \textbf{Upper} & \textbf{Estimate} & \textbf{Lower} & \textbf{Upper}\\
\midrule
0.037 & 0.005 & 0.093 & 0.047 & 0.011 & 0.106\\
\bottomrule
\end{tabular*}
\par\smallskip
\textbf{C. Same-draw disagreement and repeated correctness change}\par\vspace{2pt}
\begin{tabular*}{\linewidth}{@{\extracolsep{\fill}}rrrrrr@{}}
\toprule
\multicolumn{3}{c}{$H$} & \multicolumn{3}{c}{$\Delta_{\mathrm{rep}}$}\\
\cmidrule(lr){1-3}\cmidrule(lr){4-6}
\textbf{Estimate} & \textbf{Lower} & \textbf{Upper} & \textbf{Estimate} & \textbf{Lower} & \textbf{Upper}\\
\midrule
0.109 & 0.042 & 0.191 & 0.044 & $-0.012$ & 0.119\\
\bottomrule
\end{tabular*}
\par\smallskip
\textbf{D. Accuracy}\par\vspace{2pt}
\begin{tabular*}{\linewidth}{@{\extracolsep{\fill}}rrrrrr@{}}
\toprule
\multicolumn{3}{c}{Chronological} & \multicolumn{3}{c}{Replay}\\
\cmidrule(lr){1-3}\cmidrule(lr){4-6}
\textbf{Estimate} & \textbf{Lower} & \textbf{Upper} & \textbf{Estimate} & \textbf{Lower} & \textbf{Upper}\\
\midrule
0.855 & 0.755 & 0.941 & 0.899 & 0.822 & 0.962\\
\bottomrule
\end{tabular*}
\end{table}

\subsection{Repeated answers}
\label{app:memochat-repeated-answers}
Seven draws per route give $H_{\times}=0.106$ and $H_{\mathrm{within}}=0.069$; residual $D_{\mathrm{res}}=0.037$ has a 95\% bootstrap interval of $[0.005,0.093]$, and the plug-in answer-distribution distance is $\widehat D^\star=0.047$ with interval $[0.011,0.106]$. 

\subsection{Fixed-source sensitivity}
\label{app:memochat-fixed-source}
A post-hoc structural stratum retains the 62 query pairs whose selected source sets agree across routes, covering all 29 users. Selection reads source identifiers only; all 62 compiled contexts differ. Seven draws per route give $H_{\times}=0.109$ (95\% CI $[0.044,0.190]$), $H_{\mathrm{within}}=0.071$ ($[0.031,0.116]$), and $D_{\mathrm{res}}=0.037$ ($[0.005,0.093]$). The correctness contrast is $\Delta_{\mathrm{rep}}=0.046$ ($[-0.010,0.119]$). Thus, the residual response difference persists with fixed source membership, while its direction of task benefit remains uncertain.

\subsection{Correctness projection}
\label{app:memochat-correctness}
To test whether this variation concerns success probabilities, an exploratory analysis applies equation~\ref{eq:residual} to the binary correctness labels $c_q^{(o)}$ instead of answer options. Under the same sampling assumptions, the projected residual $D_{q,\mathrm{correct}}$ has expectation $\mathbb{E}[D_{q,\mathrm{correct}}]=(p_{q,y_q}^{(f)}-p_{q,y_q}^{(r)})^2$. The equal-user estimate is $0.038$ (95\% CI $[0.005,0.093]$) on both the complete 63-query set and the 62-query fixed-source stratum. This unsigned task-level difference can coexist with a signed mean contrast whose interval spans zero. Both strata were analyzed without selecting questions by their answer outcomes.

\section{Independent A-MEM Policy}
\label{app:external-amem}

\subsection{Policy and pairing}
\label{app:amem-policy-pairing}
A-MEM (Agentic Memory)~\citep{xu2025amem} provides a persistent policy with note evolution, linked memory units, and top-$k$ dense retrieval. The 63 query pairs span 29 unique source pools and 29 users. Each pool is ingested in chronological and replay order without the query, producing 58 constructed states; retrieval is evaluated separately for each question. Within each pair, source identities, source bodies, the task, the gold option, and the answer interface are held fixed.

\subsection{State signature}
\label{app:amem-state-signature}
Each signature applies SHA-256 to the JSON serialization of note identifiers, source-traceable links, tags, and note contexts, with object keys sorted and note insertion order retained. Sorting notes by stable source identifier and sorting each link and tag list still yields different states in all 29 pool pairs: tags and note contexts differ, while links agree. Across 154 source-matched notes in the 29 history pairs, 152 tag sets and 144 note descriptions differ; links and note bodies agree. These counts use each history once, independently of how many questions query it. Dense retrieval indexes note bodies, and the answer context concatenates the returned content fields; evolved tags and note contexts are not serialized into that prompt. Note evolution uses model calls, so this comparison concerns paired state realizations; it does not isolate the causal contribution of order to internal-state variation. 

\subsection{Boundary result}
\label{app:amem-boundary-result}
The recorded A-MEM internal-state signatures differ for all 63 pairs, whereas the selected source sets and compiled contexts agree for every pair, yielding source-set overlap $J=1.000$. Across the paired answer calls, the equal-user estimates are $H=0.103$ with a 95\% bootstrap interval of $[0.022,0.209]$ and $\Delta=-0.042$ with interval $[-0.155,0.043]$; chronological and replay accuracy are $0.896$ and $0.854$, respectively.

\begin{table}[ht]
\centering
\small
\caption{Independent A-MEM boundary check. The Lower and Upper columns are 95\% bootstrap limits with equal user weight.}
\label{tab:external-amem}
\setlength{\tabcolsep}{2pt}
\textbf{A. Changed pairs}\par\vspace{2pt}
\begin{tabular*}{\linewidth}{@{\extracolsep{\fill}}lrrrr@{}}
\toprule
 & \multicolumn{4}{c}{Changed pairs}\\
\cmidrule(lr){2-5}
\textbf{Policy} & \textbf{Signature changes} & \textbf{Source changes} & \textbf{Context changes} & \textbf{Queries}\\
\midrule
A-MEM & 63 & 0 & 0 & 63\\
\bottomrule
\end{tabular*}
\par\smallskip
\textbf{B. Route coordinates}\par\vspace{2pt}
\begin{tabular*}{\linewidth}{@{\extracolsep{\fill}}rrrrrrrrr@{}}
\toprule
\multicolumn{3}{c}{$J$} & \multicolumn{3}{c}{$H$} & \multicolumn{3}{c}{$\Delta$}\\
\cmidrule(lr){1-3}\cmidrule(lr){4-6}\cmidrule(lr){7-9}
\textbf{Estimate} & \textbf{Lower} & \textbf{Upper} & \textbf{Estimate} & \textbf{Lower} & \textbf{Upper} & \textbf{Estimate} & \textbf{Lower} & \textbf{Upper}\\
\midrule
1.000 & 1.000 & 1.000 & 0.103 & 0.022 & 0.209 & $-0.042$ & $-0.155$ & 0.043\\
\bottomrule
\end{tabular*}
\par\smallskip
\textbf{C. Accuracy}\par\vspace{2pt}
\begin{tabular*}{\linewidth}{@{\extracolsep{\fill}}rrrrrr@{}}
\toprule
\multicolumn{3}{c}{Chronological accuracy} & \multicolumn{3}{c}{Replay accuracy}\\
\cmidrule(lr){1-3}\cmidrule(lr){4-6}
\textbf{Estimate} & \textbf{Lower} & \textbf{Upper} & \textbf{Estimate} & \textbf{Lower} & \textbf{Upper}\\
\midrule
0.896 & 0.803 & 0.971 & 0.854 & 0.729 & 0.957\\
\bottomrule
\end{tabular*}
\end{table}

\subsection{Retrieval boundary}
\label{app:amem-retrieval-boundary}
The public A-MEM note-construction, evolution, link-update, and top-$k$ control flow are retained from the pinned implementation, together with its ChromaDB-backed dense retrieval using SentenceTransformer embeddings. The 58 archived states are restored into isolated collections without new note evolution and queried separately for all 63 questions. Representative queries reproduce the original retrieved source order and body content in all 58 states. Restored metadata reflect the archived final notes and can differ from the original Chroma metadata; these fields do not enter the answer context. Of the 126 complete answer prompts, 62 match the original prompts exactly and reuse those answers; the remaining 64 receive fresh answers. The resulting matrix contains 63 complete query pairs and no failed answers.

\section{Measurement Conventions and Estimands}
\label{app:measurements}

The observable coordinates are defined for query $q$ under policy $\mathcal P$ and schedule $\tau$: source overlap $J_q$ (Jaccard similarity of exposed source identities), answer disagreement $H_q$ (indicator for differing forward/replay answers), and correctness change $\Delta_q$ (replay minus forward correctness). The support-restricted overlap $J_q^T$ restricts the denominator to the union of exposed support sets $T_q\cap V_q^{(f)}$ and $T_q\cap V_q^{(r)}$; pairs where neither route exposes direct support are recorded as not applicable.

The equal-user estimate is $\overline z(\mathcal P;\tau)=\widehat{\mathbb E}_{u}[\widehat{\mathbb E}_{q\mid u}[z_q]]$; the policy contrast is $\Gamma_z(\tau)=z_{\mathrm{ord}}(\tau)-z_{\mathrm{can}}(\tau)$; the provenance contrast is gated minus preserved. Bootstrap intervals (Appendix~\ref{app:uncertainty}) resample users.

For $m$ repeated draws under fixed prompts, $H_{\times}$ averages disagreement over $m^2$ cross-route pairs, $H_{\mathrm{within}}$ averages within-route disagreement, and $\Delta_{\mathrm{rep}}$ is mean replay minus forward correctness. The plug-in distance is $\widehat D_q^\star=\tfrac12\sum_{a\in\mathcal A_q}(\widehat p_{qa}^{(f)}-\widehat p_{qa}^{(r)})^2$.

\subsection{Experiment index}
\label{app:measurements-index}
The index lists the Luna reporting matrices. Complete Qwen3-8B counterparts are retained for the core matrices and the extended answer collection; the scored LongMemEval counterpart is described in Appendix~\ref{app:longmem-scale}.
Table~\ref{tab:experiment-index} identifies the answer matrices underlying the principal comparisons. The artifact's \texttt{experiment-manifest.json} maps each experiment to its retained inputs, answers, prompt controls, model configuration, analysis, and audit files. The fresh-answer and prompt-identity analyses are reproduced by \texttt{scripts/analyze\_trace\_stats.py}; they reuse these matrices without collecting additional answers. PersonaMem estimates average draws within a question, then questions within a user, and finally users equally; paired contrasts use the same users in both conditions. LongMemEval weights questions equally.

\begin{table}[htbp]
\centering\small
\caption{Index of principal answer experiments. $m$ counts answers per route or condition and padding variant. Exact focal routes combine one formal and two fresh draws; MemoChat combines one original and six fresh draws. F-S+ and R-S+ include all structurally eligible PersonaMem restoration pairs. A-MEM uses the corrected query-specific matrix. Mechanism and streaming rows describe the full retained coverage; their primary analyses use the specified semantic subsets. NR means not reported; LongMemEval supplies no user identifiers. Luna is the reporting model for extended rows unless a model comparison is shown.}
\label{tab:experiment-index}
\setlength{\tabcolsep}{3pt}
\begin{tabular*}{\linewidth}{@{\extracolsep{\fill}}lllrrr@{}}
\toprule
\textbf{Trace} & \textbf{Experiment} & \textbf{Model} & \textbf{Queries} & \textbf{Users} & \textbf{$m$}\\
\midrule
F-C & Focal canonical repeat & Luna & 63 & 29 & 2\\
F-E & Focal exact matched routes & Luna & 63 & 29 & 3\\
R-E & Recent-3 exact matched routes & Luna & 63 & 29 & 3\\
F-S & Focal restoration & Luna & 33 & 21 & 2\\
R-S & Recent-3 restoration & Luna & 36 & 18 & 3\\
F-S+ & Focal restoration, all eligible & Luna & 78 & 33 & 2\\
R-S+ & Recent-3 restoration, all eligible & Luna & 71 & 28 & 3\\
F-U & Focal state reuse & Luna & 34 & 20 & 2\\
F-M & Mechanism schedules & Luna & 589 & 37 & 1\\
X-P & Cross-fitted streaming & Luna & 589 & 37 & 1\\
X-T & Cross-fitted targeted rescue & Luna & 2 & 2 & 2\\
X-I & Cross-fitted support identity & Luna & 9 & 3 & 2\\
M-C & MemoChat repeated answers & Luna & 63 & 29 & 7\\
M-R/M-D & MemoChat construction/display & Luna & 63 & 29 & 2\\
A-Q & A-MEM query-specific answers & Luna & 63 & 29 & 1\\
L-A & LongMemEval paired answers & Luna & 470 & NR & 1\\
L-R & LongMemEval restoration & Luna & 24 & NR & 2\\
L-A/Qwen3 & LongMemEval paired answers & Qwen3-8B & 470 & NR & 1\\
L-R/Qwen3 & LongMemEval restoration & Qwen3-8B & 24 & NR & 2\\
F-E/Sol & Focal matched routes & Sol & 63 & 29 & 1\\
F-S/Sol & Focal restoration & Sol & 33 & 21 & 2\\
\bottomrule
\end{tabular*}
\end{table}

\subsection{Answer-call metadata}
\label{app:measurements-metadata}
Retained traces record model names, reasoning effort, prompts, and draw identifiers. Historical focal and MemoChat records lack dated model snapshots or wall-clock times; newer collections (M-D, LongMemEval, cross-fitted) retain UTC timestamps and ephemeral thread identifiers but no explicit temperature or answer seed. The artifact's \texttt{runtime-metadata.json} documents the available settings and their scope. Fresh draws are additional calls to fixed prompts; independence and stationarity remain assumptions of Equation~\ref{eq:answer-distance}.

\subsection{Information lost by simpler readouts}
\label{app:measurements-readouts}
Table~\ref{tab:diagnostic-readouts} uses exact prompt equality as the reference and counts misses (changed prompt, unchanged readout) and extra flags (unchanged prompt, changed readout). Source identity misses MemoChat's post-selection context changes; single-answer indicators miss changed prompts when the answer happens to agree. Recording the full prompt resolves these distinctions.

\begin{table}[htbp]
\centering\small
\caption{Simpler readouts compared with complete-prompt identity. All rows use 63 queries and 29 users. Pairs counts query--padding comparisons (two variants for Focal); Changed counts different prompts. Miss and Extra are counts with the definitions above. Answers use the original single-draw matrices.}
\label{tab:diagnostic-readouts}
\setlength{\tabcolsep}{2pt}
\begin{tabular*}{\linewidth}{@{\extracolsep{\fill}}lrrrrrrrr@{}}
\toprule
 & & & \multicolumn{2}{c}{Source set} & \multicolumn{2}{c}{Answer} & \multicolumn{2}{c}{Correctness}\\
\cmidrule(lr){4-5}\cmidrule(lr){6-7}\cmidrule(lr){8-9}
\textbf{Policy} & \textbf{Pairs} & \textbf{Changed} & \textbf{Miss} & \textbf{Extra} & \textbf{Miss} & \textbf{Extra} & \textbf{Miss} & \textbf{Extra}\\
\midrule
Focal & 126 & 126 & 0 & 0 & 59 & 0 & 66 & 0\\
MemoChat & 63 & 63 & 62 & 0 & 59 & 0 & 60 & 0\\
A-MEM & 63 & 0 & 0 & 0 & 0 & 6 & 0 & 5\\
\bottomrule
\end{tabular*}
\end{table}

\section{Admissibility and State Observability}
\label{app:checks}

\subsection{Support after compilation}
\label{app:checks-compiled-support}
Source-level support labels apply directly when the labeled record text is preserved verbatim in the answer context. For a summary adapter, a compiled-support assessment would inspect the actual $C_q^{(o)}$ together with the question, options, and gold answer, withholding route identity and generated answers. It would require a supporting passage or a set of jointly supporting passages, with unresolved cases kept separate. The current MemoChat analysis records context differences without assigning these compiled-support labels. A source-to-state map can establish provenance, but semantic support requires this additional content-level assessment.

\subsection{Complete-pool screening}
\label{app:checks-pool-screening}
The semantic screen uses the complete source pool because screening only the selected records would make eligibility depend on the policy being evaluated. Cards preserve source positions and withhold route-dependent quantities.

Pass A marked 70 pools as admissible, 8 as ambiguous, and 68 as inadmissible; Pass B marked 77, 12, and 57, respectively, with 116 exact agreements and Cohen's $\kappa=0.634$. The largest category-pair disagreement is between admissible and inadmissible labels (12 of 30 disagreements); the retained intersection contains 63 queries from 29 users. A historical target-visible task-preservation reading retained 43 of the 63 consensus queries; intersecting these with the third semantic reading left 23 queries. Table~\ref{tab:task-preservation} preserves these estimates as an archival record. The per-card labels from that reading are unavailable, so its 43/23 membership cannot be reconstructed. The new reading below releases all labels and analyzes its own intersections. This post-hoc check sees the gold option and does not change the outcome-blind primary gate. The consensus row is the primary storage-permutation estimand; the taxonomy row for 146 queries provides broader coverage. The released cards include admissible, recency-dependent, and unresolved examples.

\begin{table}[ht]
\centering
\small
\caption{Historical target-visible task-preservation sensitivity, retained for archival transparency and excluded from the conclusion evidence. Queries and Users give the denominators. Lower and Upper are 95\% user-bootstrap limits. The strict subset also satisfies all three outcome-blind semantic readings. A and B are the two matched padding controls.}
\label{tab:task-preservation}
\setlength{\tabcolsep}{2pt}
\textbf{A. Matched padding A}\par\vspace{2pt}
\begin{tabular*}{\linewidth}{@{\extracolsep{\fill}}lrrrrrrrr@{}}
\toprule
 & & & \multicolumn{3}{c}{$D_{\mathrm{res}}$} & \multicolumn{3}{c}{$\Delta_{\mathrm{rep}}$}\\
\cmidrule(lr){4-6}\cmidrule(lr){7-9}
\textbf{Subset} & \textbf{Queries} & \textbf{Users} & \textbf{Estimate} & \textbf{Lower} & \textbf{Upper} & \textbf{Estimate} & \textbf{Lower} & \textbf{Upper}\\
\midrule
Target preserved & 43 & 25 & 0.483 & 0.294 & 0.671 & 0.442 & 0.238 & 0.633\\
Strict & 23 & 17 & 0.353 & 0.118 & 0.588 & 0.284 & 0.020 & 0.529\\
\bottomrule
\end{tabular*}
\par\smallskip
\textbf{B. Matched padding B}\par\vspace{2pt}
\begin{tabular*}{\linewidth}{@{\extracolsep{\fill}}lrrrrrrrr@{}}
\toprule
 & & & \multicolumn{3}{c}{$D_{\mathrm{res}}$} & \multicolumn{3}{c}{$\Delta_{\mathrm{rep}}$}\\
\cmidrule(lr){4-6}\cmidrule(lr){7-9}
\textbf{Subset} & \textbf{Queries} & \textbf{Users} & \textbf{Estimate} & \textbf{Lower} & \textbf{Upper} & \textbf{Estimate} & \textbf{Lower} & \textbf{Upper}\\
\midrule
Target preserved & 43 & 25 & 0.453 & 0.278 & 0.631 & 0.373 & 0.160 & 0.580\\
Strict & 23 & 17 & 0.382 & 0.176 & 0.608 & 0.314 & 0.059 & 0.569\\
\bottomrule
\end{tabular*}
\end{table}

\FloatBarrier
\subsection{Gold-visible re-reading and intersection analysis}
\label{app:checks-gold-reread}
A new Luna max reading evaluates all 146 complete-pool query cards in separate contexts using the question, options, and gold target. The frozen rubric asks whether the same gold option is supported in both displayed and reversed block order; timestamps and within-block dialogue remain fixed. It explicitly distinguishes fixed temporal evidence from arrival-order-dependent updates. Selected states, generated answers, effect estimates, and previous labels are withheld. The reading labels 108 queries admissible, 7 ambiguous, and 31 inadmissible. Its intersection with the primary set contains 62 queries from 29 users (Gold); further intersection with the third outcome-blind reading contains 24 queries from 18 users (Strict). This clarified rubric addresses the distinction between event time and import order; it constitutes a new model-assisted sensitivity analysis with different instructions from the original outcome-blind screen. Its counts do not reconstruct the historical 43/23 subsets. All cards, labels, reasons, prompts, and raw outputs are released.

The inclusion rule was fixed before collecting these labels, after the original answer experiments. We reuse all retained answer draws and match the original aggregation and 10,000 user-bootstrap resamples. Table~\ref{tab:fresh-task-repeat} reports repeated-answer and state-reuse estimates. Tables~\ref{tab:fresh-task-accuracy} and~\ref{tab:fresh-task-contrasts} report the four restoration conditions and paired accuracy differences. The Gold intersection retains 32 of 33 compactor restoration pairs and all 36 recent-policy pairs. The corresponding restoration-minus-absent estimates are 0.659 and 0.648; the Strict estimates are 0.700 and 0.594 on 12 and 19 pairs. Restoration also exceeds the non-support replacement in all four comparisons. These results condition on the released model labels and eligible intervention pairs.

\begin{table}[ht]
\centering
\small
\caption{Repeated-answer sensitivity under the new gold-visible reading. Gold and Strict are defined above. A and B denote the original matched padding controls. State reuse retains 33 of 34 target queries under Gold and 14 under Strict. Estimates give equal weight to users; Lower and Upper are 95\% user-bootstrap limits.}
\label{tab:fresh-task-repeat}
\setlength{\tabcolsep}{2pt}
\begin{tabular*}{\linewidth}{@{\extracolsep{\fill}}lrrrrrrrr@{}}
\toprule
 & & & \multicolumn{3}{c}{$D_{\mathrm{res}}$} & \multicolumn{3}{c}{$\Delta_{\mathrm{rep}}$}\\
\cmidrule(lr){4-6}\cmidrule(lr){7-9}
\textbf{Setting / subset} & \textbf{Queries} & \textbf{Users} & \textbf{Estimate} & \textbf{Lower} & \textbf{Upper} & \textbf{Estimate} & \textbf{Lower} & \textbf{Upper}\\
\midrule
Focal A / Gold & 62 & 29 & 0.518 & 0.346 & 0.682 & 0.444 & 0.261 & 0.612\\
Focal A / Strict & 24 & 18 & 0.431 & 0.222 & 0.653 & 0.361 & 0.102 & 0.602\\
Focal B / Gold & 62 & 29 & 0.502 & 0.359 & 0.646 & 0.338 & 0.141 & 0.529\\
Focal B / Strict & 24 & 18 & 0.454 & 0.241 & 0.667 & 0.356 & 0.093 & 0.611\\
State reuse / Gold & 33 & 19 & 0.419 & 0.232 & 0.610 & 0.400 & 0.197 & 0.599\\
State reuse / Strict & 14 & 10 & 0.325 & 0.100 & 0.575 & 0.263 & -0.025 & 0.562\\
\bottomrule
\end{tabular*}
\end{table}
\begin{table}[ht]
\centering
\small
\caption{Restoration accuracies on the new task-preservation intersections. The same Queries and Users apply to both panels. Present, Absent, Restored, and Replacement denote the four matched conditions; Replacement supplies no directly supporting source. Lower and Upper are 95\% user-bootstrap limits.}
\label{tab:fresh-task-accuracy}
\setlength{\tabcolsep}{2pt}
\textbf{A. Original states}\par\vspace{2pt}
\begin{tabular*}{\linewidth}{@{\extracolsep{\fill}}lrrrrrrrr@{}}
\toprule
 & & & \multicolumn{3}{c}{Present} & \multicolumn{3}{c}{Absent}\\
\cmidrule(lr){4-6}\cmidrule(lr){7-9}
\textbf{Setting / subset} & \textbf{Queries} & \textbf{Users} & \textbf{Estimate} & \textbf{Lower} & \textbf{Upper} & \textbf{Estimate} & \textbf{Lower} & \textbf{Upper}\\
\midrule
Compactor / Gold & 32 & 21 & 0.857 & 0.734 & 0.952 & 0.210 & 0.087 & 0.349\\
Compactor / Strict & 12 & 10 & 0.975 & 0.925 & 1.000 & 0.275 & 0.100 & 0.475\\
Recent / Gold & 36 & 18 & 0.940 & 0.866 & 1.000 & 0.269 & 0.144 & 0.407\\
Recent / Strict & 19 & 15 & 0.900 & 0.733 & 1.000 & 0.306 & 0.100 & 0.544\\
\bottomrule
\end{tabular*}
\par\smallskip\textbf{B. Matched interventions}\par\vspace{2pt}
\begin{tabular*}{\linewidth}{@{\extracolsep{\fill}}lrrrrrrrr@{}}
\toprule
 & & & \multicolumn{3}{c}{Restored} & \multicolumn{3}{c}{Replacement}\\
\cmidrule(lr){4-6}\cmidrule(lr){7-9}
\textbf{Setting / subset} & \textbf{Queries} & \textbf{Users} & \textbf{Estimate} & \textbf{Lower} & \textbf{Upper} & \textbf{Estimate} & \textbf{Lower} & \textbf{Upper}\\
\midrule
Compactor / Gold & 32 & 21 & 0.869 & 0.754 & 0.952 & 0.286 & 0.147 & 0.444\\
Compactor / Strict & 12 & 10 & 0.975 & 0.925 & 1.000 & 0.150 & 0.000 & 0.350\\
Recent / Gold & 36 & 18 & 0.917 & 0.833 & 0.981 & 0.167 & 0.074 & 0.273\\
Recent / Strict & 19 & 15 & 0.900 & 0.733 & 1.000 & 0.194 & 0.044 & 0.383\\
\bottomrule
\end{tabular*}
\end{table}
\begin{table}[ht]
\centering
\small
\caption{Paired restoration accuracy contrasts for Table~\ref{tab:fresh-task-accuracy}. Focal denotes the compactor. R denotes Restored, A Absent, N non-support Replacement, and P Present. Lower and Upper are 95\% user-bootstrap limits; all differences are computed within the same users.}
\label{tab:fresh-task-contrasts}
\setlength{\tabcolsep}{2pt}
\begin{tabular*}{\linewidth}{@{\extracolsep{\fill}}lrrrrrrrrrrr@{}}
\toprule
 & & & \multicolumn{3}{c}{$R-A$} & \multicolumn{3}{c}{$R-N$} & \multicolumn{3}{c}{$R-P$}\\
\cmidrule(lr){4-6}\cmidrule(lr){7-9}\cmidrule(lr){10-12}
\textbf{Setting} & \textbf{Queries} & \textbf{Users} & \textbf{Estimate} & \textbf{Lower} & \textbf{Upper} & \textbf{Estimate} & \textbf{Lower} & \textbf{Upper} & \textbf{Estimate} & \textbf{Lower} & \textbf{Upper}\\
\midrule
Focal / Gold & 32 & 21 & 0.659 & 0.480 & 0.813 & 0.583 & 0.409 & 0.750 & 0.012 & -0.036 & 0.071\\
Focal / Strict & 12 & 10 & 0.700 & 0.500 & 0.875 & 0.825 & 0.600 & 1.000 & 0.000 & 0.000 & 0.000\\
Recent / Gold & 36 & 18 & 0.648 & 0.519 & 0.773 & 0.750 & 0.639 & 0.856 & -0.023 & -0.051 & 0.000\\
Recent / Strict & 19 & 15 & 0.594 & 0.361 & 0.811 & 0.706 & 0.500 & 0.889 & 0.000 & 0.000 & 0.000\\
\bottomrule
\end{tabular*}
\end{table}

\FloatBarrier
\subsection{Cross-fitted extension}
\label{app:checks-crossfit}
Table~\ref{tab:fresh-task-crossfit} gives the reversal-route estimates for the two cross-fitted policies on the same Gold and Strict sets. Intersecting the nine-query support-change experiment with Gold retains two queries from two users; both change support identity, and neither belongs to the support-loss stratum. Reference, alternate, restored, and replacement accuracies are 0.750, 1.000, 0.875, and 0.750. Strict retains one of these queries, with corresponding accuracies 1.000, 1.000, 1.000, and 0.750. Thus this intersection supplies no support-loss restoration estimate. The artifact reports the complete memberships, condition estimates, paired intervals, structural strata, and other route schedules, including empty intersections.

\begin{table}[ht]
\centering
\small
\caption{Cross-fitted reversal-route sensitivity under the new reading. Arrival and Source denote the arrival-order policy and source-order comparator. Lower and Upper are 95\% user-bootstrap limits.}
\label{tab:fresh-task-crossfit}
\setlength{\tabcolsep}{2pt}
\begin{tabular*}{\linewidth}{@{\extracolsep{\fill}}lrrrrrrrr@{}}
\toprule
 & & & \multicolumn{3}{c}{$H$} & \multicolumn{3}{c}{$\Delta$}\\
\cmidrule(lr){4-6}\cmidrule(lr){7-9}
\textbf{Setting / subset} & \textbf{Queries} & \textbf{Users} & \textbf{Estimate} & \textbf{Lower} & \textbf{Upper} & \textbf{Estimate} & \textbf{Lower} & \textbf{Upper}\\
\midrule
Arrival / Gold & 62 & 29 & 0.124 & 0.055 & 0.203 & -0.015 & -0.088 & 0.054\\
Arrival / Strict & 24 & 18 & 0.083 & 0.000 & 0.222 & -0.083 & -0.222 & 0.000\\
Source / Gold & 62 & 29 & 0.063 & 0.000 & 0.138 & 0.017 & -0.052 & 0.086\\
Source / Strict & 24 & 18 & 0.000 & 0.000 & 0.000 & 0.000 & 0.000 & 0.000\\
\bottomrule
\end{tabular*}
\end{table}

\section{LongMemEval-S Extension}
\label{app:longmem-scale}

\subsection{Data and unit of analysis}
\label{app:longmem-data}
We use the cleaned LongMemEval-S release archived with the supplementary data. Its 500 questions have complete native session histories. Excluding the 30 abstention questions leaves 470 questions with published evidence-session identifiers: 64 single-session-user, 56 single-session-assistant, 30 single-session-preference, 121 multi-session, 127 temporal-reasoning, and 72 knowledge-update questions. Each question retains its original history of 38--62 sessions, with a median of 47 in this analysis set. We use the released answers and evidence identifiers without new semantic or support annotation. Queries define the aggregation and bootstrap unit; the release does not provide the user identifiers used for PersonaMem's user-level aggregation.

\subsection{Construction and retrieval}
\label{app:longmem-construction}
Forward follows the released session-array order; reverse import reverses that array. Session dates, the question date, and all within-session turns remain unchanged. Recent-$k$ retains the last $k$ imports for $k\in\{8,16,32\}$. Its source-order comparator first orders imports by their original source indices, so it retains the same sessions under either input permutation. We also apply the original focal clustering threshold, preference representative, and last-arrival survivor rule. These focal survivors are measured before the PersonaMem character-budget adapter. All policies then use the same BM25 top-three retrieval interface, which is specific to this extension. The retention budgets count complete sessions, whose lengths vary.

BM25 uses user-turn tokens, the focal tokenizer and stopword list, term-frequency counts, $k_1=1.5$, and $b=0.75$. Document frequencies and mean document length are computed once from the complete fixed history and shared across both paths. Only retained sessions are ranked; ties use original source index. The selected complete sessions are displayed in original source order. Gold answers and evidence-session identifiers enter only the subsequent measurements. The retained-state comparison and the exposed-context comparison are recorded separately.

\subsection{Common-interface comparison}
\label{app:longmem-common-interface}
To separate the retrieval-interface difference from the cross-benchmark direction, we apply the focal compactor with threshold 0.055 and the same fixed-full-pool BM25 top-three retrieval to both datasets. Neither arm applies the PersonaMem character cap. The PersonaMem target is the existing direct-support source set for each primary query; the LongMemEval target is its released evidence-session set. Table~\ref{tab:common-interface} reports the fraction exposed under each import order. PersonaMem first averages questions within each user; LongMemEval weights questions equally. Paired bootstrap resampling uses the corresponding unit. The shared interface retains the opposite directions, with benchmark-specific populations and support definitions remaining distinct. All reconstructed focal survivors match the retained source-level experiments; the LongMemEval exposed sets also match exactly.

\begin{table}[htbp]
\centering\small
\caption{Exposed support under common focal construction and BM25 top-three retrieval. Lower and Upper are 95\% bootstrap limits. PersonaMem uses 63 queries and 29 users; LongMemEval-S uses 470 queries.}
\label{tab:common-interface}
\setlength{\tabcolsep}{2pt}
\begin{tabular*}{\linewidth}{@{\extracolsep{\fill}}lrrrrrrrrr@{}}
\toprule
 & \multicolumn{3}{c}{Forward} & \multicolumn{3}{c}{Reverse} & \multicolumn{3}{c}{Reverse minus forward}\\
\cmidrule(lr){2-4}\cmidrule(lr){5-7}\cmidrule(lr){8-10}
\textbf{Dataset} & Estimate & Lower & Upper & Estimate & Lower & Upper & Estimate & Lower & Upper\\
\midrule
PersonaMem & 0.069 & 0.013 & 0.151 & 0.450 & 0.343 & 0.559 & 0.381 & 0.227 & 0.528\\
LongMemEval-S & 0.146 & 0.118 & 0.175 & 0.106 & 0.083 & 0.132 & -0.039 & -0.071 & -0.007\\
\bottomrule
\end{tabular*}
\end{table}

\subsection{Measurements}
\label{app:longmem-measurements}
For the published evidence-session set $T_q$, recall is $|T_q\cap S_q|/|T_q|$, where $S_q$ denotes the retained or exposed source set. All-covered is the indicator that $T_q\subseteq S_q$. Each statistic is averaged equally over questions. Intervals use 10,000 paired question-bootstrap resamples with seed 20260910. The released per-question records include every policy and both import orders; the source-level experiment issues no model calls.

\begin{table}[htbp]
\centering\small
\caption{Retained sessions on 470 LongMemEval-S questions. Source-order fixes ingestion to original source indices. Lower and Upper are 95\% query-bootstrap limits; F and R denote import order.}
\label{tab:longmem-retained}
\setlength{\tabcolsep}{1pt}
\begin{tabular*}{\linewidth}{@{\extracolsep{\fill}}lrrrrrrrrr@{}}
\toprule
 & \multicolumn{3}{c}{Source overlap} & \multicolumn{3}{c}{Evidence recall F} & \multicolumn{3}{c}{Evidence recall R}\\
\cmidrule(lr){2-4}\cmidrule(lr){5-7}\cmidrule(lr){8-10}
\textbf{Policy} & \textbf{Estimate} & \textbf{Lower} & \textbf{Upper} & \textbf{Estimate} & \textbf{Lower} & \textbf{Upper} & \textbf{Estimate} & \textbf{Lower} & \textbf{Upper}\\
\midrule
Recent-8 & 0.000 & 0.000 & 0.000 & 0.215 & 0.185 & 0.245 & 0.146 & 0.122 & 0.170\\
Recent-16 & 0.000 & 0.000 & 0.000 & 0.381 & 0.345 & 0.416 & 0.319 & 0.287 & 0.352\\
Recent-32 & 0.349 & 0.340 & 0.358 & 0.688 & 0.655 & 0.721 & 0.619 & 0.584 & 0.654\\
Source-order-8 & 1.000 & 1.000 & 1.000 & 0.215 & 0.185 & 0.245 & 0.215 & 0.185 & 0.245\\
Source-order-16 & 1.000 & 1.000 & 1.000 & 0.381 & 0.345 & 0.416 & 0.381 & 0.345 & 0.416\\
Source-order-32 & 1.000 & 1.000 & 1.000 & 0.688 & 0.655 & 0.721 & 0.688 & 0.655 & 0.721\\
Focal & 0.438 & 0.428 & 0.448 & 0.157 & 0.128 & 0.187 & 0.113 & 0.088 & 0.139\\
\bottomrule
\end{tabular*}
\end{table}

\begin{table}[htbp]
\centering\small
\caption{Exposed sessions under the shared BM25 top-three adapter. Definitions and denominator follow Table~\ref{tab:longmem-retained}.}
\label{tab:longmem-exposed}
\setlength{\tabcolsep}{1pt}
\begin{tabular*}{\linewidth}{@{\extracolsep{\fill}}lrrrrrrrrr@{}}
\toprule
 & \multicolumn{3}{c}{Source overlap} & \multicolumn{3}{c}{Evidence recall F} & \multicolumn{3}{c}{Evidence recall R}\\
\cmidrule(lr){2-4}\cmidrule(lr){5-7}\cmidrule(lr){8-10}
\textbf{Policy} & \textbf{Estimate} & \textbf{Lower} & \textbf{Upper} & \textbf{Estimate} & \textbf{Lower} & \textbf{Upper} & \textbf{Estimate} & \textbf{Lower} & \textbf{Upper}\\
\midrule
Recent-8 & 0.000 & 0.000 & 0.000 & 0.209 & 0.180 & 0.239 & 0.140 & 0.117 & 0.164\\
Recent-16 & 0.000 & 0.000 & 0.000 & 0.350 & 0.316 & 0.385 & 0.295 & 0.264 & 0.327\\
Recent-32 & 0.287 & 0.266 & 0.309 & 0.598 & 0.563 & 0.633 & 0.539 & 0.503 & 0.575\\
Source-order-8 & 1.000 & 1.000 & 1.000 & 0.209 & 0.180 & 0.239 & 0.209 & 0.180 & 0.239\\
Source-order-16 & 1.000 & 1.000 & 1.000 & 0.350 & 0.316 & 0.385 & 0.350 & 0.316 & 0.385\\
Source-order-32 & 1.000 & 1.000 & 1.000 & 0.598 & 0.563 & 0.633 & 0.598 & 0.563 & 0.633\\
Focal & 0.227 & 0.210 & 0.246 & 0.146 & 0.117 & 0.175 & 0.106 & 0.083 & 0.132\\
\bottomrule
\end{tabular*}
\end{table}

\begin{table}[htbp]
\centering\small
\caption{Paired exposure-recall change and complete evidence coverage on the same questions. All-covered requires every published evidence session; it measures exposure, not answer correctness.}
\label{tab:longmem-allcovered}
\setlength{\tabcolsep}{1pt}
\begin{tabular*}{\linewidth}{@{\extracolsep{\fill}}lrrrrrrrrr@{}}
\toprule
 & \multicolumn{3}{c}{Recall R minus F} & \multicolumn{3}{c}{All-covered F} & \multicolumn{3}{c}{All-covered R}\\
\cmidrule(lr){2-4}\cmidrule(lr){5-7}\cmidrule(lr){8-10}
\textbf{Policy} & \textbf{Estimate} & \textbf{Lower} & \textbf{Upper} & \textbf{Estimate} & \textbf{Lower} & \textbf{Upper} & \textbf{Estimate} & \textbf{Lower} & \textbf{Upper}\\
\midrule
Recent-8 & -0.070 & -0.110 & -0.029 & 0.096 & 0.070 & 0.123 & 0.040 & 0.023 & 0.060\\
Recent-16 & -0.055 & -0.112 & 0.003 & 0.185 & 0.151 & 0.221 & 0.134 & 0.104 & 0.166\\
Recent-32 & -0.059 & -0.115 & -0.003 & 0.415 & 0.372 & 0.460 & 0.360 & 0.317 & 0.404\\
Source-order-8 & 0.000 & 0.000 & 0.000 & 0.096 & 0.070 & 0.123 & 0.096 & 0.070 & 0.123\\
Source-order-16 & 0.000 & 0.000 & 0.000 & 0.185 & 0.151 & 0.221 & 0.185 & 0.151 & 0.221\\
Source-order-32 & 0.000 & 0.000 & 0.000 & 0.415 & 0.372 & 0.460 & 0.415 & 0.372 & 0.460\\
Focal & -0.039 & -0.071 & -0.007 & 0.102 & 0.077 & 0.130 & 0.064 & 0.043 & 0.087\\
\bottomrule
\end{tabular*}
\end{table}

\subsection{Matched Answers on Native Histories}
\label{app:longmem-restoration}
The released single-session-user subset has 64 non-abstention questions with one evidence-session identifier each. Recent-8 and the shared BM25 interface expose that session on exactly one construction path for 24 questions; all enter the experiment. Present and Absent are the two original exposures. Restored inserts the designated evidence session into Absent, replacing one complete session. Replacement substitutes the highest-ranked non-designated session outside Absent that has the same canonical insertion position. The two shared sessions remain unchanged. Each condition contains three complete sessions with original timestamps, question date, and question text; neutral padding outside the evidence equalizes token counts under \texttt{o200k\_base}.

Luna and Qwen3-8B each produce two free-text answers per condition in fresh ephemeral sessions with runtime tools disabled, totaling 192 answers per model. The original benchmark question and gold answer are used unchanged. A separate Terra max judge receives only the question, released gold answer, and model response. Its yes/no rubric follows LongMemEval's single-session-user evaluation template; this is an adapted judge rather than the benchmark's original scoring model. We aggregate the two judged answers within each question, then weight questions equally. Intervals resample 24 questions 10,000 times; condition seeds are 20260910--20260913 and contrast seeds are 20261010--20261013 in table order. User identifiers are unavailable. Present and Restored remain different contexts in every pair. Degenerate bootstrap intervals reflect identical observed question-level outcomes and do not imply certainty about unseen questions.

\begin{table}[htbp]
\centering\small
\caption{LongMemEval four-condition answer accuracy and paired contrasts, 24 questions. Luna and Qwen3-8B answers are scored by the same masked Terra max judge. Lower and Upper are 95\% question-bootstrap limits.}
\label{tab:longmem-answers}
\setlength{\tabcolsep}{2pt}
\begin{tabular*}{\linewidth}{@{\extracolsep{\fill}}lrrrrrr@{}}
\toprule
\textbf{Condition or contrast} & \multicolumn{3}{c}{\textbf{Luna}} & \multicolumn{3}{c}{\textbf{Qwen3-8B}}\\
\cmidrule(lr){2-4}\cmidrule(lr){5-7}
 & \textbf{Estimate} & \textbf{Lower} & \textbf{Upper} & \textbf{Estimate} & \textbf{Lower} & \textbf{Upper}\\
\midrule
Present & 1.000 & 1.000 & 1.000 & 0.875 & 0.750 & 1.000\\
Absent & 0.000 & 0.000 & 0.000 & 0.000 & 0.000 & 0.000\\
Restored & 1.000 & 1.000 & 1.000 & 0.917 & 0.792 & 1.000\\
Replacement & 0.042 & 0.000 & 0.125 & 0.042 & 0.000 & 0.125\\
\midrule
Present minus Absent & 1.000 & 1.000 & 1.000 & 0.875 & 0.750 & 1.000\\
Restored minus Absent & 1.000 & 1.000 & 1.000 & 0.917 & 0.792 & 1.000\\
Replacement minus Absent & 0.042 & 0.000 & 0.125 & 0.042 & 0.000 & 0.125\\
Restored minus Replacement & 0.958 & 0.875 & 1.000 & 0.875 & 0.667 & 1.000\\
\bottomrule
\end{tabular*}
\end{table}
\subsection{Answer Coverage Across Task Types}
\label{app:longmem-coverage}
The Recent-8 forward/reverse answer comparison includes every non-abstention question in the released set, across all six task types. Inclusion is independent of evidence exposure: questions with support on both paths, one path, or neither path all remain. Construction and BM25 exposure reproduce the corresponding source-level rows exactly. Every prompt contains three complete native sessions in source order, preserving session timestamps, the question date, and the original question. Neutral padding outside the evidence matches total prompt length within each pair under \texttt{o200k\_base}.

Each route receives one new open-ended answer from Luna and Qwen3-8B, yielding 940 answers per model. The common instruction requires personal facts and prior events to come from the supplied records, permits general background knowledge for requested recommendations, and disables tools and external retrieval. A separate Terra max judge receives only the question, released gold answer (or preference rubric), and model response. The task-specific templates are archived from the benchmark's evaluation code: factual and multi-session tasks accept equivalent answers or sufficient intermediate steps and reject partial information, temporal tasks retain the specified off-by-one duration tolerance, knowledge-update tasks accept the updated answer even if older information is also mentioned, and preference tasks assess correct use of personal information. These are the released rubrics applied with a different grading model.

A response is counted as correct when the judge returns yes. We compute forward and reverse accuracy and their paired difference, weighting questions equally, both overall and within each task type (Table~\ref{tab:longmem-coverage-main}). The task-level table reports Luna; Table~\ref{tab:longmem-coverage-models} gives the corresponding overall comparison with Qwen3-8B. Intervals use 10,000 paired question-bootstrap resamples with seed 20260910 and percentile indices 250 and 9750. One answer per route supplies no within-prompt repeat estimate, so this experiment reports correctness contrasts without $D_{\mathrm{res}}$. The earlier four-condition answers remain a separate matrix. Frozen inputs, completed responses, masked grading prompts, and completion-based technical exclusions are retained in \texttt{data/longmem\_coverage/}.

\begin{table}[htbp]
\centering\small
\caption{Recent-8/BM25 answer accuracy on LongMemEval-S for Luna. Lower and Upper give 95\% question-bootstrap limits.}
\label{tab:longmem-coverage-main}
\setlength{\tabcolsep}{1pt}
\begin{tabular*}{\linewidth}{@{\extracolsep{\fill}}lrrrrrrrrrr@{}}
\toprule
 & & \multicolumn{3}{c}{Forward} & \multicolumn{3}{c}{Reverse} & \multicolumn{3}{c}{Reverse minus forward}\\
\cmidrule(lr){3-5}\cmidrule(lr){6-8}\cmidrule(lr){9-11}
\textbf{Task} & Queries & Estimate & Lower & Upper & Estimate & Lower & Upper & Estimate & Lower & Upper\\
\midrule
User & 64 & 0.266 & 0.156 & 0.375 & 0.109 & 0.047 & 0.188 & -0.156 & -0.297 & -0.016\\
Assistant & 56 & 0.286 & 0.179 & 0.411 & 0.232 & 0.125 & 0.339 & -0.054 & -0.214 & 0.107\\
Preference & 30 & 0.267 & 0.133 & 0.433 & 0.067 & 0.000 & 0.167 & -0.200 & -0.367 & -0.067\\
Multi-session & 121 & 0.058 & 0.017 & 0.099 & 0.017 & 0.000 & 0.041 & -0.041 & -0.083 & 0.000\\
Temporal & 127 & 0.126 & 0.071 & 0.189 & 0.071 & 0.031 & 0.118 & -0.055 & -0.126 & 0.016\\
Update & 72 & 0.458 & 0.347 & 0.569 & 0.069 & 0.014 & 0.139 & -0.389 & -0.528 & -0.236\\
\midrule
All & 470 & 0.206 & 0.170 & 0.245 & 0.081 & 0.057 & 0.106 & -0.126 & -0.170 & -0.083\\
\bottomrule
\end{tabular*}
\end{table}

\begin{table}[htbp]
\centering\scriptsize
\caption{Overall Recent-8/BM25 answer comparison on 470 LongMemEval-S questions. Both answer models use the same masked Terra max judge; Lower and Upper give 95\% question-bootstrap limits.}
\label{tab:longmem-coverage-models}
\setlength{\tabcolsep}{2pt}
\begin{tabular*}{\linewidth}{@{\extracolsep{\fill}}lrrrrrrrrr@{}}
\toprule
 & \multicolumn{3}{c}{Forward} & \multicolumn{3}{c}{Reverse} & \multicolumn{3}{c}{Reverse minus forward}\\
\cmidrule(lr){2-4}\cmidrule(lr){5-7}\cmidrule(lr){8-10}
\textbf{Answer model} & \textbf{Estimate} & \textbf{Lower} & \textbf{Upper} & \textbf{Estimate} & \textbf{Lower} & \textbf{Upper} & \textbf{Estimate} & \textbf{Lower} & \textbf{Upper}\\
\midrule
Luna & 0.206 & 0.170 & 0.245 & 0.081 & 0.057 & 0.106 & -0.126 & -0.170 & -0.083\\
Qwen3-8B & 0.204 & 0.168 & 0.243 & 0.089 & 0.064 & 0.115 & -0.115 & -0.157 & -0.072\\
\bottomrule
\end{tabular*}
\end{table}

For multi-session tasks, a corresponding restoration intervention would return the required evidence bundle and compare it with an equally sized non-designated bundle under a declared token budget. Bundles exceeding three sessions require a larger exposure budget.

\end{document}